\pdfoutput=1    
\documentclass{article}

\usepackage[numbers]{natbib}
\usepackage[final]{neurips_2019} 

\usepackage[utf8]{inputenc}

\usepackage[labelfont=bf]{caption} 

\usepackage{CJKutf8}
\usepackage{bold-extra} 

\usepackage{textcomp} 

\usepackage{changepage} 

\usepackage{float} 
\usepackage{afterpage} 

\usepackage{array} 
\usepackage{booktabs} 

\usepackage{hyperref} 

\usepackage[
	activate={true,nocompatibility},
	tracking=true
]{microtype}

\usepackage{xcolor}
\usepackage{multirow}
\usepackage{subcaption}
\usepackage{wrapfig}
\usepackage{amsmath}
\usepackage{adjustbox}

\definecolor{myLinkColor}{rgb}{0.18,0.39,0.62}
\hypersetup{
    colorlinks=true,
    linkcolor=myLinkColor,
    filecolor=myLinkColor,
    urlcolor=myLinkColor,
    citecolor=myLinkColor,
}

\usepackage[disable]{todonotes} 
\usepackage{graphicx} 
\usepackage{tabularx} 
\usepackage{booktabs} 
\usepackage{multirow}
\usepackage{longtable}

\usepackage[titles]{tocloft}
\usepackage{caption} 
\usepackage{chngcntr}
\counterwithin{figure}{section}
\counterwithin{table}{section}

\title{Group Diffusion Transformers are \\ Unsupervised Multitask Learners}
\author{
Lianghua Huang\thanks{Equal contribution. Emails: {
\textlangle xuangen.hlh, ww413411, wuzhifan.wzf \textrangle@alibaba-inc.com
}}
\And
Wei Wang\footnotemark[1]
\And
Zhi-Fan Wu\footnotemark[1]
\And
Huanzhang Dou\thanks{Zhejiang University. The work was done when the author was an intern at Tongyi Lab.}
\AND
Yupeng Shi
\And
Yutong Feng
\And
Chen Liang\thanks{Institute of Automation, Chinese Academy of Sciences. The work was done when the author was an intern at Tongyi Lab.}
\And
Yu Liu
\And
Jingren Zhou
\AND
\\
{\large Tongyi Lab}
}
\date{2024}

\begin{document}

\maketitle
\begin{abstract}
   
While large language models (LLMs) have revolutionized natural language processing with their task-agnostic capabilities, visual generation tasks such as image translation, style transfer, and character customization still rely heavily on supervised, task-specific datasets. In this work, we introduce \textbf{Group Diffusion Transformers (GDTs)}, a novel framework that unifies diverse visual generation tasks by redefining them as a \textbf{group generation} problem. In this approach, a set of related images is generated simultaneously, optionally conditioned on a subset of the group. GDTs build upon diffusion transformers with minimal architectural modifications by concatenating self-attention tokens across images. This allows the model to implicitly capture cross-image relationships (\textit{e.g.}, identities, styles, layouts, surroundings, and color schemes) through caption-based correlations. Our design enables scalable, unsupervised, and task-agnostic pretraining using extensive collections of image groups sourced from multimodal internet articles, image galleries, and video frames. We evaluate GDTs on a comprehensive benchmark featuring over 200 instructions across 30 distinct visual generation tasks, including picture book creation, font design, style transfer, sketching, colorization, drawing sequence generation, and character customization. Our models achieve competitive \textbf{zero-shot} performance without any additional fine-tuning or gradient updates. Furthermore, ablation studies confirm the effectiveness of key components such as data scaling, group size, and model design. These results demonstrate the potential of GDTs as scalable, general-purpose visual generation systems.
\end{abstract}


\setcounter{footnote}{0}
\makeatletter
\let\@fnsymbol\@arabic
\makeatother

%
%
\section{Introduction}
\label{section:Introduction}
\begin{figure*}[p]
\begin{center}
\includegraphics[width=\linewidth]{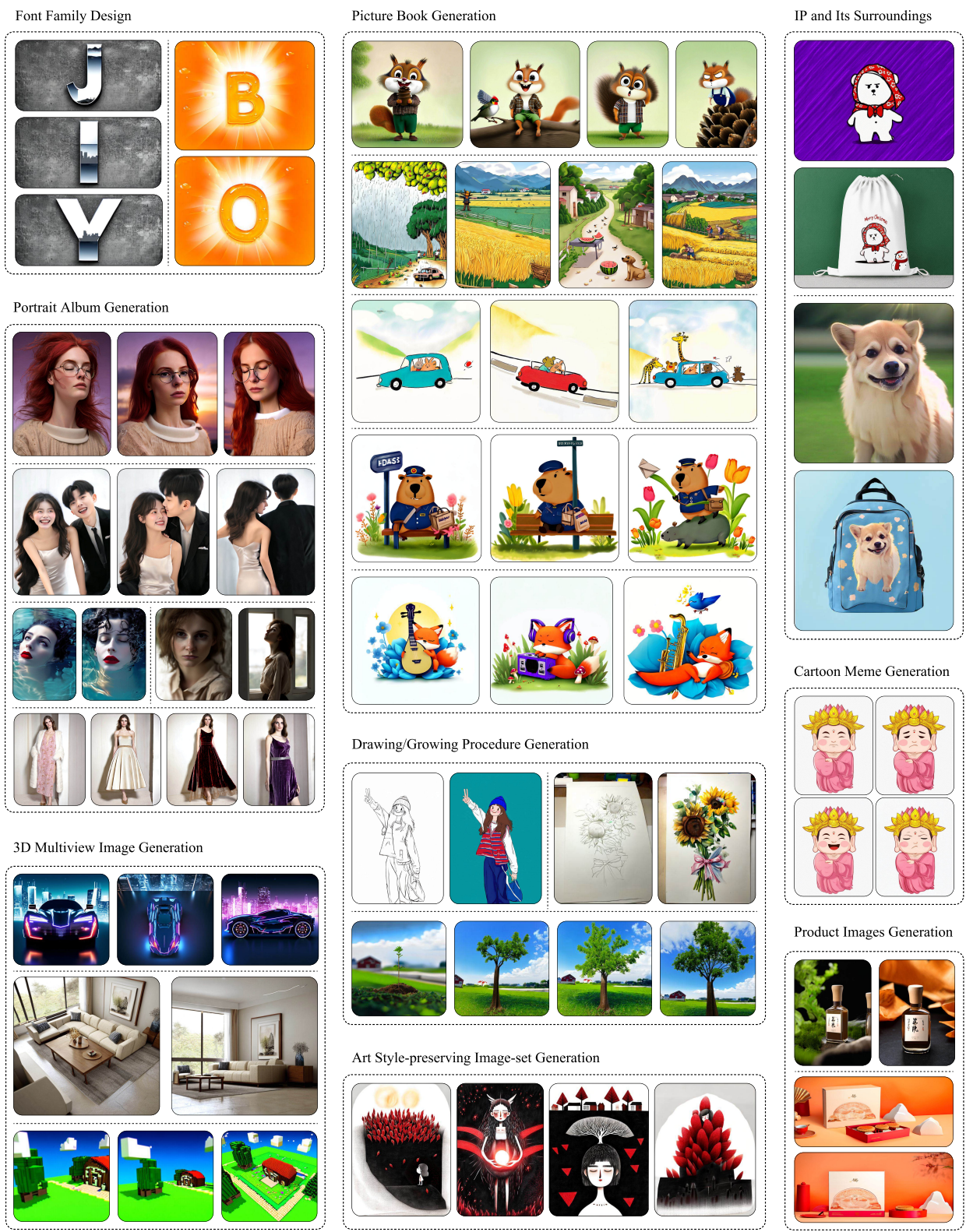}
\end{center}
\caption{\textbf{Group Diffusion Transformers perform a vast array of visual generation tasks in a unified framework termed group generation.}~~~Note that \textbf{NO} task-specific dataset and \textbf{NO} additional gradient update is applied. The model is automatically generalized to these tasks after unsupervised training on image groups. For simplicity, textual descriptions of images are omitted here, which can be found in Appendix.}
\label{figure:intro_cases}
\end{figure*}

The advent of large language models (LLMs) has brought a paradigm shift in natural language processing (NLP) \cite{radford2019language, raffel2020exploring, brown2020language, ouyang2022training, zhang2022opt, touvron2023llama, touvron2023llama2, dubey2024llama}, enabling a wide range of tasks to be approached in a task-agnostic manner. These models, trained on vast corpora, can generate coherent and contextually relevant content across various domains without the need for task-specific fine-tuning, setting a new standard for what is achievable in NLP. However, this level of task generalization has yet to be fully realized in the field of visual generation. Unlike NLP, visual generation tasks -- such as pose transfer \cite{shen2023advancing, lu2024coarse}, image translation \cite{ho2024every, rodatz2024pattern}, customization \cite{jones2024customizing, wei2023elite}, stylization \cite{huang2024diffstyler, yang2023pixel}, and font creation \cite{wang2023cf, yang2024fontdiffuser} -- remain largely siloed, relying heavily on supervised learning paradigms. These tasks often demand extensive task-specific datasets and additional modules, such as LoRAs \cite{jones2024customizing, smith2023continual, luo2023lcm}, adapters \cite{ye2023ip, mou2024t2i}, visual encoders \cite{giannone2022few, kumar2024semantica, xu2024prompt}, and ControlNets \cite{zhang2023adding, zhao2024uni}, to achieve satisfactory performance.

\begin{figure*}[t]
\clearpage
\begin{center}
\includegraphics[width=1.0\linewidth]{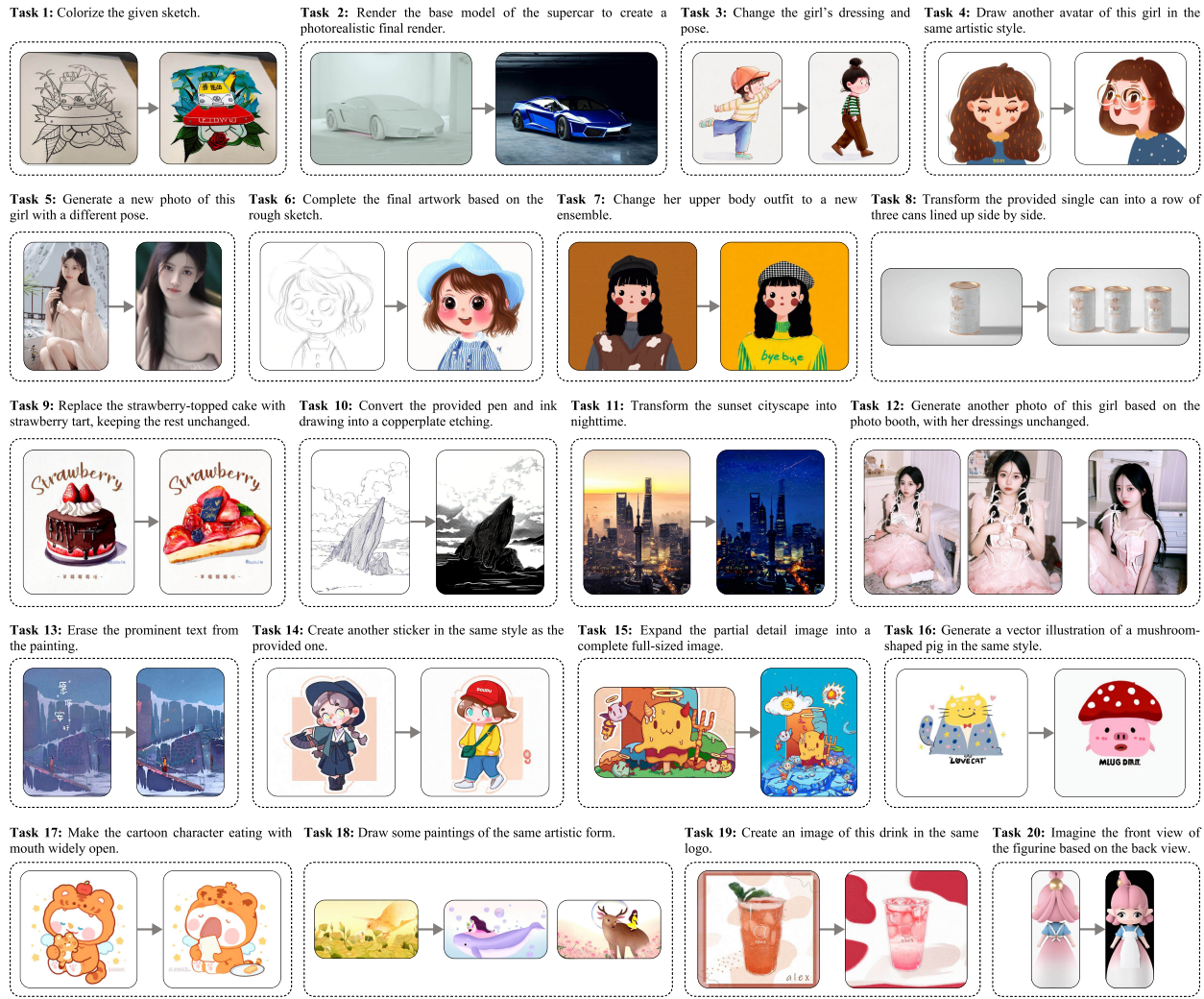}
\end{center}
\caption{\textbf{When conditioned on a subset of the group data, Group Diffusion Transformers could perform conditional group generation in the inpainting fashion.}~~~Note that the model is automatically generalized to these tasks after unsupervised training on image groups. Textual descriptions of images are omitted here (can be found in Appendix), and we summarize them into brief task descriptions. }
\vspace{-1em}
\label{figure:intro_inp_cases}
\end{figure*}

This reliance on specialized data and architectures presents significant challenges for scalability and generalization. First, it limits scalability by failing to leverage the vast amount of weakly supervised data available on the Internet; creating and curating task-specific datasets is human-laboring. Second, it restricts models' adaptability to unseen tasks. Third, cross-task adaptation is lacking, particularly in compositional control, where multiple tasks are implicitly managed. For example, consider creating a picture book, characters, environments, and attire must be dynamically adjusted, requiring decisions on which elements to keep consistent and which to vary. Finally, we hypothesize that training on single-task, shallow-domain datasets leads to the lack of generalization in real-world applications. To truly unlock the potential of visual generation, it is crucial to develop models capable of performing a wide range of tasks in a task-agnostic manner. This demands a shift in how we conceptualize and approach these tasks.

Our key insight is that most, \textit{if not all}, visual generation tasks can be reformulated within a unified framework that we term the \textbf{group generation} problem. In this framework, the objective is to generate a set of correlated data, or a \textit{group}, optionally conditioned on a subset of this group. For instance, tasks such as generating picture books \cite{jin2023generating, wang2023script}, font images \cite{wang2023cf, yang2024fontdiffuser}, or emoticons \cite{mittal2020photo} involve producing multiple images with distinct yet related descriptions simultaneously. The inherent correlations are implicitly captured through the relationships among these descriptions. Conversely, tasks like sketching \cite{voynov2023sketch, wang2023diffsketching}, colorization \cite{zabari2023diffusing, carrillo2023diffusart, liang2024control}, character-specific image generation \cite{zdenek2023handwritten, kou2023character}, and multiview image generation from a single image \cite{liu2023syncdreamer, shi2023zero123++} can be framed as conditional group generation problems, where a subset of the group data is provided as a reference. Figure~\ref{figure:intro_cases} and ~\ref{figure:intro_inp_cases} provide examples of group generation and conditional group generation. By reframing these tasks as group generation problems, we leverage the power of unsupervised learning to address a broad spectrum of tasks without the need for task-specific supervision, simplifying the learning process and broadening applicability.

One of the most compelling advantages of the \textbf{group generation} framework is its natural alignment with the vast amount of data available on the Internet. Multimodal articles, image galleries, and multi-shot videos are just a few examples of readily accessible sources of group data. Each of these sources inherently captures the relationships between different data elements, offering a form of free supervision that is both scalable and diverse. The availability of such abundant group data not only reduces the need for labor-intensive data annotation but also enables the training of models on a wide array of tasks simultaneously, further enhancing generalizability.

To address the group generation problem, we introduce a minimalistic modification to diffusion transformers \cite{Peebles_2023_ICCV, Esser2024ScalingRF, Chen2023PixArtFT}, termed \textbf{Group Diffusion Transformers (GDTs)}. The core idea is to concatenate self-attention tokens across a group of inputs, allowing the model to learn the correlations and variations within the group. This modification is straightforward, requiring minimal changes to the underlying architecture of diffusion transformers (DiTs), yet it significantly enhances the model's ability to capture relationships among multiple generated data. To address reference-based generation problems, such as style transfer \cite{huang2024diffstyler, yang2023pixel} and image translation \cite{ho2024every, rodatz2024pattern}, we incorporate techniques like SDEdit \cite{Meng2021SDEditGI} and inpainting \cite{xie2023smartbrush, xu2024prompt}. These methods enable the model to generate the remaining elements of a group when conditioned on a subset of inputs. Figure~\ref{figure:gdt_framework} provides a detailed architectural overview of GDTs. The straightforward design of GDTs makes it easy to implement and shows promise for efficient scaling.

To evaluate the capabilities of our model, we first introduce a user interface that can automatically convert user instructions into textual descriptions of the target image group to support group generation. Then, we construct a  comprehensive benchmark that covers a wide range of visual generation tasks, both with and without reference images. All tasks are performed in a zero-shot setting, without any parameter or architectural modifications. Despite the absence of task-specific supervision during training, our model demonstrates promising performance across most tasks in human rating and quantitative evaluation. 
Finally, we conduct ablation studies to examine the impact of key components in our framework, such as data scale, group size, model design and quality tuning, on overall performance.

%
%
\section{Approach}
\label{section:Approach}
The core of our approach is to reformulate visual generation tasks into a \textit{group generation} problem and solve it using minimally modified diffusion transformers. We begin by detailing how these tasks are reformulated, followed by a comprehensive introduction to our model, its architecture, the data employed, the training procedure, and the user interface for inference.

    \subsection{Problem Formulation}
    \label{section:Problem Formulation}
    We propose that a vast array of visual generation tasks can be unified under a single framework we term the \textbf{group generation} problem. In this framework, the objective is to generate a group of $n$ elements $\mathbf{x} = \{\mathbf{x}_1, \mathbf{x}_2, \cdots, \mathbf{x}_n\}$, where each element is conditioned on its respective context (\textit{e.g., image descriptions}) $\mathbf{c} = \{\mathbf{c}_1, \mathbf{c}_2, \cdots, \mathbf{c}_n\}$. The relationships among these elements are implicitly defined by the interdependencies within their contextual conditions. Optionally, a subset of $0\leq m < n$ elements of $\mathbf{x}$ can be provided as reference data, with the task being to generate the remaining $(n - m)$ elements. This formulation naturally encapsulates a variety of tasks:

\begin{itemize}
    \item \textbf{Text-to-Image:} A special case where the group size $n=1$ and the reference subset size $m=0$. The task is to generate a single image from a textual description.
    \item \textbf{Font Generation:} Here, the group size $n>1$ corresponds to the number of characters to generate, with $m=0$.
    \item \textbf{Picture Book Generation:} Similar to font generation, the group size $n>1$ corresponds to the number of picture book pages, with $m=0$. The descriptions capture the connections and variations across the pages.
    \item \textbf{Identity Preservation:} Here, the group size $n>1$ corresponds to the number of photos with the same identities to generate, with $m=0$. Identity-specific information is reflected in the descriptions, such as names or other identifiers.
    \item \textbf{Local Editing:} In this task, the group size is $n=2$ with a reference subset size $m=1$. One reference image is provided, and the model generates the edited image based on the differences captured in their descriptions.
    \item \textbf{Image Translation:} Similarly, the group size is $n=2$ with a reference subset size $m=1$. A reference image from one domain is converted to another domain according to their descriptions.
    \item \textbf{Subject Customization:} The task involves generating $(n - m)\geq 1$ images, where $1\leq m < n$ character images are used as references.
    \item \textbf{Style Preservation:} In this task, $(n - m)\geq 1$ corresponds to the number of stylized images to be generated, with $m=1$ being the reference image guiding the target style. 
\end{itemize}

These examples illustrate just a few of the many tasks that can be naturally expressed within the \textit{group generation} framework. Across these tasks, the task hints are naturally embedded within the group element descriptions, much like how a human might communicate with a designer. This unified framework simplifies the approach to diverse visual generation tasks and paves the way for scalable, generalized solutions.

    \subsection{Model and Architectures}
    \label{section:Model and Architectures}
    \begin{figure}[t]
\begin{center}
\includegraphics[width=1.0\linewidth]{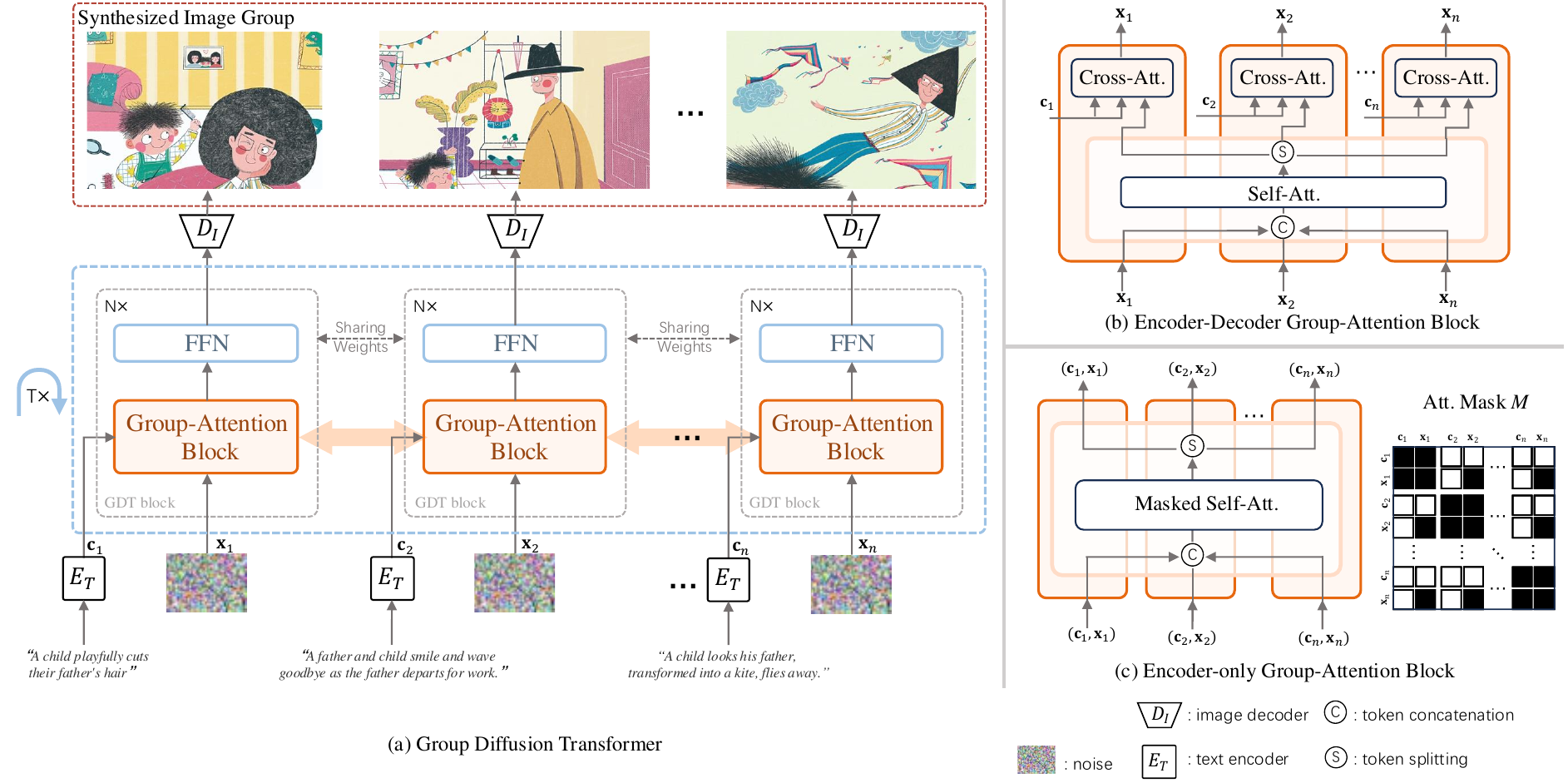}
\end{center}
\caption{\textbf{The overview of Group Diffusion Transformer, which takes minimal adaptations for the encoder-decoder and encoder-only visual generation architectures.}~~~We make a straightforward modification on self-attention blocks by concatenating image tokens across group inputs, allowing to learn cross-image correlations.}
\label{figure:gdt_framework}
\end{figure}

To tackle the group generation problem, it is crucial to establish connections between multiple group elements during the generation process, allowing the model to perceive and utilize the correlations among these elements. Our approach involves a straightforward modification: concatenating tokens across group inputs within the self-attention blocks of diffusion transformers. This enables tokens from different data elements to interact with one another throughout the model's layers.

For different text-conditioned visual generation architectures, we make minimal adaptations to accommodate our approach:

\begin{itemize}
    \item \textbf{Encoder-Decoder:} In architectures like PixArt \cite{Chen2023PixArtFT}, each transformer block includes a self-attention operation for the image, cross-attention for interaction between image and text, and a feed-forward network. We choose to concatenate all the image tokens in self-attention blocks, which allows every token attends to all the image tokens within the group. After self-attention operation, concatenated image tokens are split back correspondingly. Then, in cross-attention blocks, each image token attends only to the text embeddings associated with its respective description. This setup is illustrated in Figure~\ref{figure:gdt_framework}~(b).

    \item \textbf{Encoder-Only:} Examples like Stable Diffusion 3 \cite{Esser2024ScalingRF} and FLUX \cite{flux1} feature transformer blocks with self-attention blocks and feed-forward networks. We modify the self-attention operation into a masked version, 
    which is depicted in Figure~\ref{figure:gdt_framework}~(c). Specifically, image tokens $\mathbf{x}_{i}$ as well as  text tokens $\mathbf{c}_i$ are first concatenated with each other all over the group. Then, we perform the masked self-attention, where the mask is designed for allowing every image token attends to all tokens across the group while allowing context tokens only attend to their associated image tokens as well as themselves. 
    Concretely, let $M{(\mathbf{a}_j, \mathbf{b}_k)}$ indicate the attention mask for tokens in $\mathbf{a}_j$ and $ \mathbf{b}_k$, where $ \mathbf{a}, \mathbf{b}\in \{\mathbf{c}, \mathbf{x}\}, 0\le j,k \le n$. Then, $M{(\mathbf{a}_j, \mathbf{b}_k)}$ is decided by
    \begin{equation}
        M{(\mathbf{a}_j, \mathbf{b}_k)} = 
        \begin{cases} 
            1 & \text{if } (j = k) \text{ or } (\mathbf{a} \in \mathbf{x} \text{ and }  \mathbf{b} \in \mathbf{x}) \\ 
            0 & \text{else} 
        \end{cases}.
    \end{equation}
  
\end{itemize}

    \subsection{Training Dataset}
    \label{section:Training Dataset}
    We focus on image-related tasks in this work, which requires a high-quality, large-scale, and diverse image group dataset. While existing multimodal datasets like MINT-1T \cite{awadalla2024mint} are large, they fall short of our pretraining needs due to low image quality and biased group type distribution relative to real-world visual generation applications. Thus, we construct our own dataset by sourcing image groups from multimodal Internet articles.

\begin{wrapfigure}{r}{0.47\linewidth}
\centering
\includegraphics[width=1.0\linewidth]{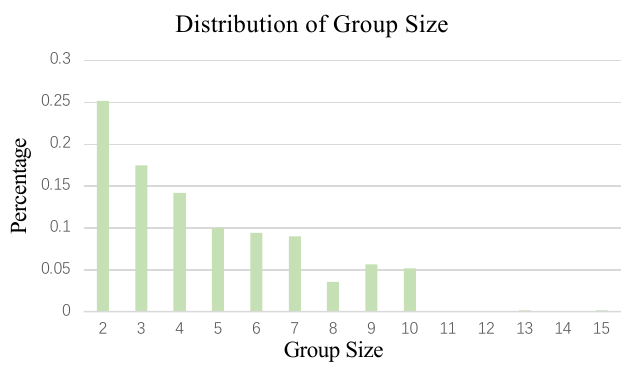}
\caption{\textbf{Distribution of group size in our training dataset.}}
\label{figure:groupsizes}
\end{wrapfigure}

Our dataset creation process involve several key steps: (1) We collect a substantial amount of multimodal data, extracting images while preserving their original order to maintain group integrity. (2) A small subset of these image groups is manually annotated as either positive (suitable for retention) or negative (to be discarded). (3) Using these annotations, we train a binary classifier to score and filter the collected image groups. (4) We perform deduplication across and within groups to eliminate redundant groups and images. After preprocessing, we compile a dataset of approximately 500,000 image groups, with the distribution of group size illustrated in Figure~\ref{figure:groupsizes}.

\begin{figure}[h]
\begin{center}
\includegraphics[width=1.0\linewidth]{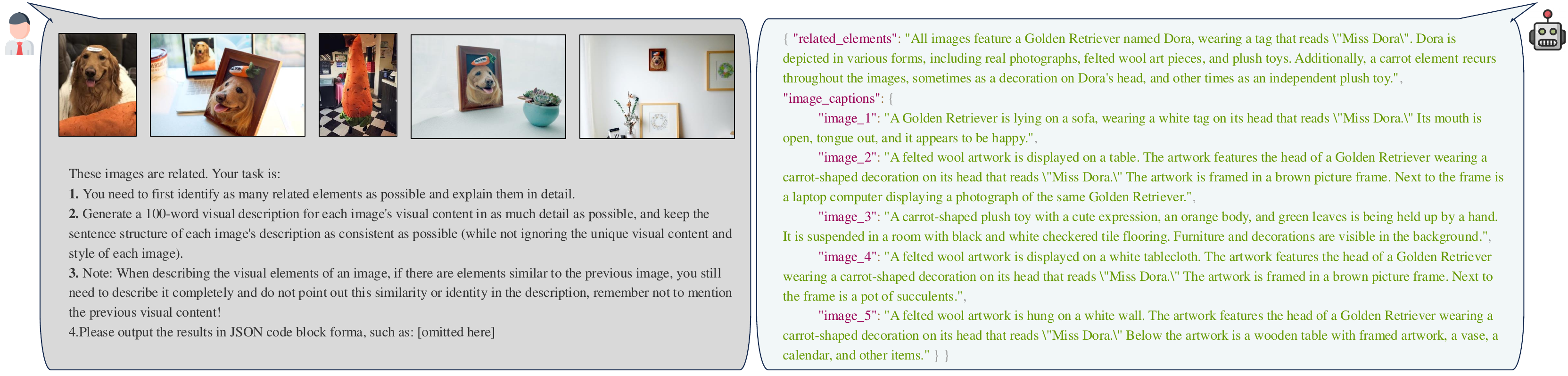}
\end{center}
\caption{\textbf{Example of our training dataset, where the group images are captioned through prompting our internal MLLMs.}~~~}
\label{figure:gemini_case}
\end{figure}

The next crucial step is to generate descriptions that accurately capture the correlations among the images within each group. To achieve this, we utilize our internal multimodal large language models (MLLMs), iteratively testing and refining prompts to ensure the generated descriptions are stable and applicable across different group types. In Figure~\ref{figure:gemini_case}, we show the prompt we used, as well as the resulting group image descriptions.

While pretraining on our large-scale dataset provides a solid foundation for learning correlations with Group Diffusion Transformers (GDTs), it is common practice in visual generation tasks to conduct a supervised fine-tuning stage to enhance generation details and aesthetics. To this end, we curate a smaller, high-quality subset of approximately 10,000 image groups. These groups were selected for their strong correlations, high image quality, aesthetic appeal, and diversity. Fine-tuning our pretrained models on this curated dataset significantly improves both the image quality and content consistency in group generation, where the comparison can be found in Section~\ref{section:Quality Tuning}.

    \subsection{Training Process}
    \label{section:Training Process}
    We initialize the Group Diffusion Transformers (GDTs) with weights from pre-trained text-to-image models, such as PixArt-$\alpha$ \cite{Chen2023PixArtFT} and Stable Diffusion 3 \cite{Peebles_2023_ICCV}. Since GDTs introduce no additional parameter to the existing diffusion transformers, the pretrained weights are fully compatible. During both pretraining and supervised fine-tuning, we uniformly sample group sizes ranging from 1 to 4, dynamically adjusting the batch size to maintain consistent GPU memory usage. This approach ensures balanced performance across different group sizes. The model undergoes pretraining for approximately 100,000 steps, followed by fine-tuning on a curated dataset for around 5,000 steps. All training is conducted on A100 GPUs. We adopt the same hyperparameter settings as the official models in PixArt-$\alpha$ and Stable Diffusion 3.

    \subsection{User Interface}
\begin{figure}[t]
\begin{center}
\includegraphics[width=1.0\linewidth]{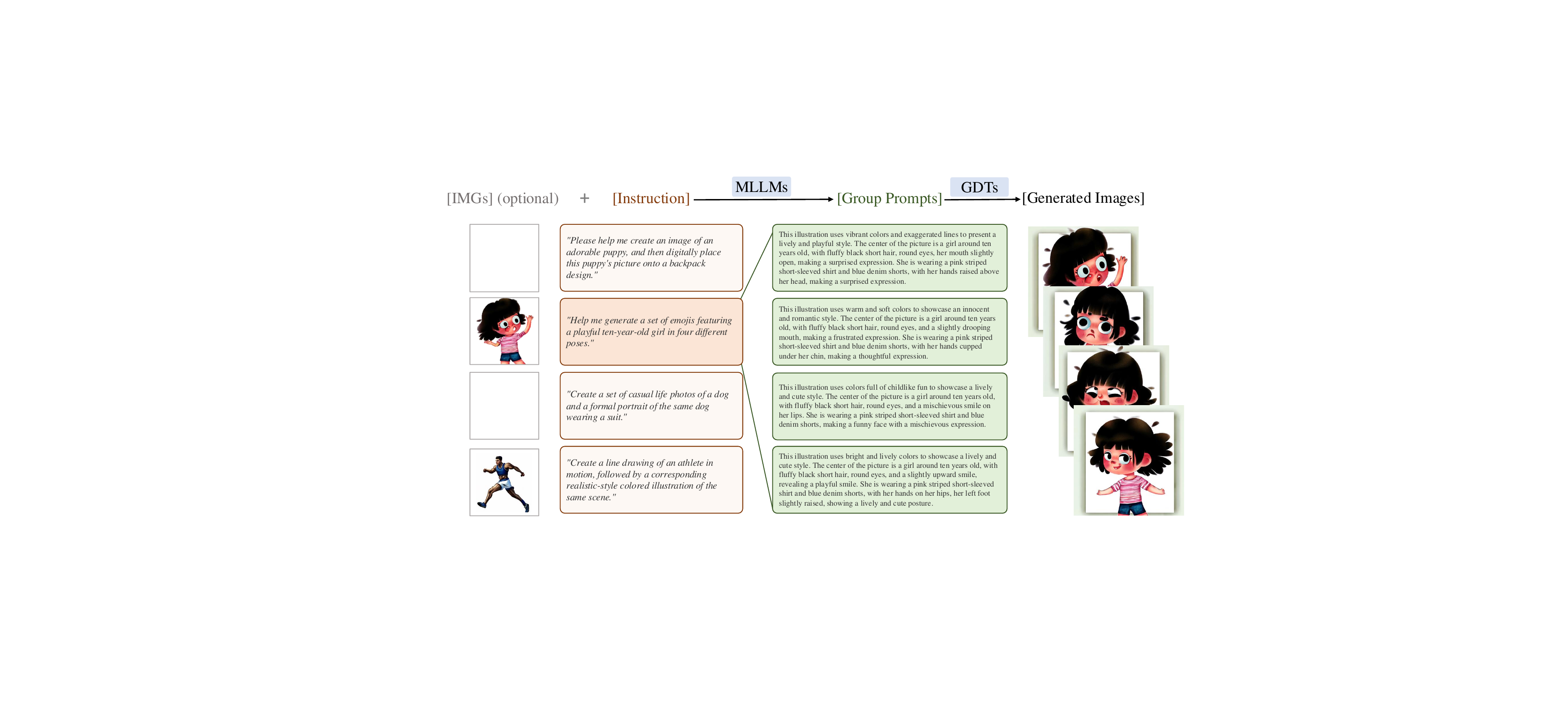}
\end{center}
\caption{\textbf{We build a user interface that automatically converts the user instruction into group prompts using MLLMs, which is useful in the inference stage of GDTs.}~~~}
\label{figure:user interface}
\end{figure}

    Considering it is tedious to write a group of prompts in the inference stage, we build a user interface to provide a convenient interaction with GDTs. As illustrated in Figure~\ref{figure:user interface}, we follow the pipeline of [\textbf{\texttt{Instruction}}] $\rightarrow$ [\textbf{\texttt{Group Prompts}}] $\rightarrow$ [\textbf{\texttt{Generated Images}}] for group generation, and [\textbf{\texttt{IMGs}}] + [\textbf{\texttt{Instruction}}] $\rightarrow$ [\textbf{\texttt{Group Prompts}}] $\rightarrow$ [\textbf{\texttt{Generated Images}}] for conditional group generation. Specifically, we prompt MLLMs to convert the user instruction into group prompts, where the MLLMs are asked to analyze the number of group prompts and the corresponding tasks. For example, if the instruction is ``Draw a line sketch of a female character and the corresponding colored photo", the MLLM can deduce that this instruction should be transformed into two prompts, categorizing the task as sketch coloring.

\section{Benchmark}
\label{section:Benchmark}

\begin{wrapfigure}{l}{0.5\linewidth}
\centering
\includegraphics[width=1.0\linewidth]{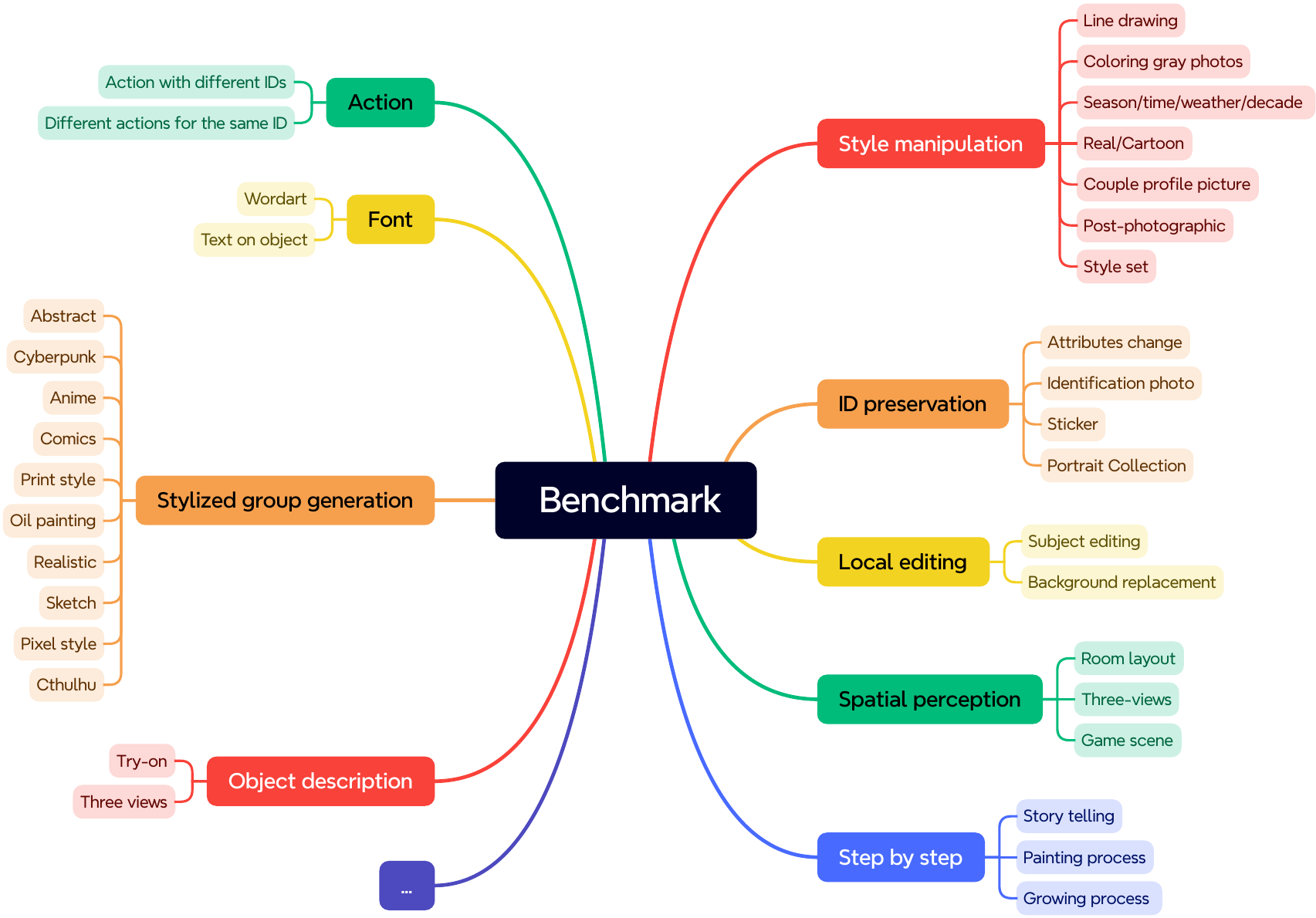}
\caption{\textbf{Overview of our benchmark, covering about 30 distinct types of generation tasks.}}
\vspace{-3em}
\label{figure:benchmark}
\end{wrapfigure}

Given the diverse nature of visual generation tasks, evaluating the performance of our GDTs presents unique challenges. Therefore, we design a benchmark~\footnote{Benchmark is available at \href{https://drive.google.com/drive/folders/1e7muQDqJ-kV4H5yzGMCzCHORpz3ut1CN?usp=sharing}{Google Drive}.} that spans a wide array of tasks as shown in Figure~\ref{figure:benchmark}. Specifically, our benchmark consists of over 200 instructions, each corresponding to one of 30 distinct types of visual generation tasks. This diversity enables a thorough assessment of the generalization capabilities of GDTs across various scenarios. 

This evaluation suit encompasses tasks such as identity preservation, local editing, subject customization, font generation, stylized group generation, and step-by-step generation. Among these coarse-grained categories, further fine-grained tasks are expanded. For example, step-by-step generation contains subtasks like story telling~\cite{Zhou2024storydiffusion}, painting process~\cite{song2024processpainterlearnpaintingprocess}, and growth process. Besides, all the textual descriptions in this benchmark are created through our user interface.

%
%
\section{Results}
\label{section:Results}

    \vspace{-0.5em}
    \subsection{User Study}
    \label{section:User Study}
    
\begin{wraptable}{h}{0.45\textwidth}
\footnotesize
\centering
\vspace{-1em}
\caption{\textbf{User study on our benchmark.}~~Human evaluation on three questions in a five-point scale.}
\begin{tabular}{lccc}
\toprule
\textbf{Models}                & \textbf{Q1}   & \textbf{Q2}   & \textbf{Q3}   \\ \midrule
\textbf{group generation} & & & \\ 
PixArt-$\alpha$           & 3.44 & 3.89 & 3.78 \\
Stable Diffusion 3              & 3.20 & 3.35 & 3.29 \\
\midrule
\textbf{conditional group generation} & & & \\ 
PixArt-$\alpha$ & 3.15 & 3.56 & 3.68 \\
Stable Diffusion 3 & 3.02 & 3.27 & 3.34 \\ \bottomrule
\end{tabular}
\label{tab:user_study}
\vspace{-1em}
\end{wraptable}

\label{section:user study}

\begin{figure}[t]
\begin{center}
\includegraphics[width=1.0\linewidth]{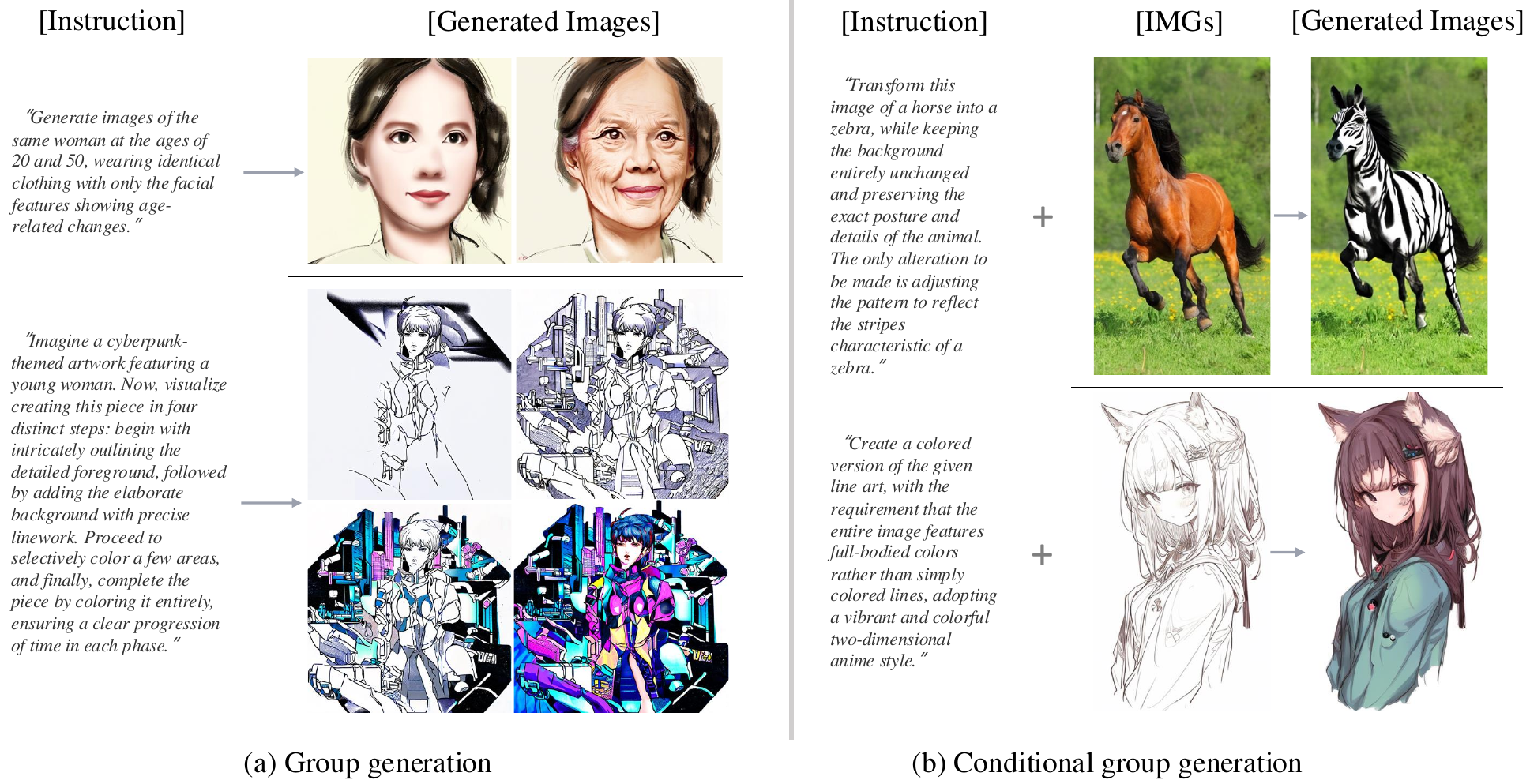}
\end{center}
\caption{\textbf{Qualitative results of GDTs on our benchmark, including group generation and conditional group generation.}~~~}
\vspace{-1em}
\label{figure:benchmark cases}
\end{figure}

We first qualitatively evaluate the generated results of GDTs on our proposed benchmark as shown in Figure~\ref{figure:benchmark cases}. GDTs could perform both group generation and conditional group generation according to the user instructions. Note that the task scope of this benchmark is effectively limited by our imagination, but thanks to our task-agnostic pretraining, GDTs can theoretically be generalized to \textit{arbitrary} visual generation tasks. 

In our user study, we mainly adopt human ratings to assess the performance of GDTs on the benchmark. Three questions are included to measure the prompt following ability, content consistency within the image group, and the overall instruction following ability, namely: \textbf{Q1: Prompt following on each image within the group}:  \textbf{Q2: Content consistency among generated group images, regardless of prompts}, \textbf{Q3: Instruction following on the generated group images.} Evaluators are asked to rate on three questions in the scale from 1 to 5, where 5 signifies perfection and 1 denotes the lowest quality. The final evaluation score is derived from the average ratings across all tasks, which serves as a robust indicator of the overall performance and its potential for real-world applications. The human-rated results are illustrated in Table~\ref{tab:user_study}, where GDTs achieve overall satisfaction (higher than 3) on all of the three questions.

    \subsection{Ablation Analysis}
    \label{section:Ablation Analysis}
    
\vspace{-0.5em}
\subsubsection{Metrics}
\label{section:Metrics}
While our benchmark with over 200 instructions could well evaluate model's capabilities on a five-point scale, we would like to compare these ablated models in a more nuanced and quantitative manner in our ablation experiments. Therefore, we mainly present the objective metrics like FID and CLIP score. To be specific, we measure image fidelity by calculating FID on the validation set using 50k images. We assess content consistency and prompt adherence within each group by averaging CLIP similarities across every image-image and image-text pairs, respectively. 
In terms of reference-based generation, we adopt the same metrics but exclude pairs that involve the reference images themselves, as well as pairs between reference images and their corresponding texts.

\begin{table}
    \caption{\textbf{Performance evaluation on key components of GDTs.} We investigate the impacts of data scale, group size, model design, and quality tuning on encoder-decoder and encoder-only models. }
    \vspace{-0.5em}
    \centering
    \renewcommand{\arraystretch}{1.03} 
    \begin{tabular}{p{4.3cm} c c c c c c c c}
    \toprule
    \multirow{3}{*}{\bf{Settings}} & \multicolumn{3}{c}{\textbf{PixArt-$\alpha$ (Encoder-Decoder)}} & \multicolumn{4}{c}{\textbf{Stable Diffusion 3 (Encoder-Only)}} \\ 
    \cmidrule{2-4} \cmidrule{6-8}
     & \shortstack{{FID-50k} \\ \\ {}} & \shortstack{{Content} \\ {Consistency}} & \shortstack{{Prompt} \\ {Adherence}} & & \shortstack{{FID-50k} \\ \\ {}} & \shortstack{{Content} \\ {Consistency}} & \shortstack{{Prompt} \\ {Adherence}} \\ 
    \midrule
    \textbf{Data Scaling} &   &     &   \\
    5k groups & {8.40} & {0.747}   & 0.291 & & 8.95 & 0.740 & 0.298\\
    50k groups & {12.06} &{0.767} &  0.293 & & 10.92 & 0.760 & 0.302\\
    500k groups & {15.91} &{0.778} &  0.300 &  & 11.30 & 0.761 & 0.305 \\
    \midrule
    \textbf{Group Size} &   &     &   \\
    groupsize = 2 & {15.69} & {0.784} & 0.299 & & 12.37 & 0.763 & 0.301\\
    groupsize = 4 & {18.19} & {0.761} & 0.291 & & 13.85 & 0.739 & 0.298\\
    groupsize = 8 & {48.26} & {0.701} & 0.252 & & 18.28 & 0.701 & 0.290\\
    \midrule
    \textbf{Inpainting} &   &     &   \\
    SDEdit & {15.71} &{0.702} &  0.299 & & 12.15 & 0.751 & 0.303\\
    trainable & {10.91} & {0.725}   & 0.287 & & 10.94 & 0.755 & 0.298\\
    \midrule
    \textbf{Quality Tuning} &   &     &   \\
    before & {15.91} &{0.778} &  0.300 &  & 11.30 & 0.761 & 0.305\\
    after & {12.53} & {0.792}   & 0.298 & & 10.03 & 0.781 & 0.303\\
    \bottomrule
    \end{tabular}
    \label{table:ablation}
    \vspace{-1em}
\end{table}

\vspace{-0.5em}
\subsubsection{Data Scaling}
\label{section:Data Scaling}
Without the demand of task-specific supervision, it is quite easy to acquire a large abundance of group data from the Internet. We scale the training data to 5k, 50k, and 500k groups, to explore the impact of data scale in GDTs. As illustrated in Table~\ref{table:ablation}, with the increase of the amount of training data, GDTs behave increasingly better in content consistency and prompt adherence. Interestingly, we find that FID would become lower when training on less data, {\color{black}which may be that it is easier to overfit to small datasets.} We plan to further scale up our data to the level of hundreds of millions of groups in the future, in order to fully leverage the potential of GDTs.

\vspace{-0.5em}
\subsubsection{Group Size}
\label{section:Impact of Group Size}
We gradually increase the upper limit of group size to  2, 4, and 8, and perform inference based on that limit. Note that doubling the group size will, in turn, double the sequence length in self-attention, leading to a corresponding increase in computational complexity, so we cap the maximum group size at 8 in our ablation. From the ablated results in Table~\ref{table:ablation}, we find that larger group sizes lead to a more pronounced performance decline in image quality, content consistency, and prompt adherence. The reason may be that it is more difficult to learn the complex relationships across a large group of images. Besides, the scarcity of data of large group sizes prevents the model from being adequately trained. In the future, we would greatly scale our training data.

\vspace{-0.5em}
\subsubsection{SDEdit or Inpainting}
\label{section:SDEdit or Inpainting}
When conditioned on a subset of the group data, using methods like SDEdit \cite{Meng2021SDEditGI} or trainable inpainting \cite{xie2023smartbrush, xu2024prompt}, GDTs can be instructed to generate the remaining data of the group. Specifically, SDEdit is a training-free technique which provides the reference images that are added with the same noise step as the generated images during the denoising stage. In contrast, trainable inpainting concatenates the reference image to the noised one in channel dimension, allowing the model to ``copy" the reference images and generate the remaining ones. In our ablation study, as illustrated in Table~\ref{table:ablation}, it is observed that trainable inpainting performs better in image quality and content consistency, while the training-free SDEdit is good at prompt adherence. We adopt the model design of trainable inpainting in our GDTs.

\vspace{-0.5em}
\subsubsection{Quality Tuning}
\label{section:Quality Tuning}
While quality tuning is a common practice in visual generation models to enhance aesthetic appeal, we investigate its impact under the paradigm of group generation. As illustrated in Table~\ref{table:ablation}, after the supervised fine-tuning on a small subset of high-quality image groups, GDTs exhibit significantly better image quality. We also find that quality tuning helps generating image groups with higher content consistency, while barely compromising the adherence to textual descriptions.

\section{Related Work}
\label{section:Related Work}
\subsection{Text-to-Image Generation}
The emergence of DDPM~\cite{NEURIPS2020_4c5bcfec} has catalyzed rapid advancements in text-to-image (T2I) generation. Earlier frameworks focused on T2I generation in pixel space, exemplified by GLIDE~\cite{nichol2022glidephotorealisticimagegeneration} and Imagen~\cite{saharia2022photorealistictexttoimagediffusionmodels}. In contrast, Stable Diffusion~\cite{Rombach_2022_CVPR} introduced latent space for T2I generation, while DALLE-2 (unCLIP)\cite{Ramesh2022HierarchicalTI} expanded this to a multimodal latent space. EMU\cite{dai2023emuenhancingimagegeneration} demonstrated that supervised fine-tuning on a small set of appealing images can significantly enhance generation quality. Unlike U-Net architectures, several approaches, including DiT~\cite{Peebles_2023_ICCV}, Pixart~\cite{Chen2023PixArtFT}, HunyuanDiT~\cite{li2024hunyuanditpowerfulmultiresolutiondiffusion}, and SD3~\cite{esser2024scalingrectifiedflowtransformers}, adopt transformers as their backbone.

\subsection{Controllable Text-to-Image Generation}

\textbf{Personalization.} The personalization task~\cite{cui2024idadapterlearningmixedfeatures,ViPer,Ham_2024_CVPR,wang2024instantid,tao2024storyimager} seeks to capture and utilize concepts as generative conditions, such as subject~\cite{li2023blipdiffusion,kumari2022customdiffusion}, person~\cite{xiao2023fastcomposer,li2023photomaker,Chen_Fang_Liu_He_Huang_Mao_2024,chen2023photoverse}, style~\cite{liu2023stylecrafter,sohn2023styledrop}, and image~\cite{ye2023ip-adapter,Xu_2023_ICCV,ramesh2022hierarchicaltextconditionalimagegeneration}. Textual Inversion~\cite{gal2022textual} embeds user-provided concepts into optimized words within the text embedding space, while DreamBooth~\cite{ruiz2022dreambooth} uses low-frequency words to represent concepts, subsequently updating UNet parameters. Subject-driven methods utilize additional encoders to represent facial identity information, as seen in Face0~\cite{10.1145/3610548.3618249} and DreamIdentity~\cite{Chen_Fang_Liu_He_Huang_Mao_2024}, which employ pre-trained face recognition models. A variety of personalization technologies continue to emerge to satisfy diverse application needs.

\textbf{Spatial Control.} Given the challenge of accurately representing the spatial structure of generated images through text~\cite{Li_2023_CVPR}, such as layout~\cite{liu2024trainingfreecompositescenegeneration,cheng2023layoutdiffuseadaptingfoundationaldiffusion,dou2024gvdiffgroundedtexttovideogeneration}, human pose, and segmentation masks, controllable T2I generation with various positional signals is garnering increasing attention in AIGC literature~\cite{Li_2023_CVPR,bansal2024universal,huang2023composer,liu2023customizable,mou2023t2i}. ControlNet~\cite{zhang2023adding} introduces an additional encoder with "zero-initialization," while UniControl~\cite{qin2023unicontrol} encodes conditions using mixtures of experts (MoE) adaptors.

\textbf{Advanced Control.} To address the complex needs of real applications, controlled generation technology is evolving in several new directions. Attend-and-Excite~\cite{attendexcite} refines cross-attention mechanisms to enhance text alignment. Composer~\cite{huang2023composer} and Cocktail~\cite{hu2023cocktail} employ joint training for multi-condition generation. Cones~\cite{Liu2023ConesCN} fine-tunes concept neurons post-personalization to improve quality and enable multi-subject generation. Universal Guidance~\cite{bansal2023universalguidancediffusionmodels}, EMU2~\cite{sun2024generativemultimodalmodelsincontext}, and FreeDom~\cite{yu2023freedom} strive to achieve universal controllable T2I generation.

\subsection{Attention Adjustment in Diffusion Models}
Some previous works \cite{hertz2024style,Zhou2024storydiffusion,zeng2024jedi,hertz2022prompt} have shown that adjusting the attention operation in diffusion models could help preserve the content and style across the generated images. Particularly, JeDi~\cite{zeng2024jedi}, StyleAligned~\cite{hertz2024style}, and StoryDiffusion~\cite{Zhou2024storydiffusion} modify the self-attention operation to allow multiple images to attend to each other. Although GDTs adopt a similar design, we are fundamentally different with the following capabilities: (1) general-purpose visual generation system; (2) no need for task-specific learning or finetuning; (3) easily scalable, unsupervised visual pretraining; (4) enabling zero-shot task generalization.

\section{Conclusion and Limitations}
\label{section:conclusion}
We reformulate most visual generation tasks into a \textit{group generation} problem, thereby introducing a unified framework named \textbf{Group Diffusion Transformers} (GDTs). We present that with scalable, unsupervised, and task-agnostic pretraining on group data, GDTs could achieve competitive zero-shot performance on a vast array of visual generation tasks. Our results demonstrate the potential of GDTs as scalable, general-purpose visual generation systems.

Moreover, we point out that there is still a discrepancy in image quality between GDTs and the state-of-the-art text-to-image models. The amount of group data for pretraining is also not sufficient yet, which has not fully unleashed the model's capabilities. We are optimistic that with an enlarged group dataset, we can further optimize the model's performance and reduce the discrepancy. In the future, we also plan to extend the time dimension of GDTs to enable multi-shot video generation, which can be naturally expressed under our group generation framework.

\clearpage
\newpage
\bibliographystyle{unsrtnat}
\bibliography{bib}

\clearpage
\newpage
\appendix
\setcounter{figure}{0}
\section{Appendix}

{\tiny
\begin{longtable}{@{}m{0.15\textwidth} m{0.3\textwidth} | m{0.15\textwidth} m{0.3\textwidth}@{}} 
\caption{Textual descriptions of the images in Figure 1.1.}\label{appendix_tab1}\\
\toprule
\textbf{Image} & \textbf{Description} & \textbf{Image} & \textbf{Description} \\ 
\midrule
\endhead

\multicolumn{4}{r}{\emph{Continued on next page}} \\ 
\endfoot

\bottomrule
\endlastfoot

\centering
\includegraphics[width=\linewidth, height=3cm, keepaspectratio]{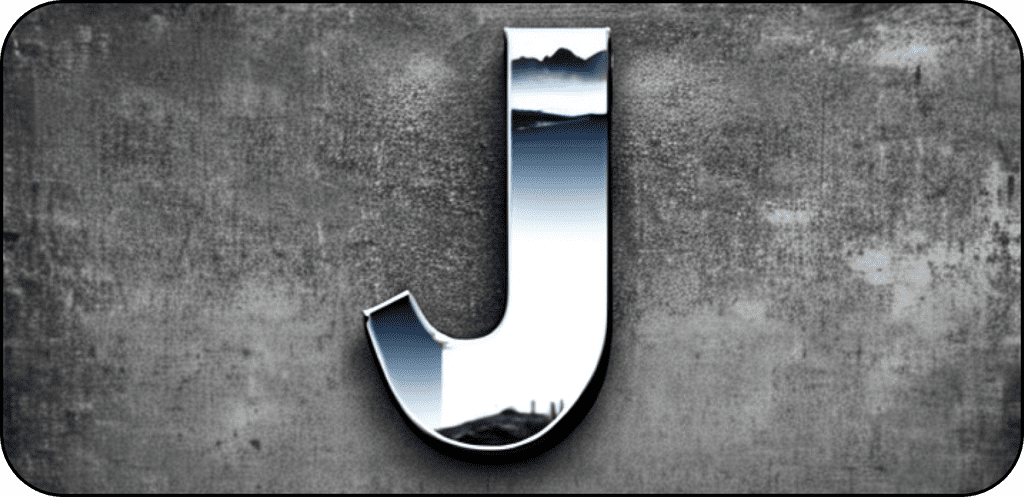} & 
The image presents a large three-dimensional letter ``J", made of polished metal with a smooth surface and a strong mirror reflection.  The interior of the letter reflects a blurry cityscape or mountain range, primarily in dark blue and gray-white tones. The letter is placed against a rough gray cement wall, creating a sharp contrast that enhances the metallic texture and three-dimensionality. & 
\centering 
\includegraphics[width=\linewidth, height=3cm, keepaspectratio]{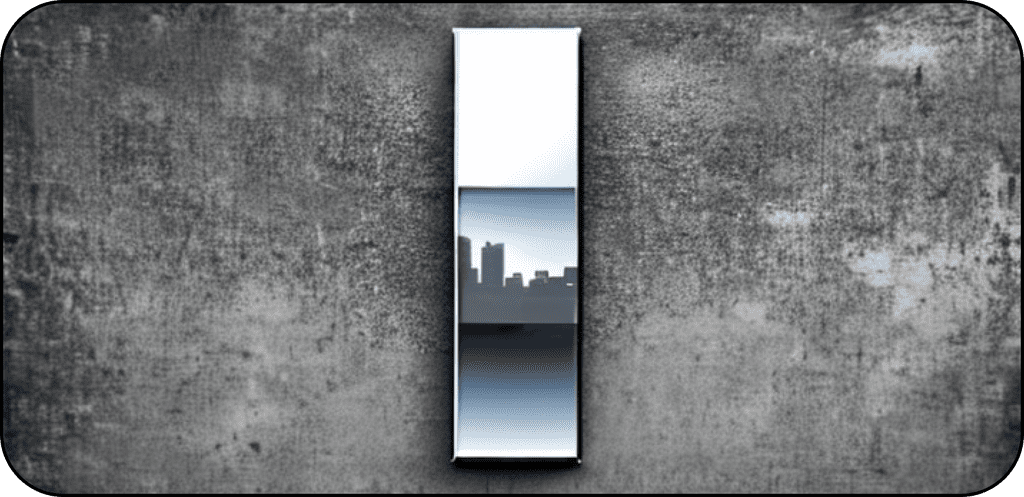} & 
The image is dominated by a large three-dimensional letter ``I", made of polished metal with a mirror effect, reflecting the surrounding light. Inside the letter is a clear black and white silhouette of a city, with densely packed buildings. The bottom is dark blue, contrasting with the bright white above. The background is a rough gray cement wall, creating visual tension with the metallic letter. \\

\midrule
\centering
\includegraphics[width=\linewidth, height=3cm, keepaspectratio]{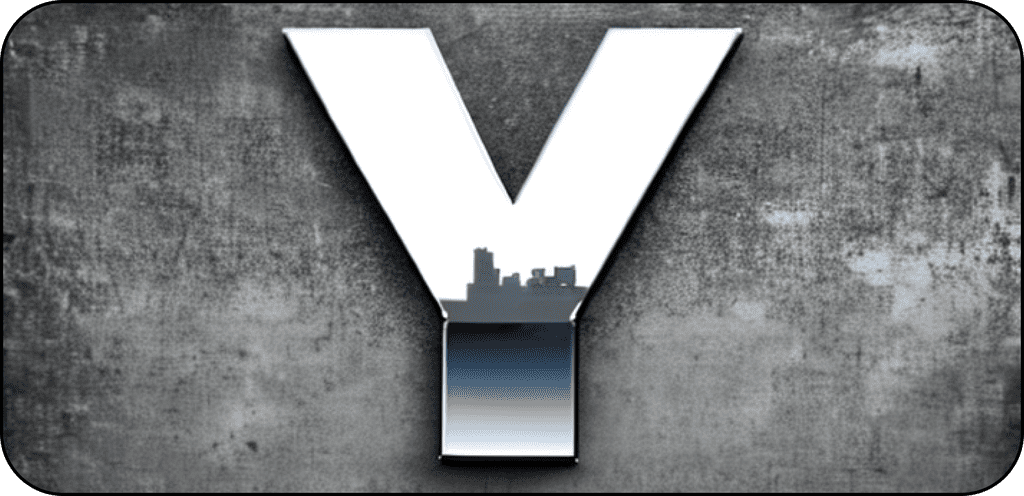} & 
The image shows a large three-dimensional letter ``Y", also made of polished metal with a mirror reflection.  The letter reflects a relatively clear city outline; the buildings are arranged irregularly, and the overall color is dark. The bottom of the letter transitions to dark blue, contrasting with the gray cement wall in the background, creating an industrial and modern feel. The letter is three-dimensional and has a strong texture. & 

\centering
\includegraphics[width=\linewidth, height=3cm, keepaspectratio]{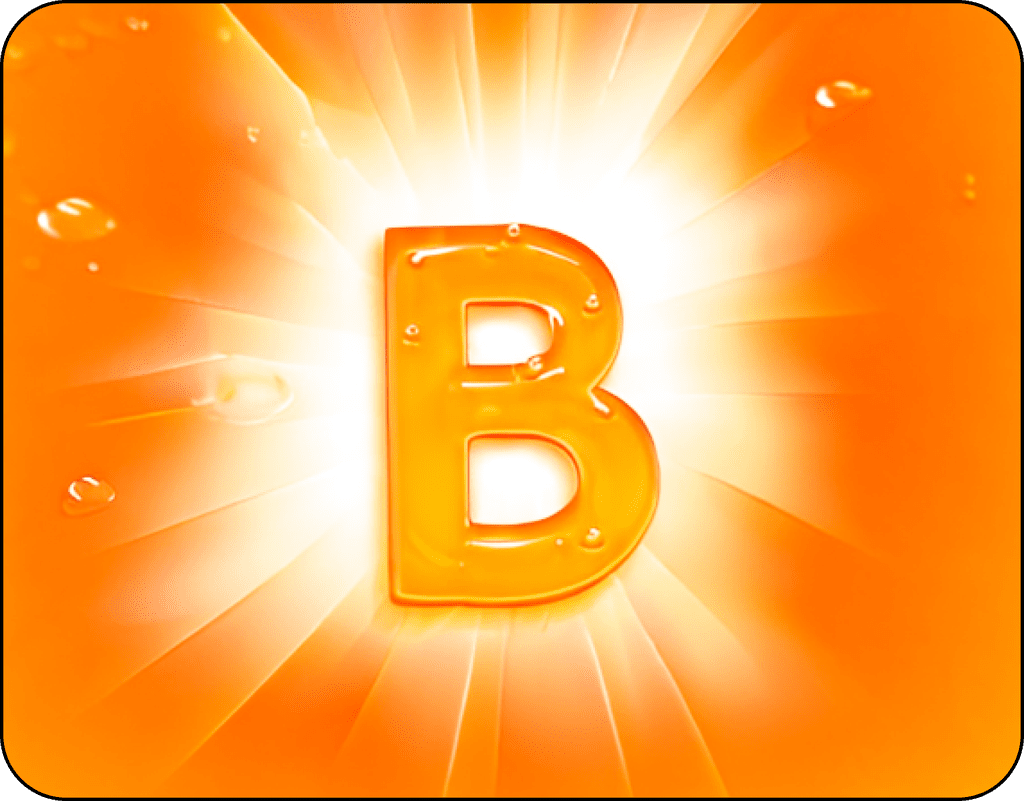} & 
The main subject of the image is a transparent orange letter ``B", with a fine water droplet-like texture on its surface, giving it a crystal-clear texture. The letter ``B" is located in the center of the image, surrounded by bright orange rays of light radiating outwards, creating a dazzling light effect.  Small orange droplets are scattered in the background, echoing the texture on the surface of the letter. The overall tone is warm, bright, and full of vitality.
\\

\midrule
\centering
\includegraphics[width=\linewidth, height=3cm, keepaspectratio]{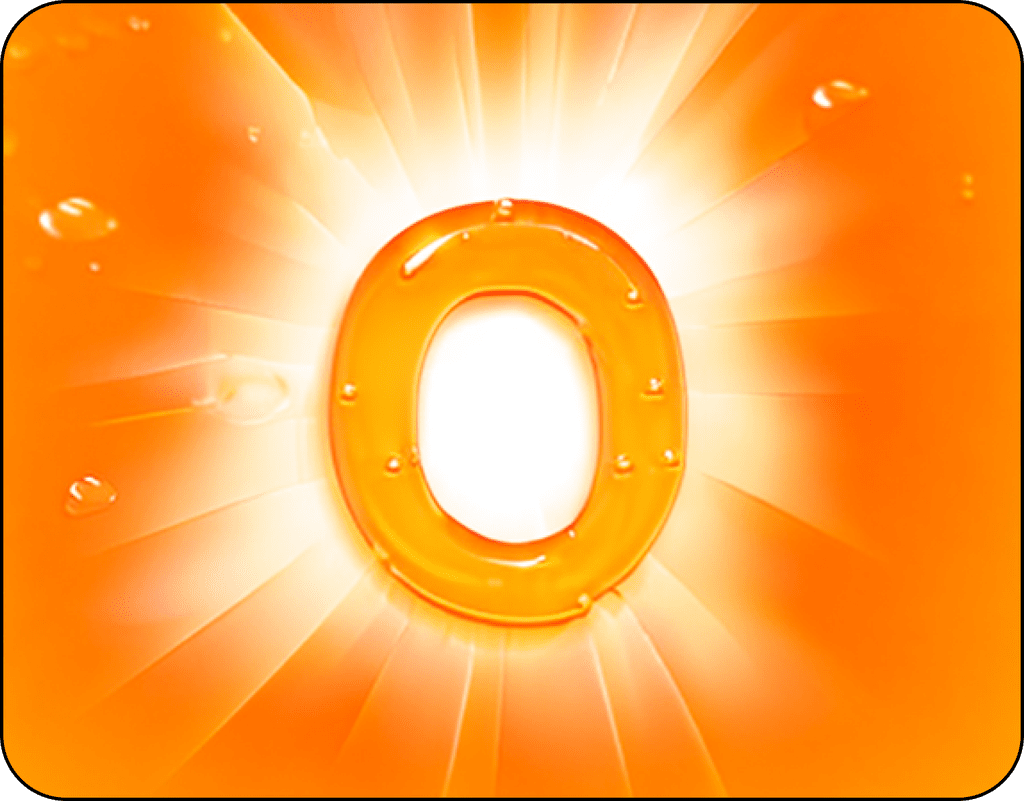} & 
The main subject of the image is a transparent orange letter ``O", with a fine water droplet-like texture on its surface, giving it a crystal-clear texture. The letter ``O" is located in the center of the image, surrounded by bright orange rays of light radiating outwards, creating a dazzling light effect. Small orange droplets are scattered in the background, echoing the texture on the surface of the number. The overall tone is warm, bright, and full of vitality.
& 

\centering
\includegraphics[width=0.6\linewidth, height=3cm, keepaspectratio]{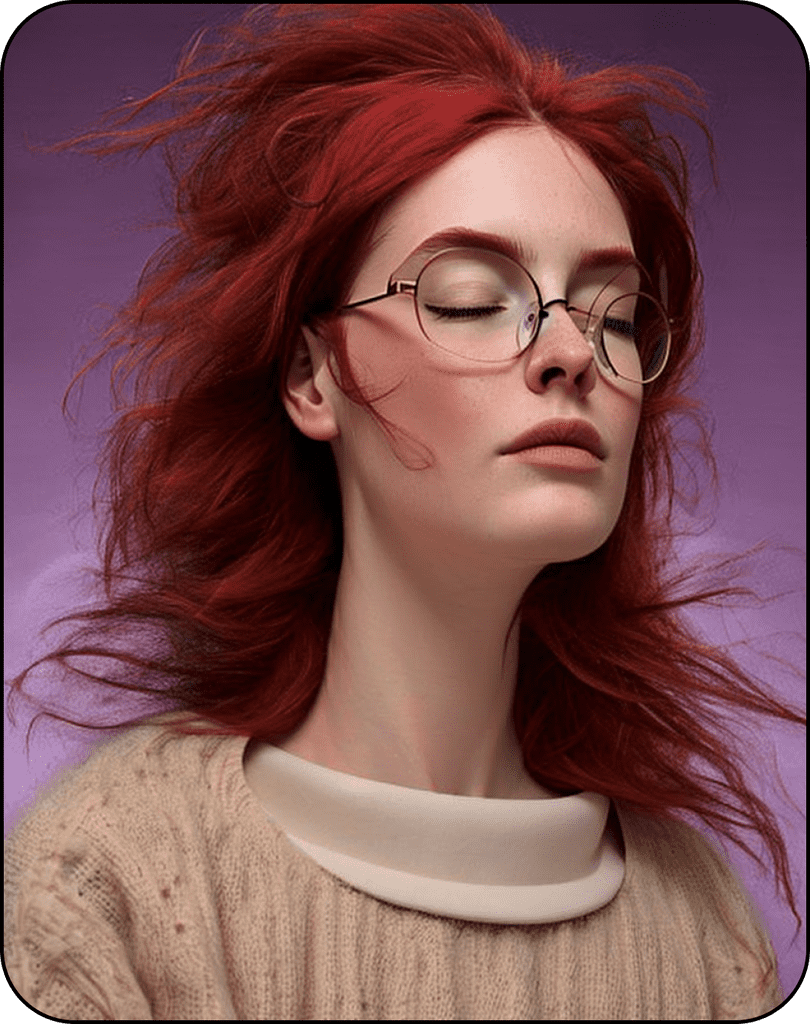} & 
A young woman with long red hair is shown with her eyes closed, wearing a beige sweater with a white collar underneath and round glasses. The background features a purple backdrop, and her hair is blowing in the wind.
\\

\midrule
\centering
\includegraphics[width=0.6\linewidth, height=3cm, keepaspectratio]{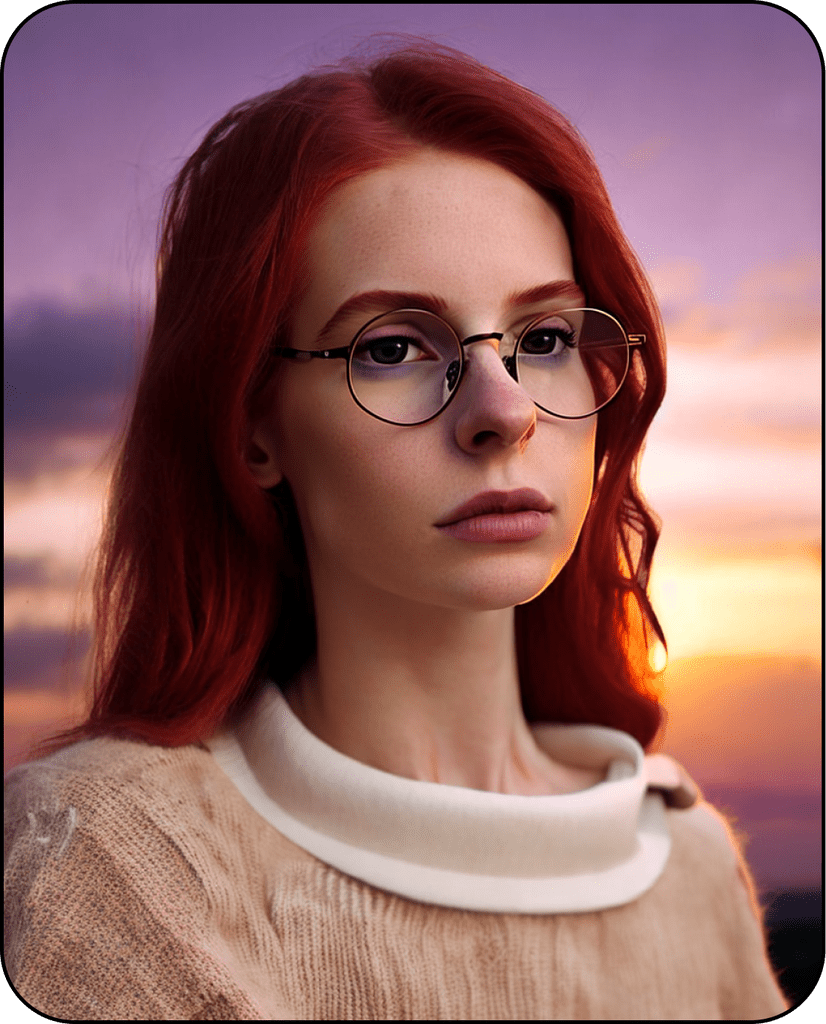} & 
A young woman with long red hair is shown wearing a beige sweater with a white collar underneath and round glasses. She turns her head to look at the camera, with a sunset and a purple backdrop in the background.
& 

\centering
\includegraphics[width=0.6\linewidth, height=3cm, keepaspectratio]{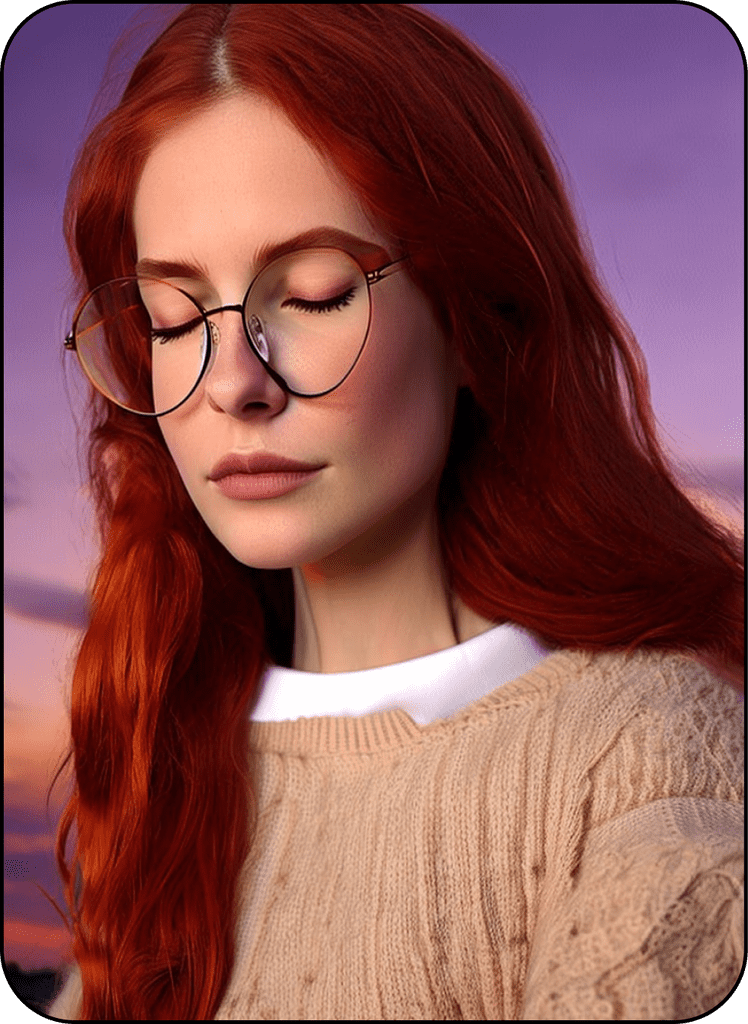} & 
A young woman with long red hair is shown with her eyes closed, wearing a beige sweater with a white collar underneath and round glasses. The background features a sunset and a purple backdrop.
\\

\midrule
\centering
\includegraphics[width=0.6\linewidth, height=3cm, keepaspectratio]{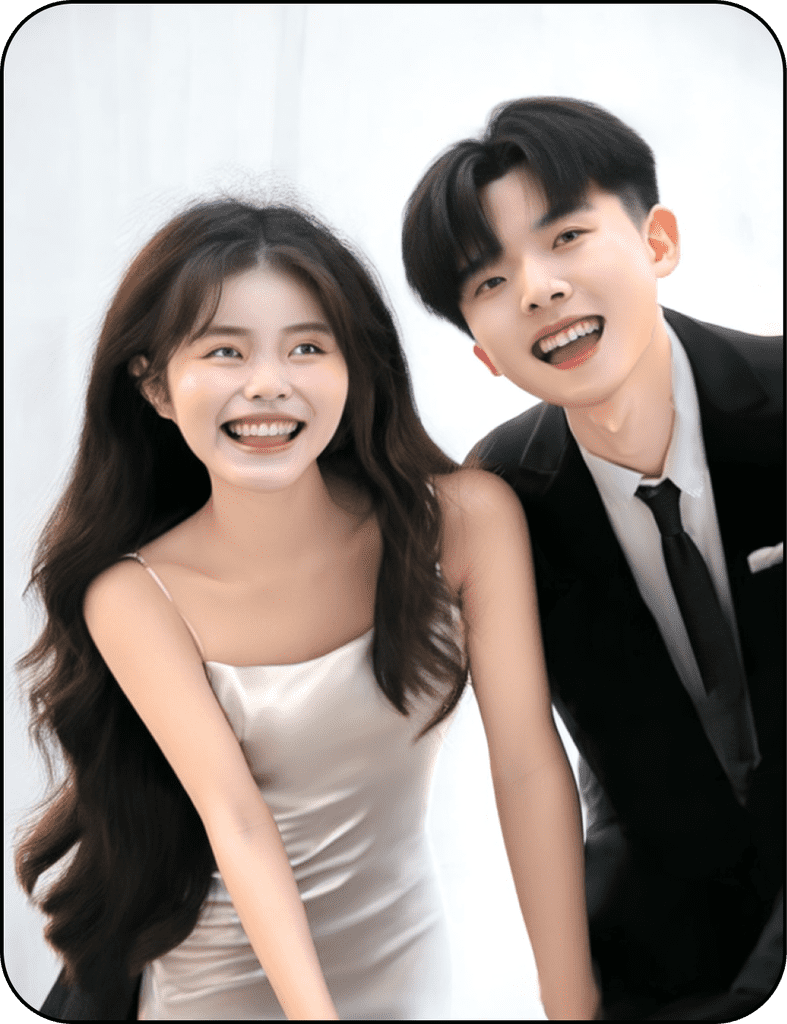} & 
In the photo, a young couple is all smiles. The girl is wearing a white slip dress and the boy is wearing a black suit with a white shirt and black tie. They are holding hands and look very sweet and happy.
& 

\centering
\includegraphics[width=0.6\linewidth, height=3cm, keepaspectratio]{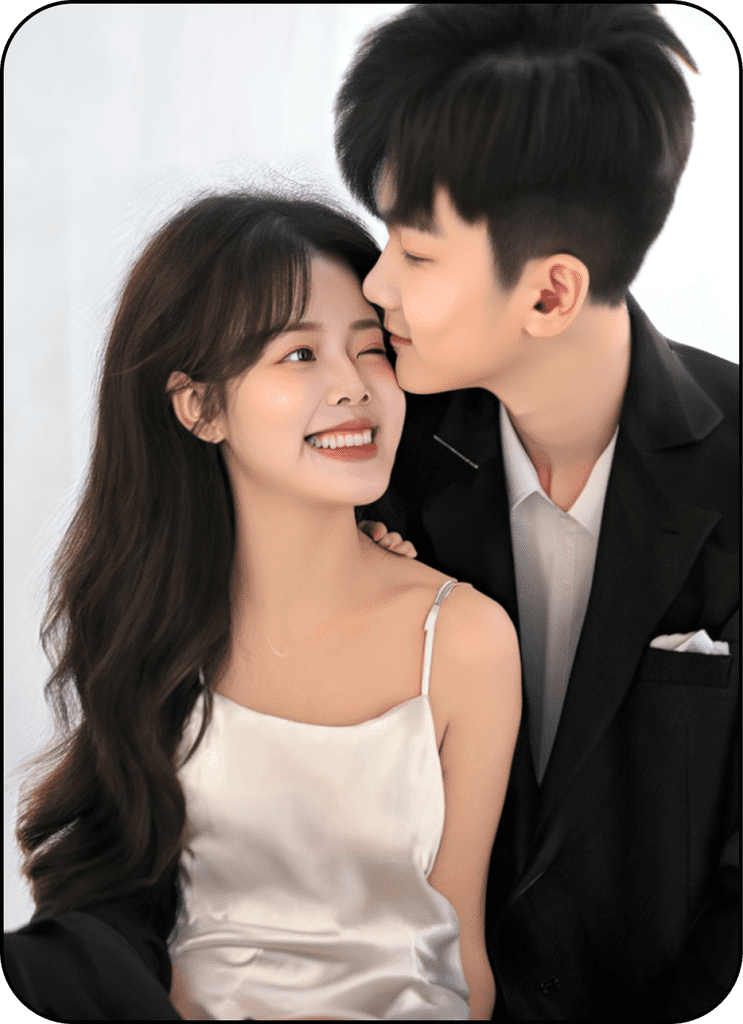} & 
In the photo, a young couple cuddling intimately. The girl is wearing a white slip dress and the boy is wearing a black suit. They are looking into each other’s eyes and the girl has a sweet smile on her face, overflowing with happiness.
\\

\midrule
\centering
\includegraphics[width=0.6\linewidth, height=3cm, keepaspectratio]{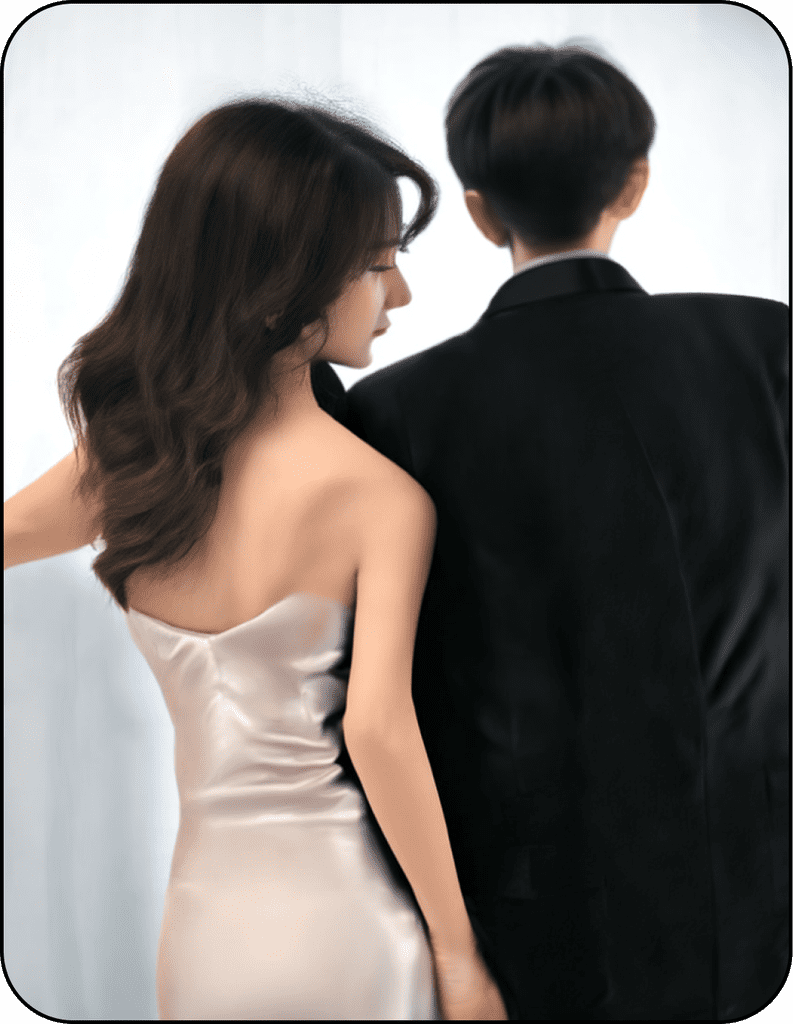} & 
In the photo, a young couple is in an intimate pose. The girl is wearing a white slip dress with her back to the camera, revealing her bare back. The boy is wearing a black suits and looking at the girl’s back affectionately. 
& 

\centering
\includegraphics[width=0.6\linewidth, height=3cm, keepaspectratio]{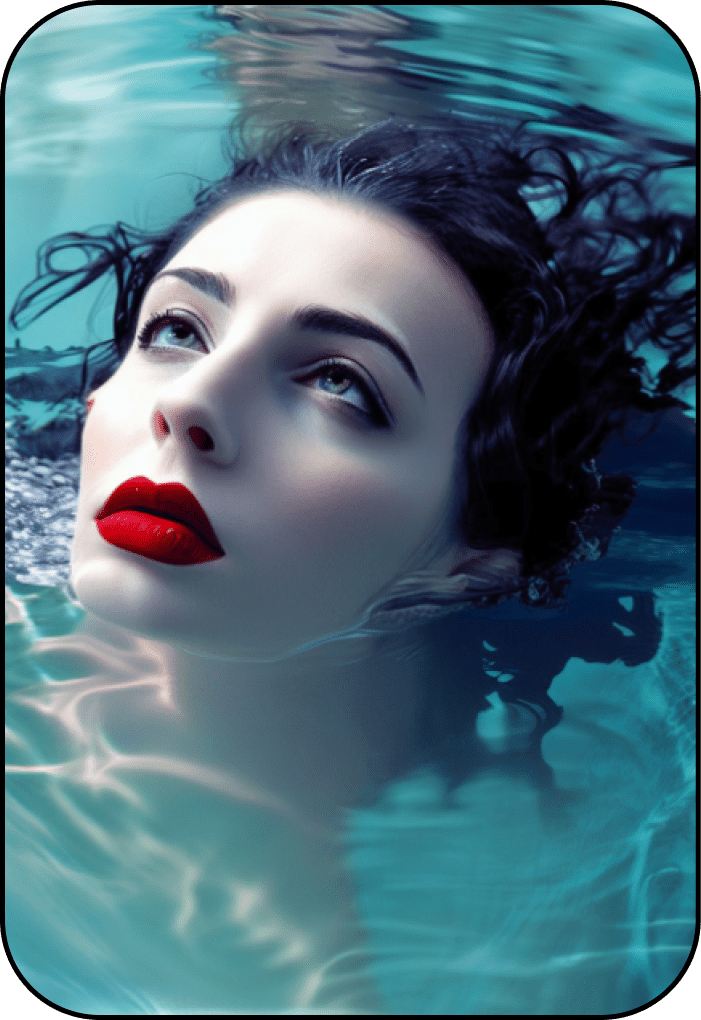} & 
A young woman floats in the water, her head slightly lifted and her gaze directed upwards. Her dark hair spreads around her, contrasting sharply with her fair skin and vibrant red lipstick. The water surrounds her face and hair, creating a dreamlike atmosphere.
\\

\midrule
\centering
\includegraphics[width=0.6\linewidth, height=3cm, keepaspectratio]{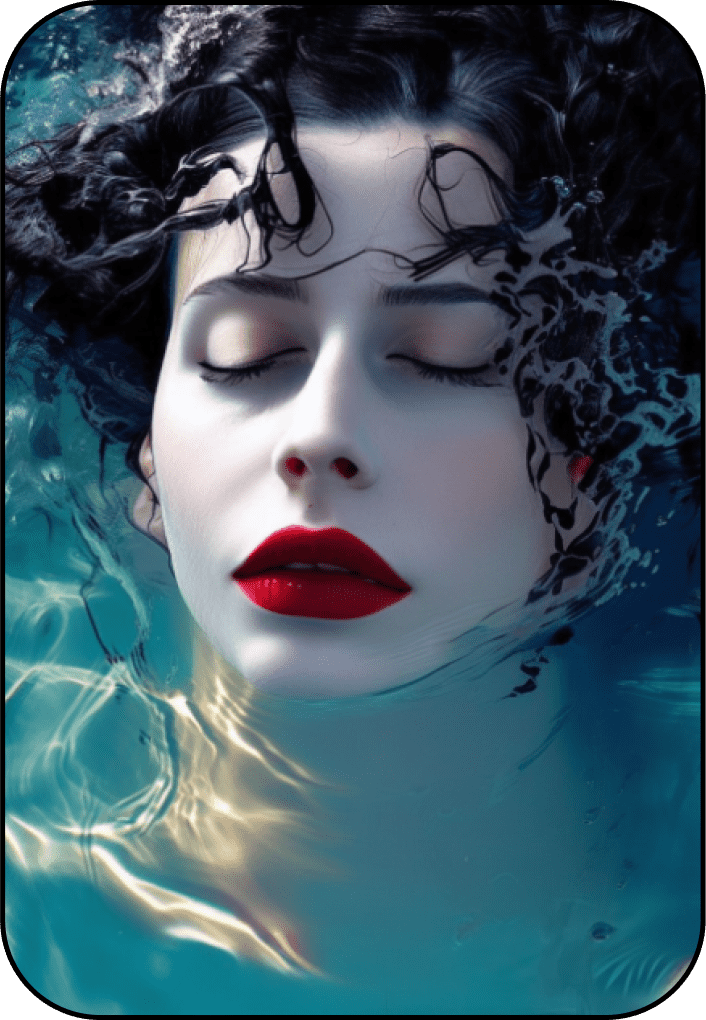} & 
A young woman floats in the water with her eyes closed. Her dark hair spreads around her, contrasting with her fair skin and vibrant red lipstick. The water surrounds her face, creating a peaceful and serene atmosphere.
& 

\centering
\includegraphics[width=0.6\linewidth, height=3cm, keepaspectratio]{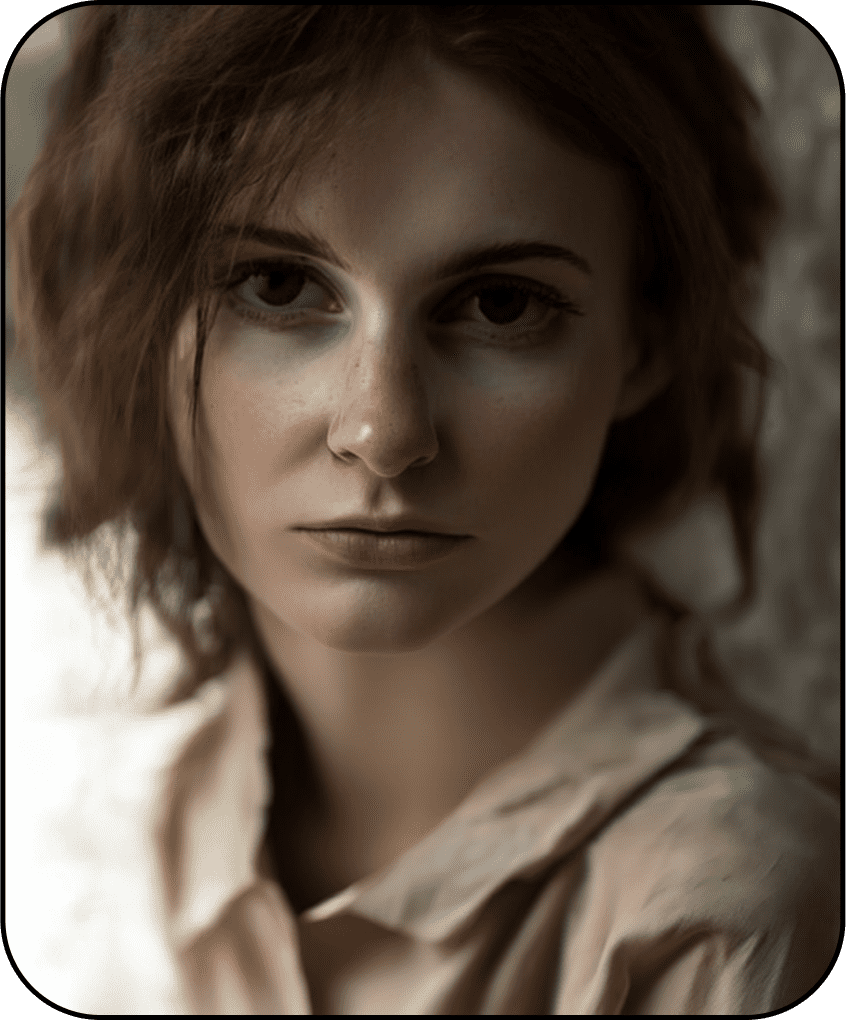} & 
A young woman leans on a chest, gazing directly at the viewer. She has fair skin, her brown hair falls loosely around her shoulders, and her dark eyes hold a gentle expression. She wears a light-colored shirt, the neckline slightly open, revealing the graceful lines of her neck.
\\

\midrule
\centering
\includegraphics[width=0.6\linewidth, height=3cm, keepaspectratio]{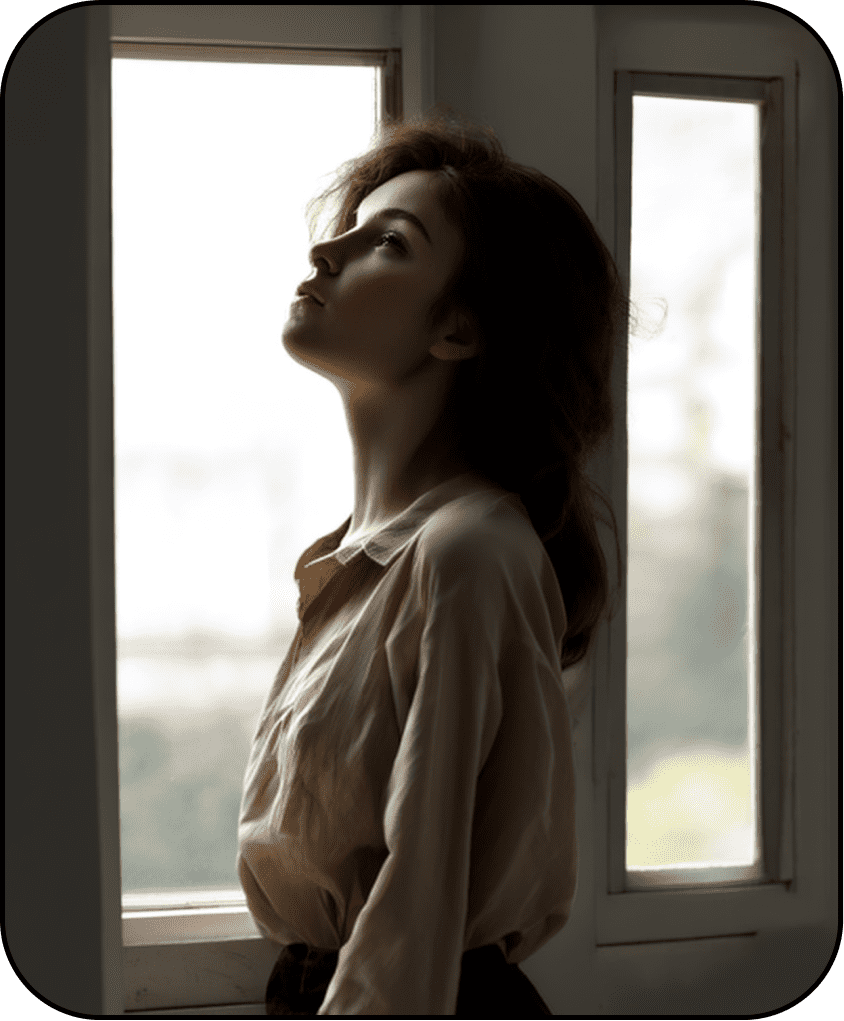} & 
A young woman stands by the window, her head tilted slightly upwards as if feeling the caress of a gentle breeze. She has fair skin, her brown hair falls naturally, and her dark eyes are filled with hope for the future. She wears a light-colored shirt with loose, comfortable sleeves and dark pants, creating a simple and elegant look.
& 

\centering
\includegraphics[width=0.6\linewidth, height=3cm, keepaspectratio]{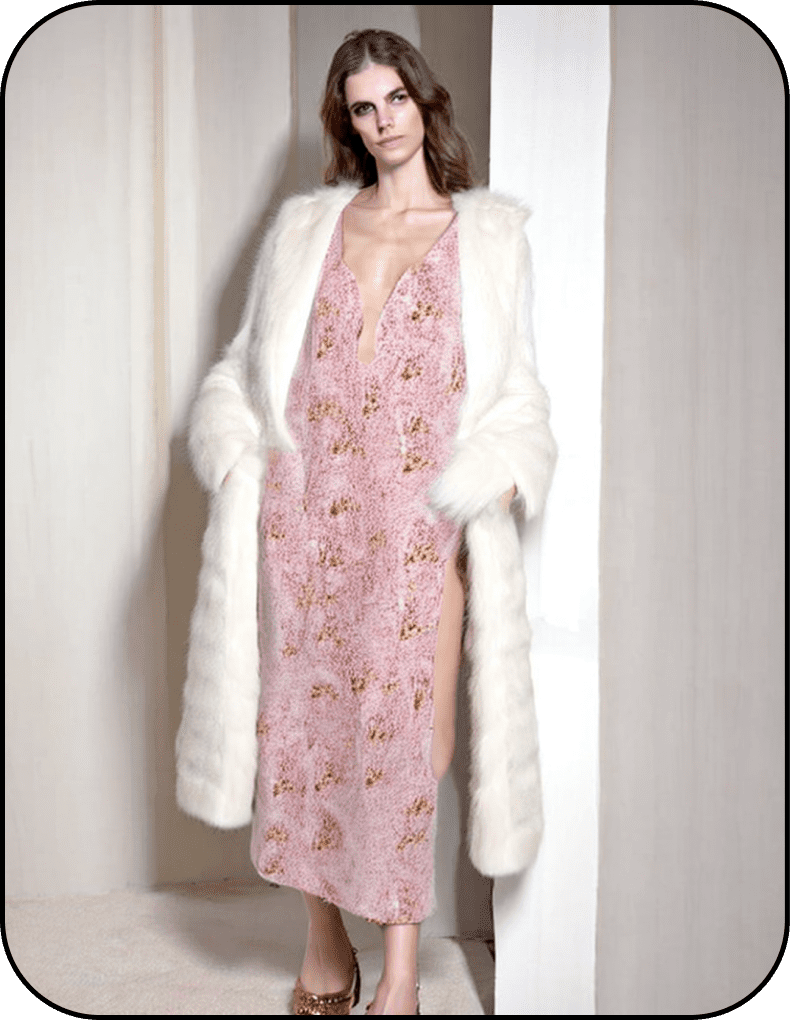} & 
A fashion photography piece showcasing a female model in a pink printed maxi dress layered over with a long white furry coat. The model has long brown hair and sophisticated makeup; she stands against a light-colored backdrop in an elegant pose with her hands in her pockets. The dress has a V-neck design, is lightweight in texture, and features a soft print. The coat is fluffy and soft, with a high-end feel. The model is wearing gold strapped heels.
\\

\midrule
\centering
\includegraphics[width=0.6\linewidth, height=3cm, keepaspectratio]{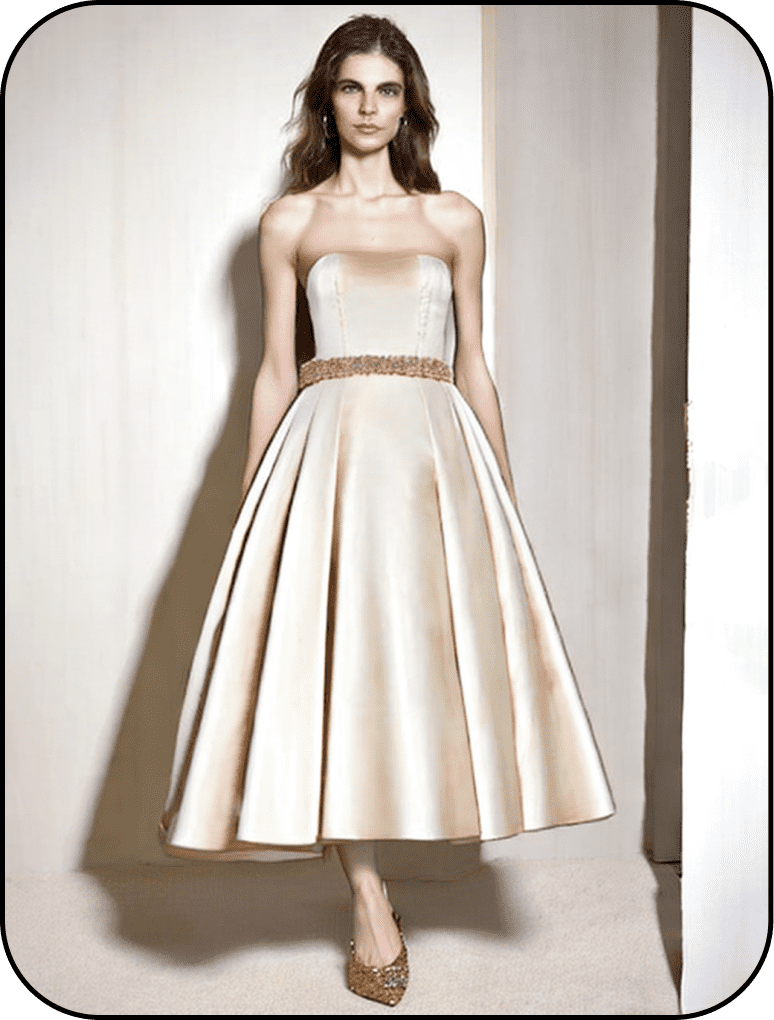} & 
Fashion photography showcasing a female model in a champagne-colored strapless A-line dress. The dress is smooth in texture, with a strong drape, and an A-line skirt, cinched at the waist with a golden belt. The model's hairstyle and makeup are consistent with the previous image, and the background is similarly simple and light-colored. The model’s pose is elegant, with her gaze directed forward. She wears gold heels that complement the dress.
& 

\centering
\includegraphics[width=0.6\linewidth, height=3cm, keepaspectratio]{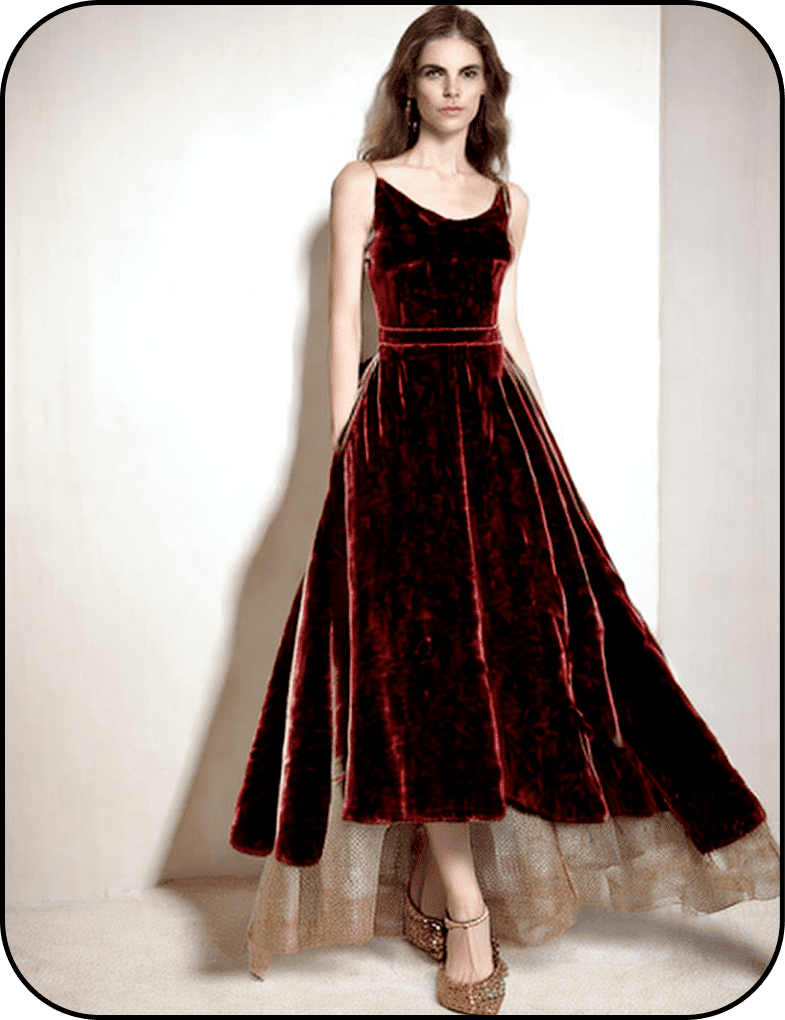} & 
A fashion photography image displaying a female model in a burgundy velvet maxi dress. The dress is a halter neck style, with a defined waist, and an asymmetric hemline, trailing on one side while revealing the ankle on the other. The dress is thick in texture, with a rich color, and the skirt flows slightly.  The model's hairstyle and makeup remain consistent, against a simple light-colored background. The model's posture is elegant, and she wears gold high-heeled shoes.
\\
\midrule

\centering
\includegraphics[width=0.6\linewidth, height=3cm, keepaspectratio]{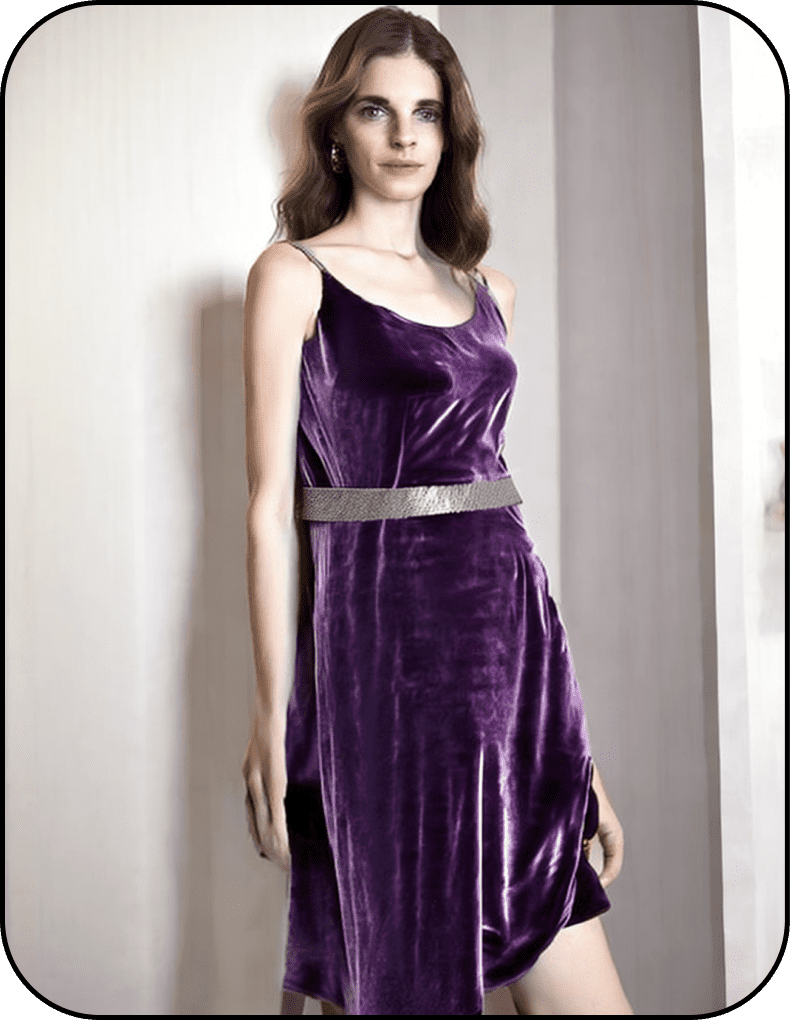} & 
A fashion photography piece featuring a female model in a purplish-red velvet slip dress. The dress is simply cut and form-fitting, accented with a thin silver belt at the waist. The model maintains the consistent hairstyle and makeup from other images, against a light-colored background.  The model poses elegantly, with a confident smile. Her makeup is refined, with simple earrings. The model wears light gold heeled shoes.
& 

\centering
\includegraphics[width=0.8\linewidth, height=3cm, keepaspectratio]{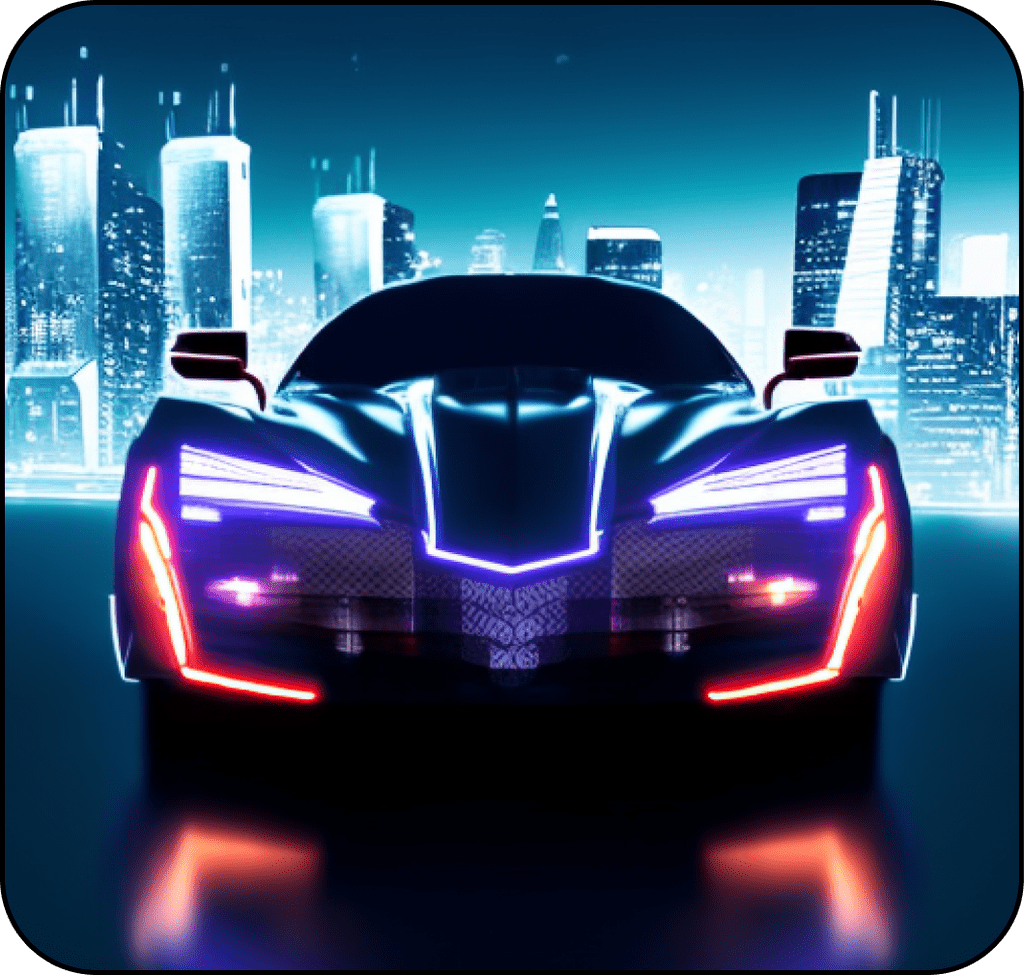} & 
A futuristic cyberpunk car faces the camera, showcasing its front design. The car boasts sleek lines, intricate details, and dazzling lighting effects. The backdrop is a technologically advanced cityscape at night. 
\\
\midrule

\centering
\includegraphics[width=0.8\linewidth, height=3cm, keepaspectratio]{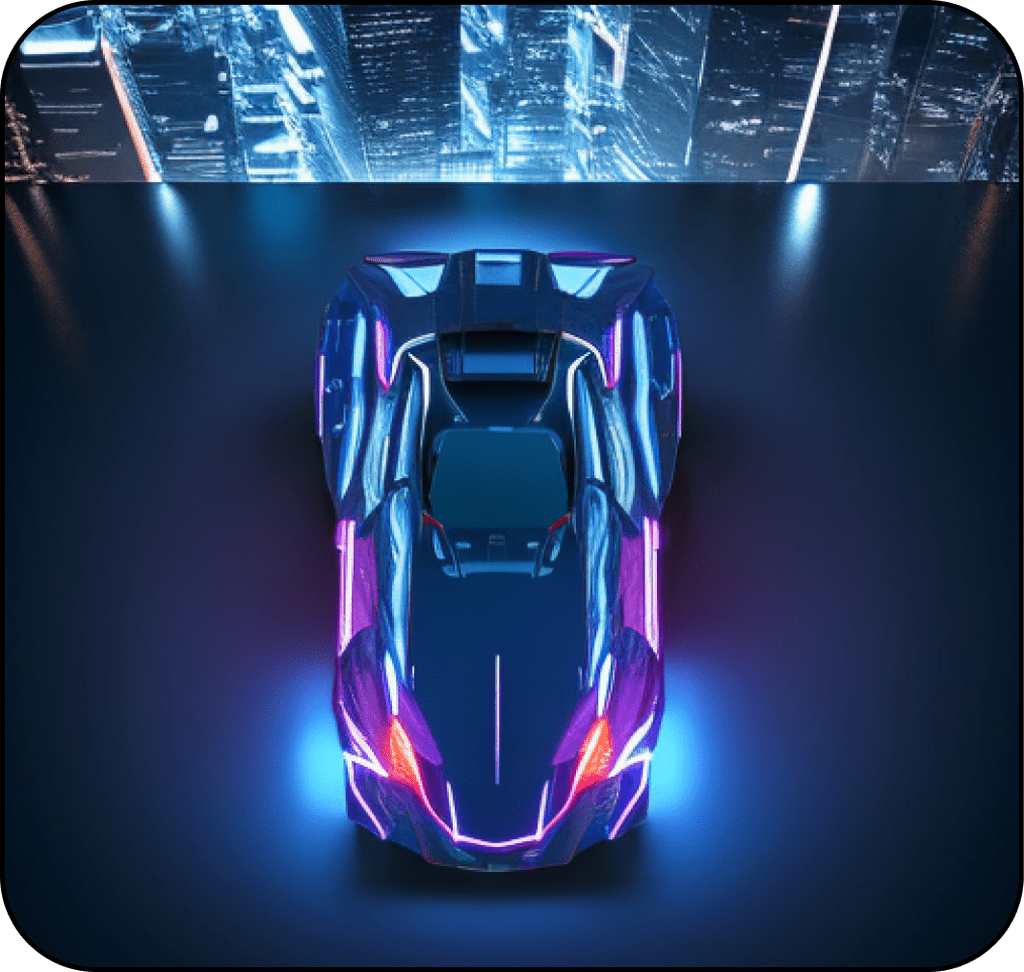} & 
A futuristic cyberpunk car is seen from above, showcasing its roof design and overall layout. The car boasts sleek lines, intricate details, and dazzling lighting effects. The backdrop is a technologically advanced cityscape at night.
& 

\centering
\includegraphics[width=0.8\linewidth, height=3cm, keepaspectratio]{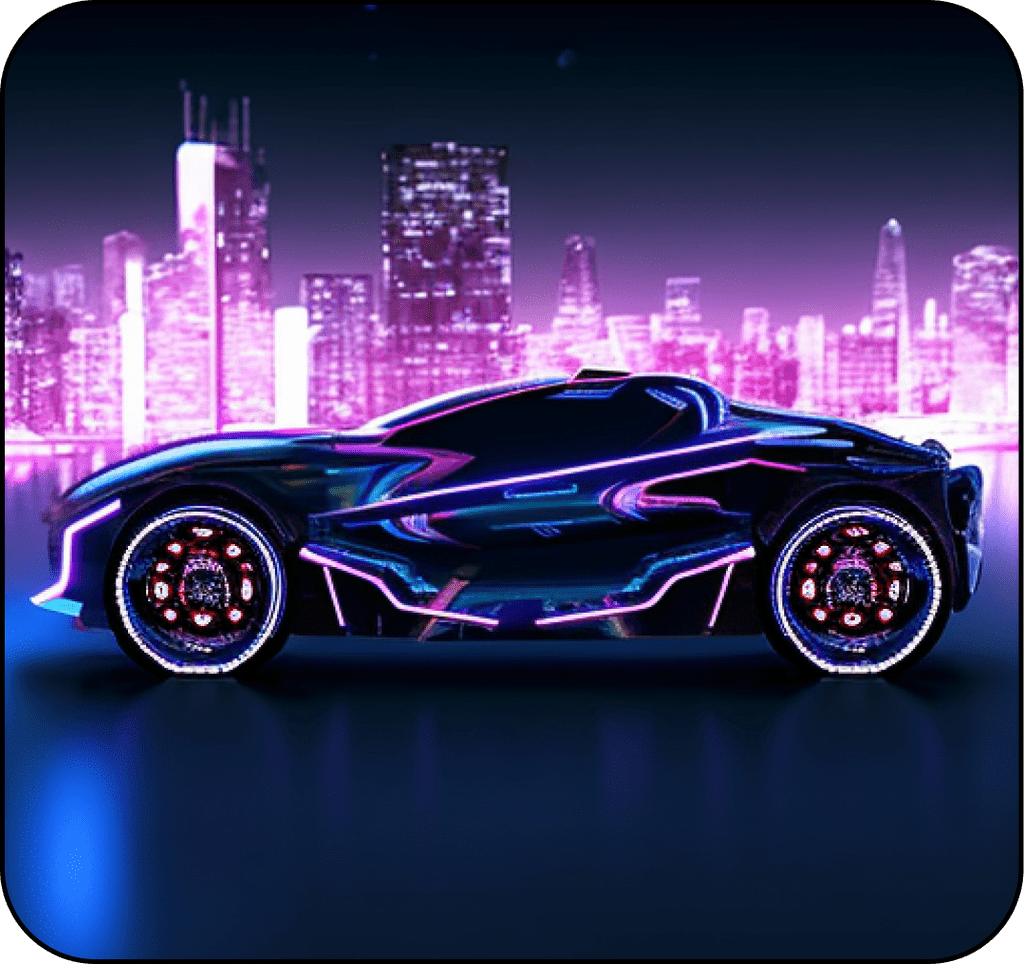} & 
A futuristic cyberpunk car is presented sideways, showcasing its side profile and streamlined body. The car boasts sleek lines, intricate details, and dazzling lighting effects. The backdrop is a technologically advanced cityscape at night.
\\
\midrule

\centering
\includegraphics[width=0.8\linewidth, height=3cm, keepaspectratio]{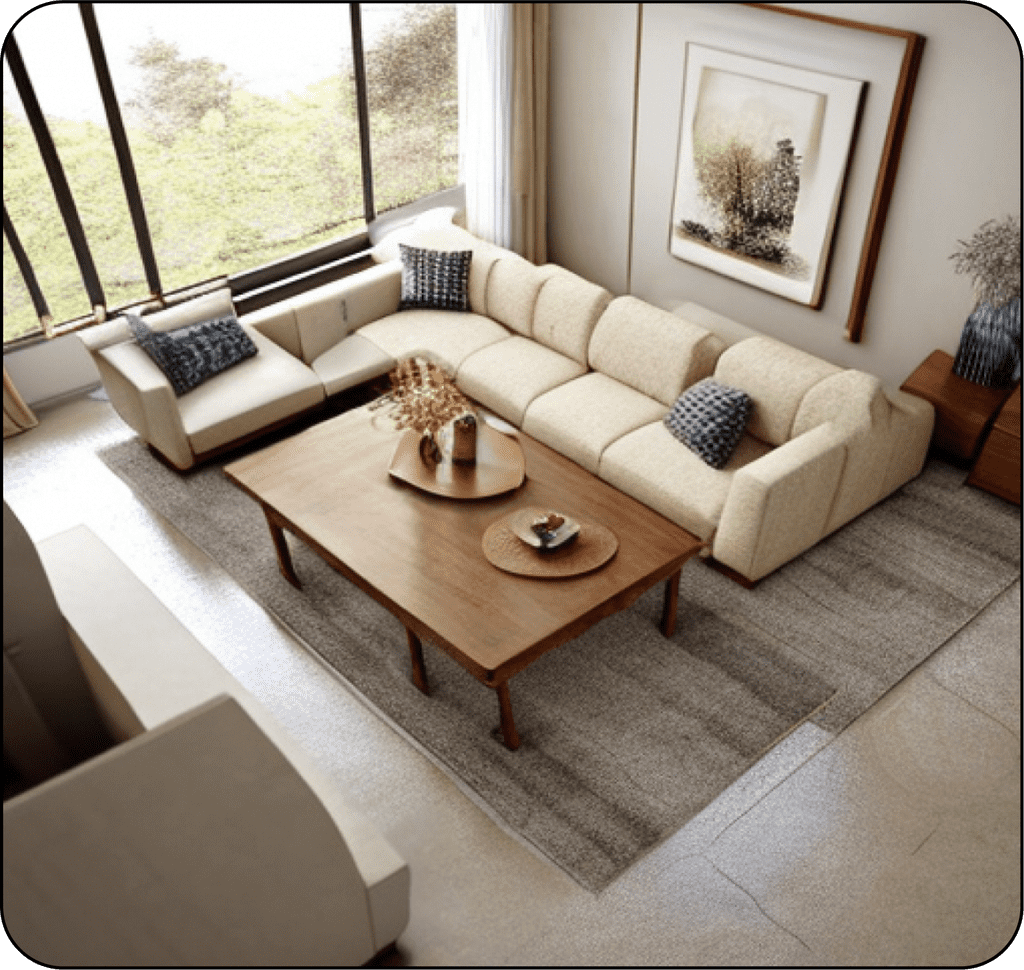} & 
The image shows a modern minimalist living room from an overhead perspective. A beige fabric L-shaped sofa occupies the center of the space, complemented by a dark wood coffee table and a light-colored rug. Decorative items are placed on the coffee table. A floor-to-ceiling window dominates one wall, showcasing lush greenery outside. An abstract artwork hangs on the wall, and a wooden side table is visible in a corner. The overall color scheme is soft, creating a relaxed and comfortable atmosphere.
& 

\centering
\includegraphics[width=0.8\linewidth, height=3cm, keepaspectratio]{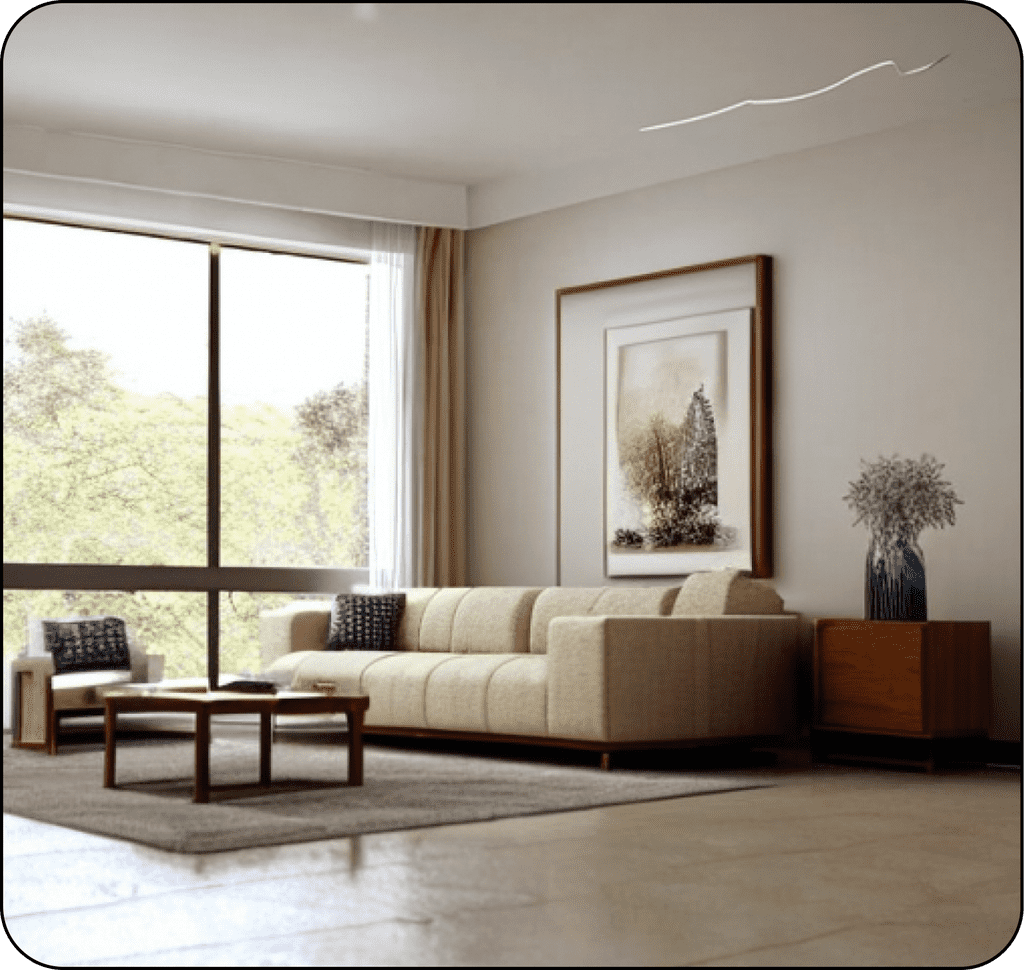} & 
The image depicts a modern minimalist living room. A beige fabric sofa is centrally positioned, featuring smooth lines and soft, comfortable material. A dark wood coffee table with a simple and elegant design sits in front of the sofa. A floor-to-ceiling window occupies one wall, revealing a pleasant outdoor view. An abstract artwork hangs on the wall, next to a wooden side cabinet displaying a blue vase. The overall space is spacious and bright, with a clean and sophisticated design style.
\\
\midrule

\centering
\includegraphics[width=0.8\linewidth, height=3cm, keepaspectratio]{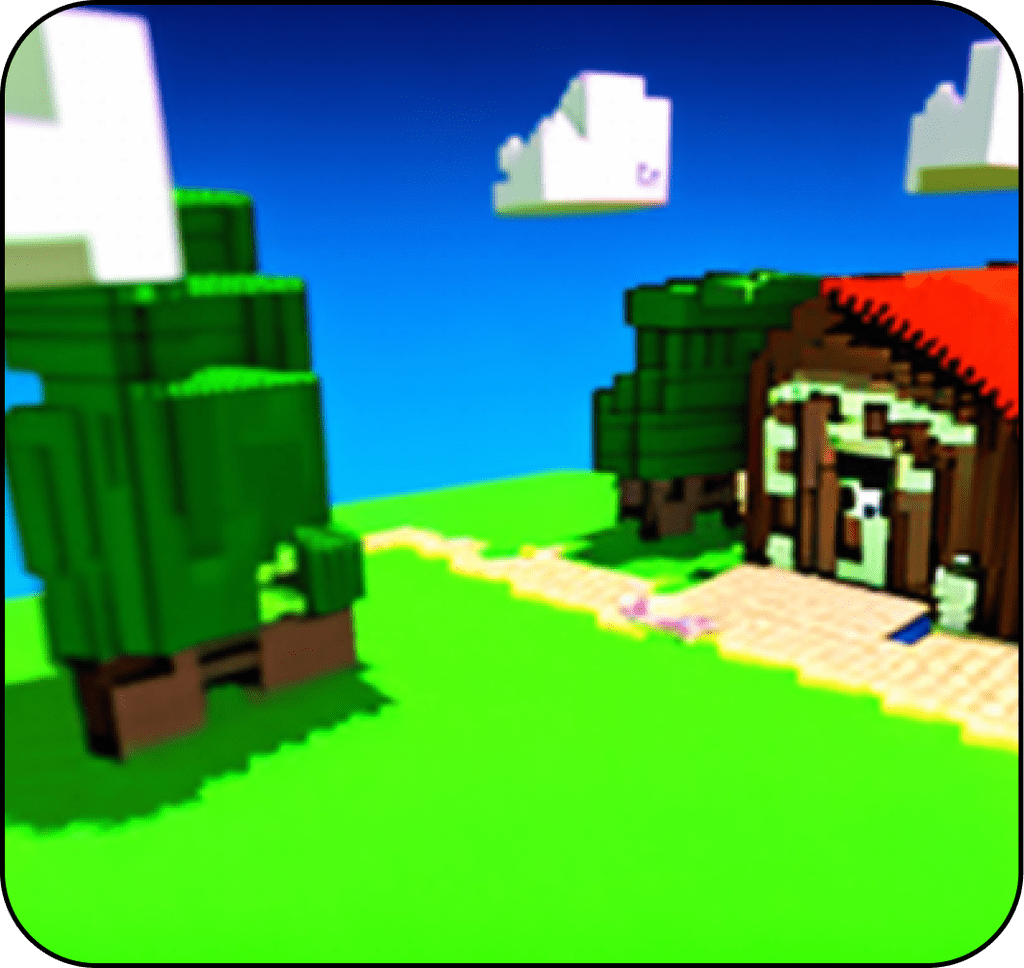} & 
A front view of a Minecraft-style game scene, the picture is full of pixelated style. In the foreground is green grass dotted with colorful flowers. Not far away is a house built of oak, with a path paved with stone bricks in front of the house. Around the house are planted several lush trees with leaves in different shades of green. Several white clouds float in the sky.
& 

\centering
\includegraphics[width=0.8\linewidth, height=3cm, keepaspectratio]{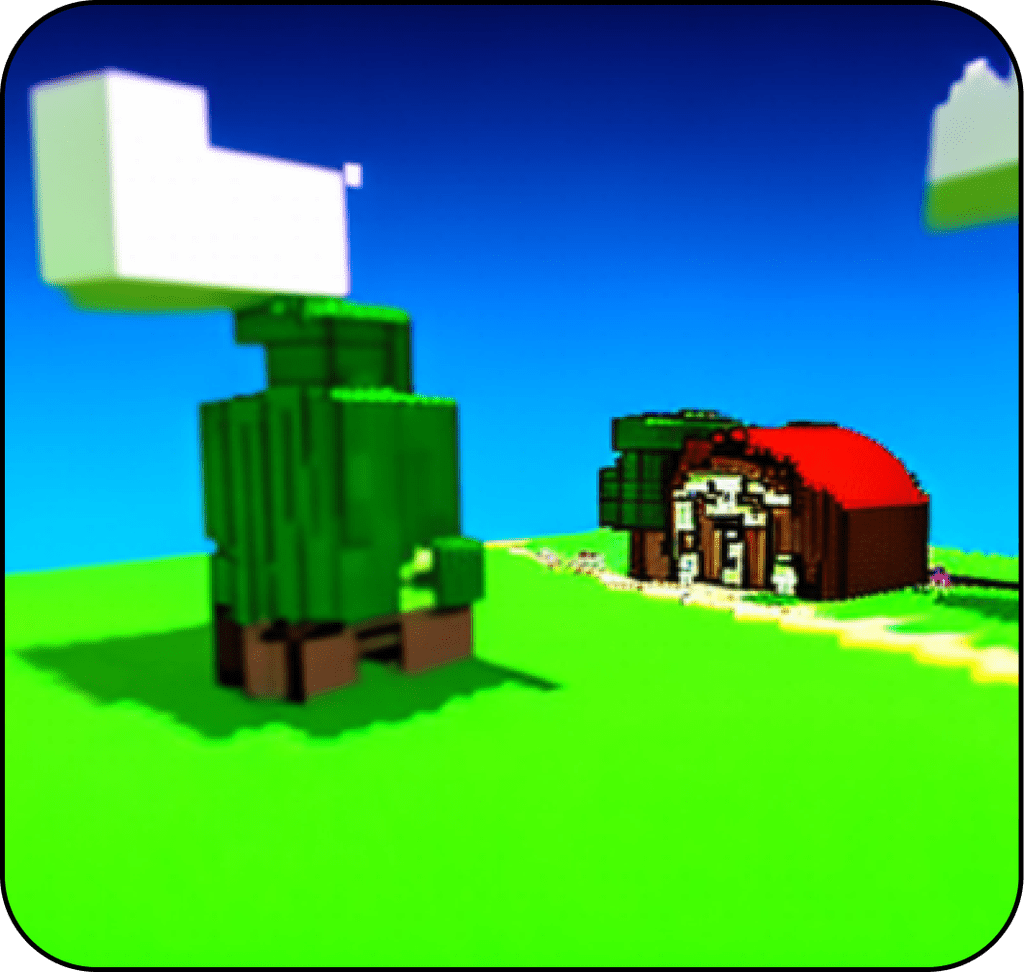} & 
A side view of a Minecraft-style game scene, the picture is full of pixelated style. In the foreground is green grass dotted with colorful flowers. The grass stretches into the distance, where it meets the blue sky. On the left side of the picture, you can see a house built of oak, with a path paved with stone bricks in front of the house. Several lush trees are planted next to the house, with leaves in different shades of green. Several white clouds float in the sky.
\\
\midrule

\centering
\includegraphics[width=0.8\linewidth, height=3cm, keepaspectratio]{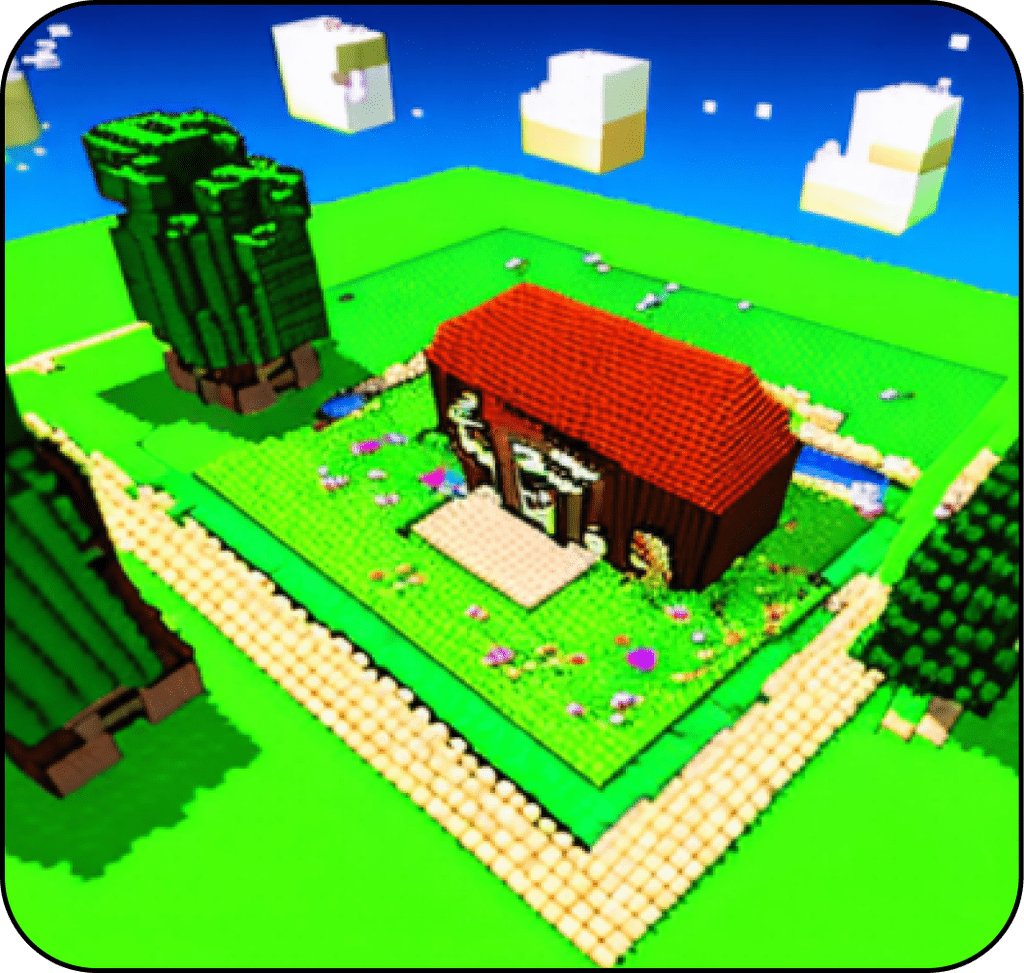} & 
A top view of a Minecraft-style game scene, the picture is full of pixelated style. In the center of the picture is a green grass with colorful flowers dotted on it. There is a house built of oak on the grass, with a path paved with stone bricks in front of the house. Several lush trees planted around are the house, with leaves in different shades of green. A winding river can be seen beside the house. Several white clouds float in the sky.
& 

\centering
\includegraphics[width=0.8\linewidth, height=3cm, keepaspectratio]{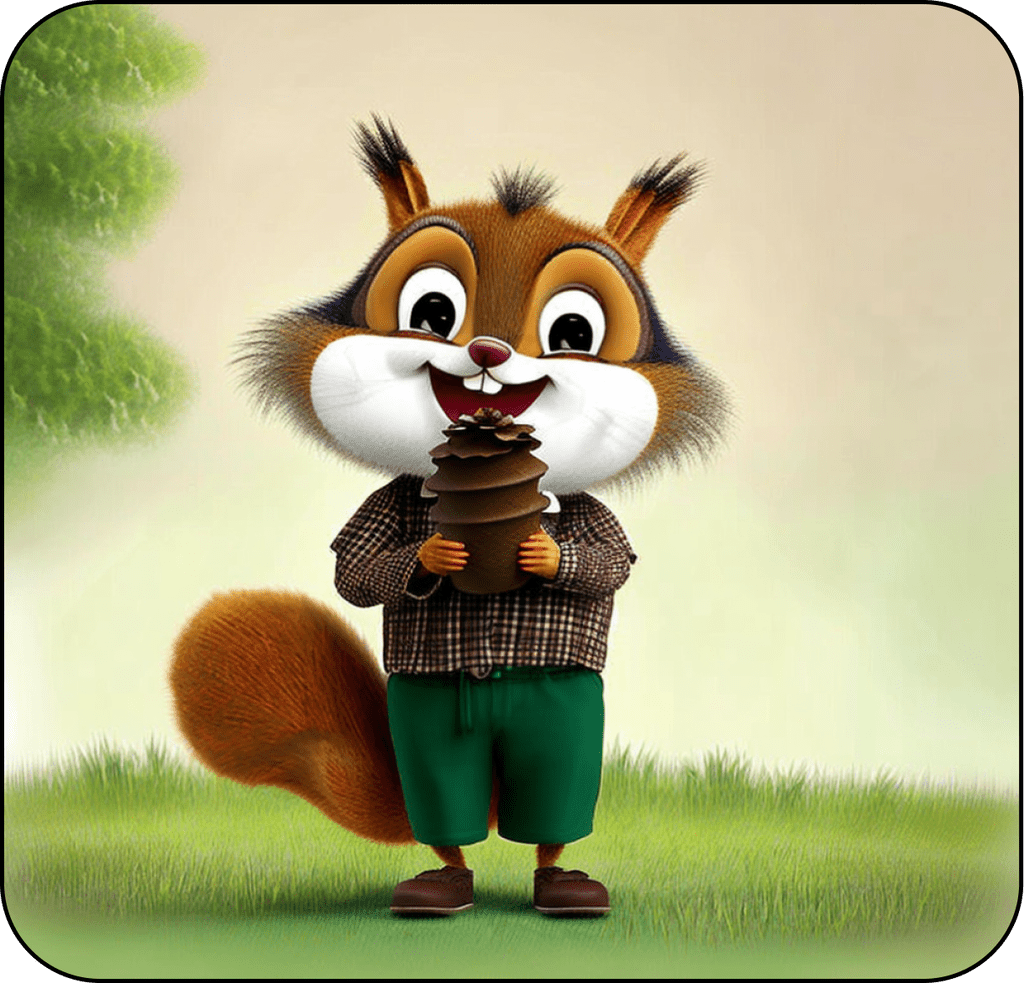} & 
A fluffy squirrel, dressed in a brown plaid shirt and green pants, stands on the grass. It has a wide grin, wide-open eyes, and holds a giant pine cone, looking delighted. 
\\
\midrule

\centering
\includegraphics[width=0.8\linewidth, height=3cm, keepaspectratio]{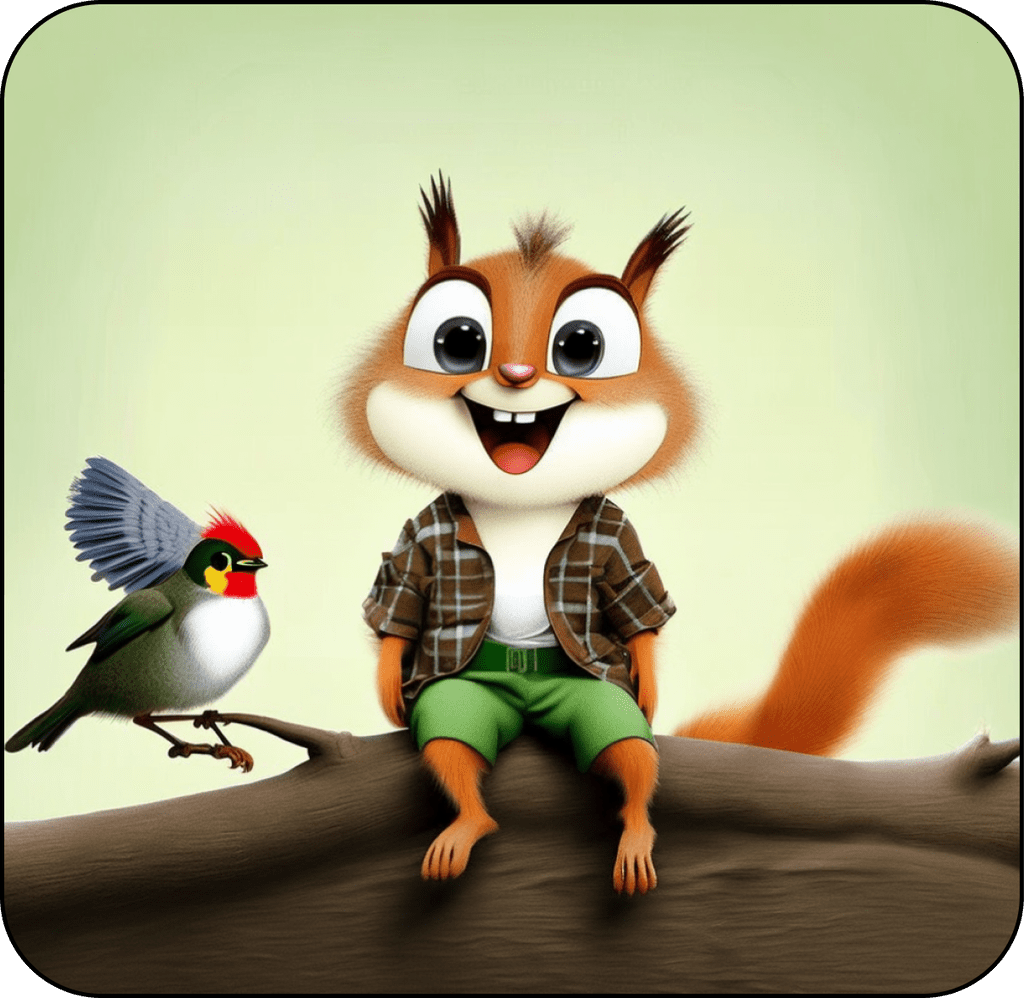} & 
A fluffy squirrel, dressed in a brown plaid shirt and green pants, sits on a branch. It has a wide grin, wide-open eyes, and looks happy. A small bird sits on the branch next to it, and another bird sits on a higher branch. 
& 

\centering
\includegraphics[width=0.6\linewidth, height=3cm, keepaspectratio]{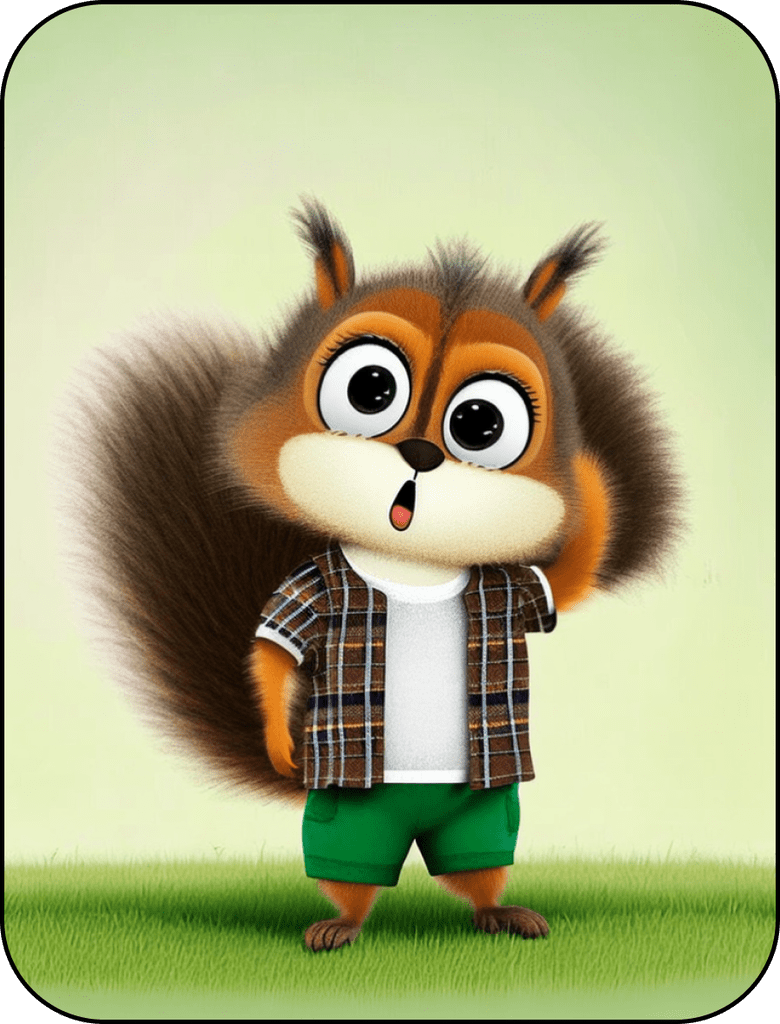} & 
A fluffy squirrel, dressed in a brown plaid shirt and green pants, stands on the grass. It has wide-open eyes and a slightly open mouth, looking surprised. 
\\
\midrule

\centering
\includegraphics[width=0.6\linewidth, height=3cm, keepaspectratio]{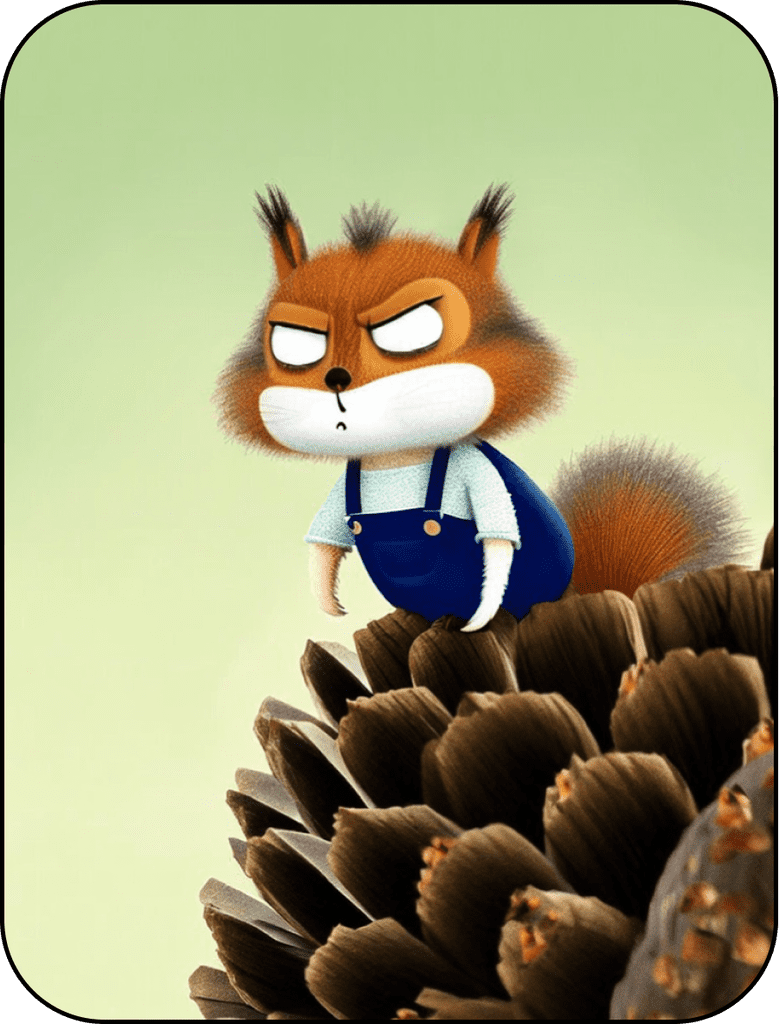} & 
A fluffy squirrel, dressed in blue overalls, lies on top of a giant pine cone. It has a frown, wide-open eyes, and a slightly open mouth, looking angry. 
& 

\centering
\includegraphics[width=0.6\linewidth, height=3cm, keepaspectratio]{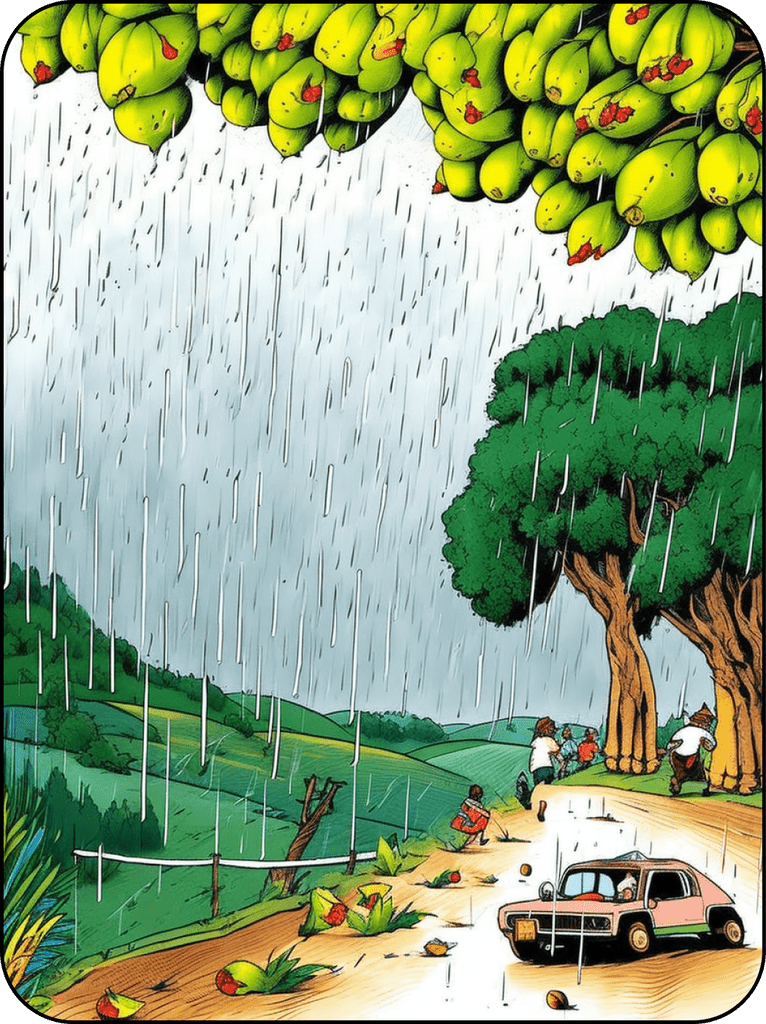} & 
The picture depicts a sudden downpour in the countryside. Large raindrops fall from the sky as people on the hillside run downhill to seek shelter. The strong wind bends the trees and blows fruit to the ground. A car drives through the rain on a country road. 
\\
\midrule

\centering
\includegraphics[width=0.6\linewidth, height=3cm, keepaspectratio]{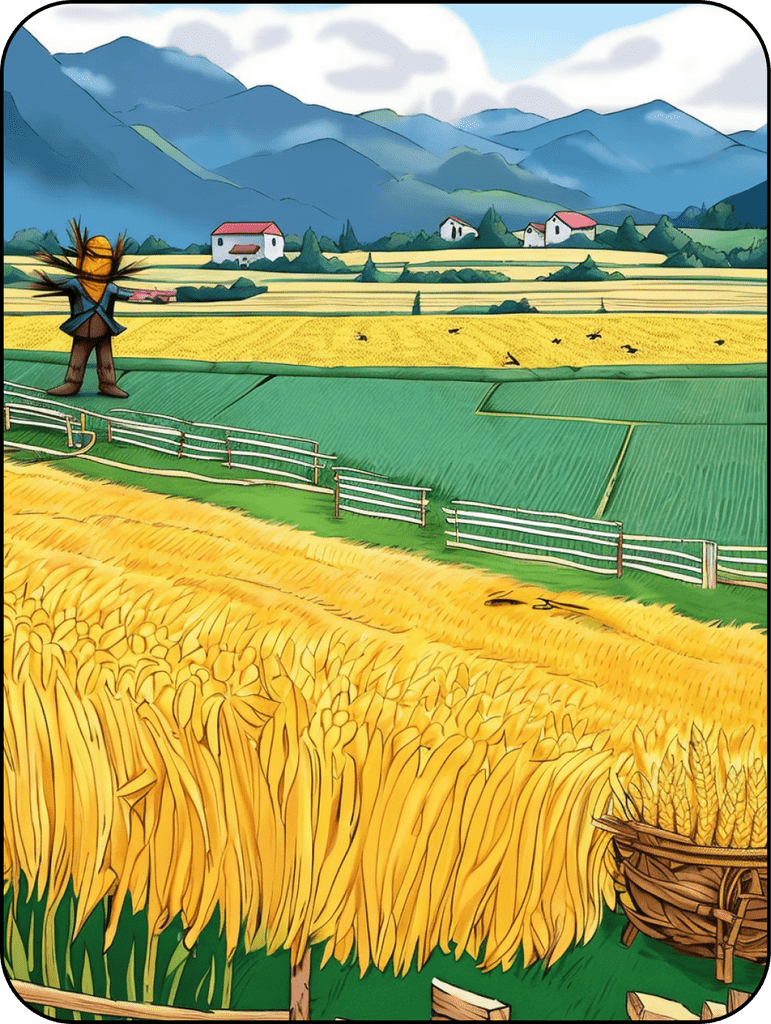} & 
The picture depicts a bountiful harvest scene in the countryside. Golden wheat fields stretch as far as the eye can see, with rolling mountains and a few houses in the distance. A scarecrow stands in the wheat field, guarding the upcoming harvest. A few birds perch on a fence by the field, seemingly admiring the beautiful scenery. 
& 

\centering
\includegraphics[width=0.6\linewidth, height=3cm, keepaspectratio]{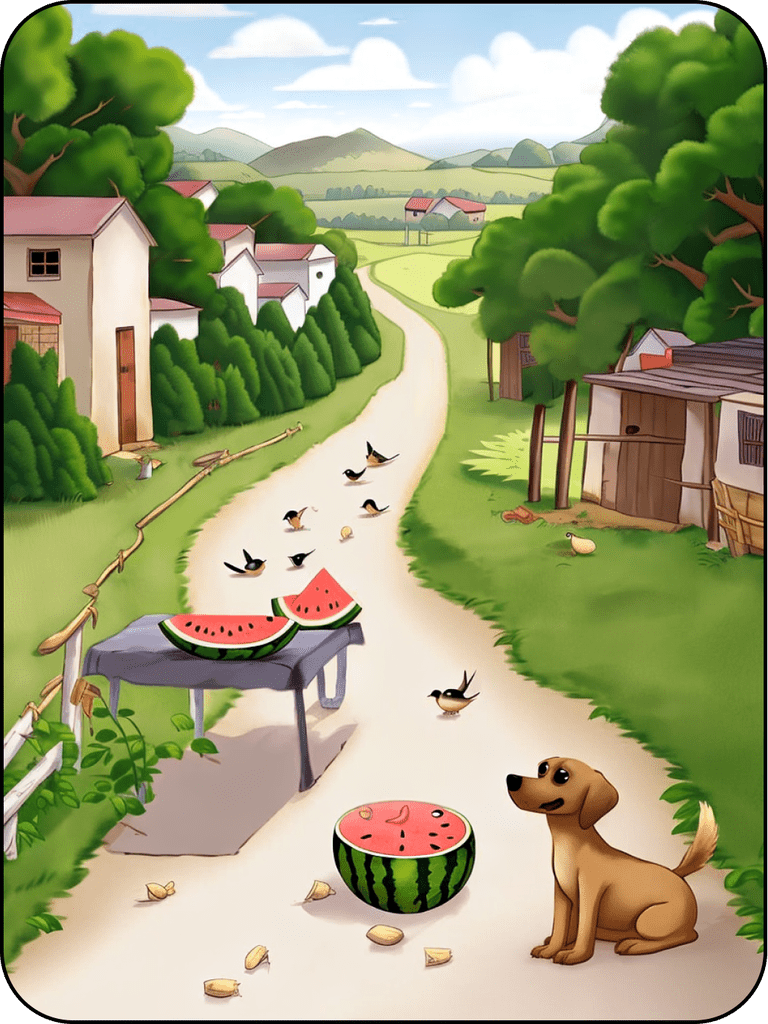} & 
The picture depicts a tranquil rural scene. A country road winds through the picture, flanked by several houses surrounded by lush greenery. On a table by the road lie slices of watermelon with a few birds pecking at the crumbs on the ground. A dog lies under the table, seemingly anticipating the delicious treat. 
\\
\midrule

\centering
\includegraphics[width=0.6\linewidth, height=3cm, keepaspectratio]{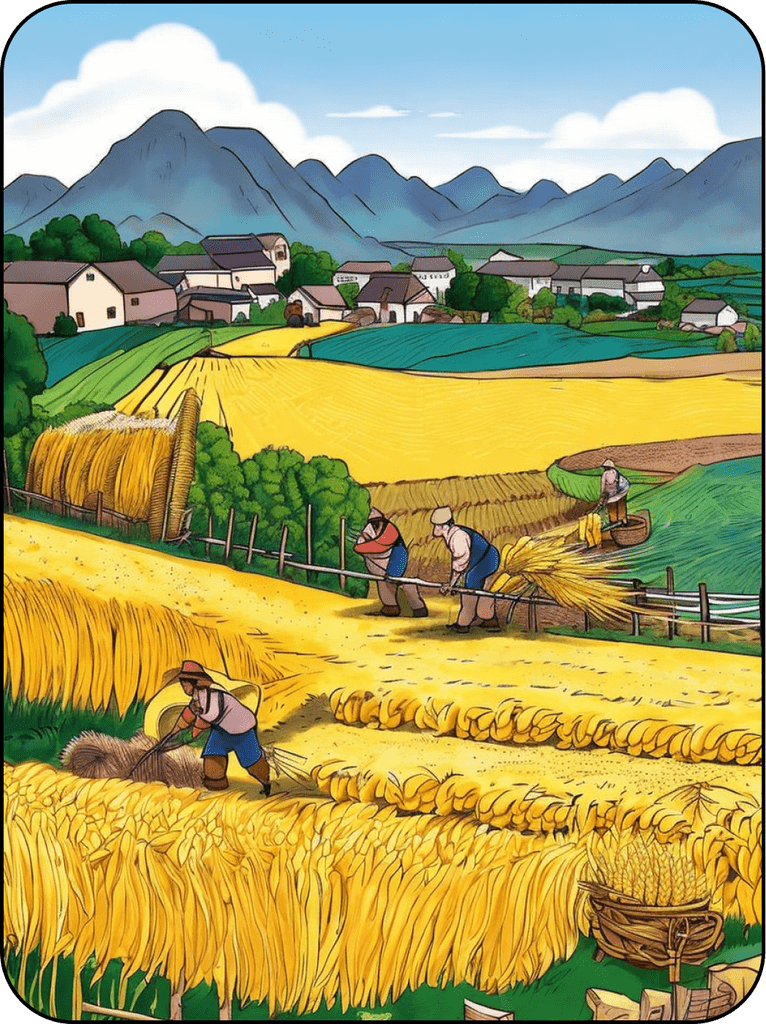} & 
The picture depicts a busy harvest scene in the countryside. In the golden wheat fields, farmers are diligently harvesting the ripe wheat. They tie the wheat stalks into bundles and stack them at the edge of the field. In the distance are rolling mountains, with houses and farmland at the foot of the mountains, forming a beautiful rural landscape. 
& 

\centering
\includegraphics[width=0.8\linewidth, height=3cm, keepaspectratio]{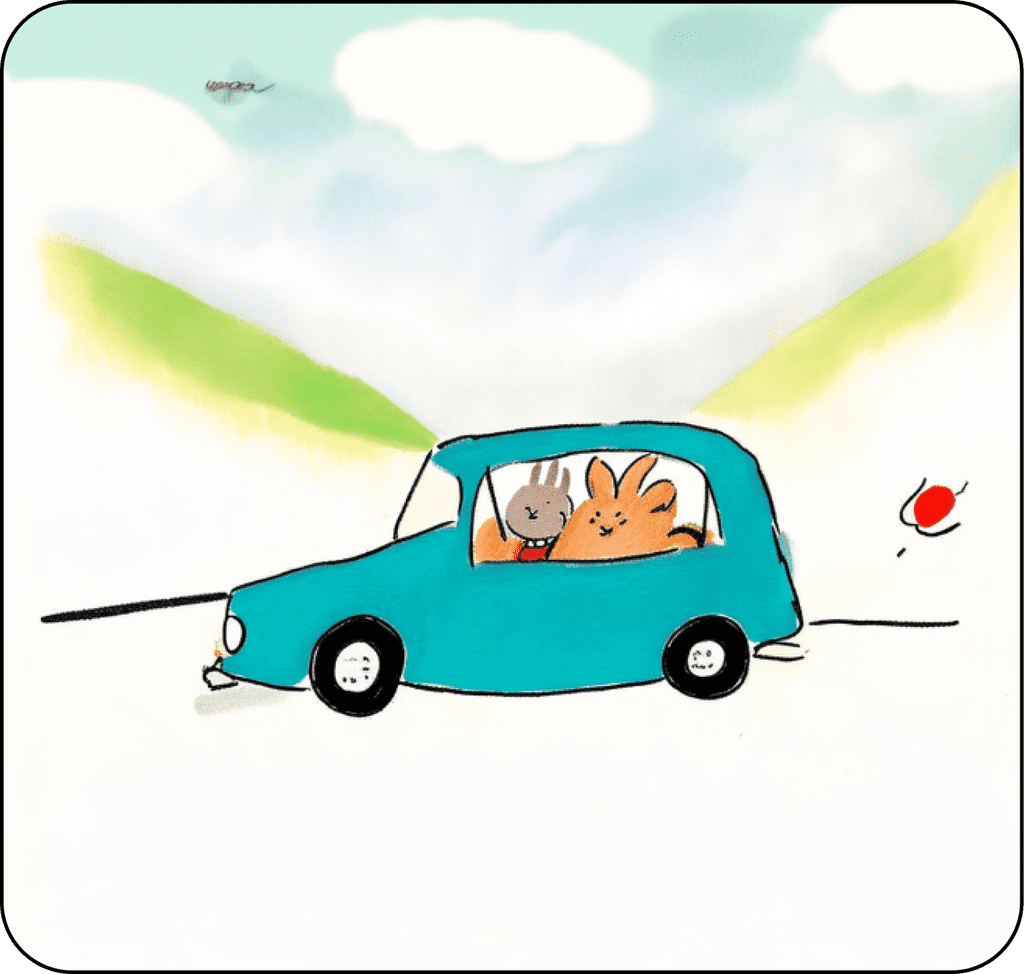} & 
A blue-green car is driving on the road with two rabbits in it. 
\\
\midrule

\centering
\includegraphics[width=0.8\linewidth, height=3cm, keepaspectratio]{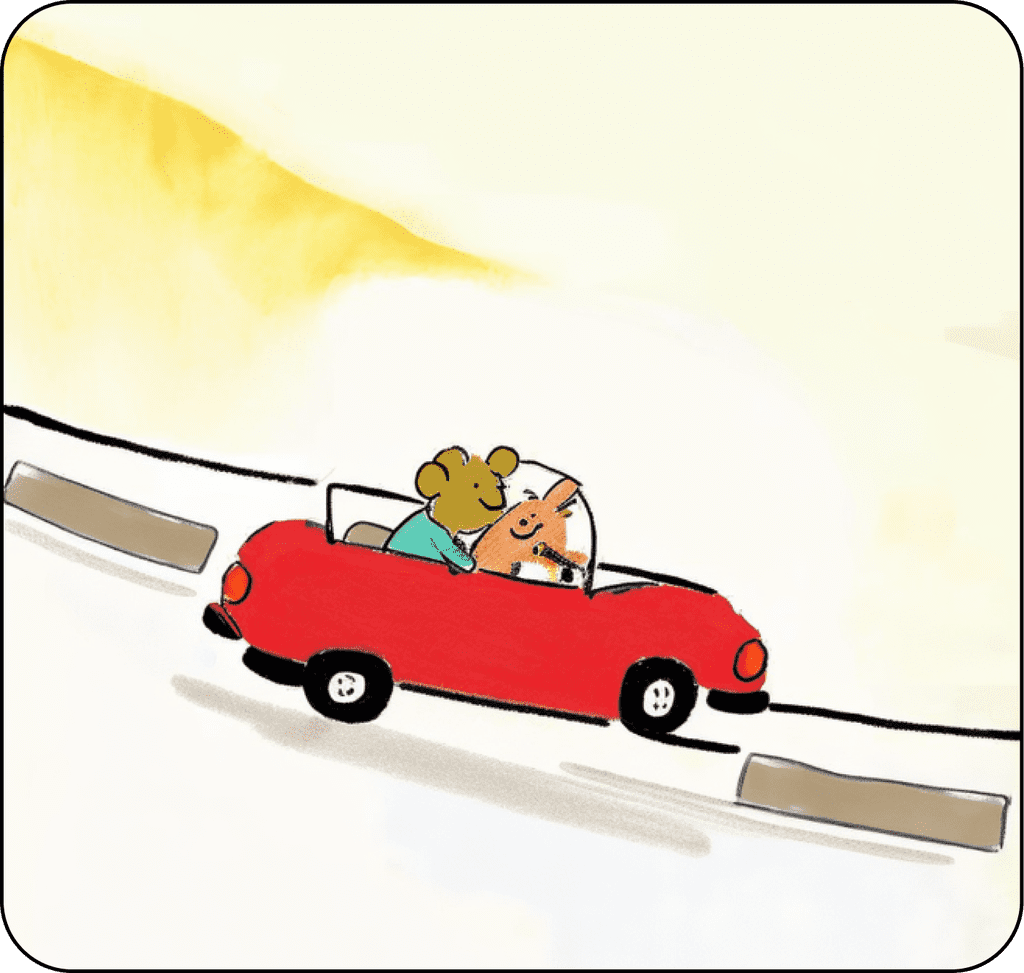} & 
A red car is driving on the road with two mice in it. 
& 

\centering
\includegraphics[width=0.8\linewidth, height=3cm, keepaspectratio]{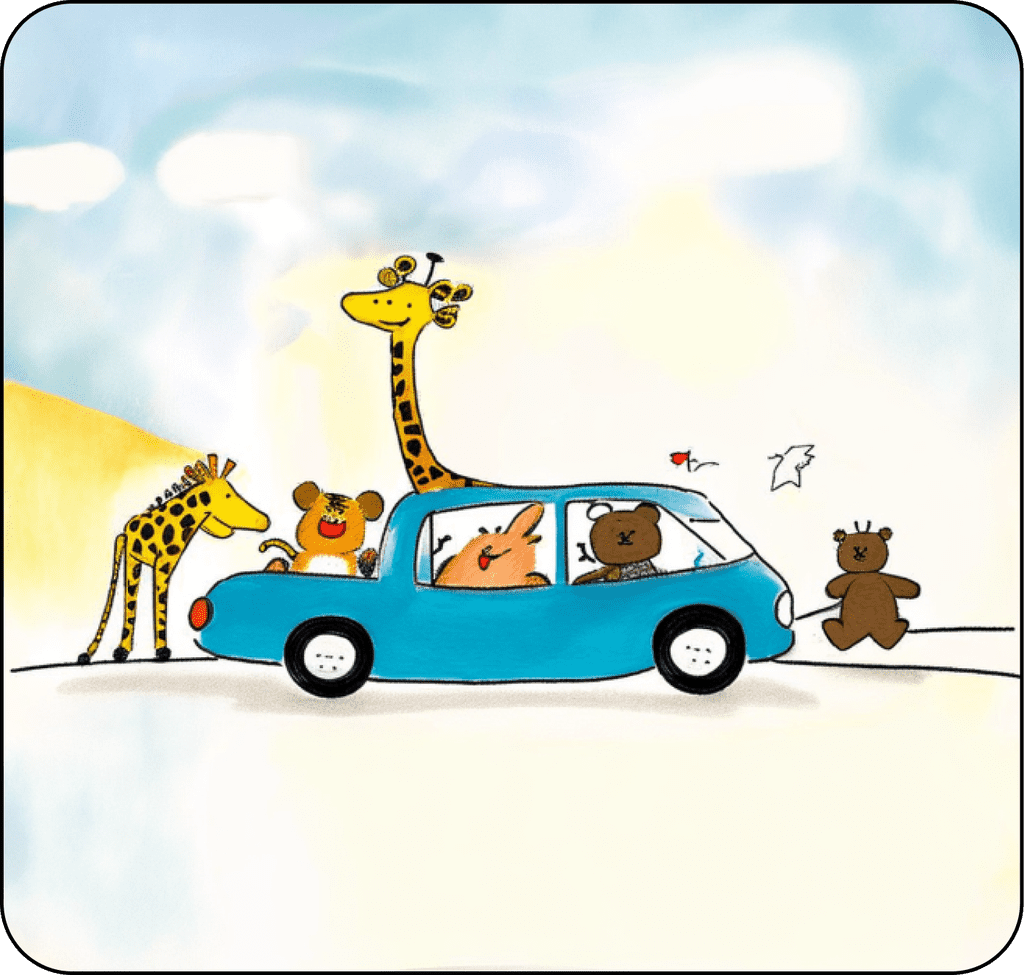} & 
A blue car is driving on the road with four animals inside: a giraffe, a tiger, a frog, and a teddy bear. 
\\
\midrule

\centering
\includegraphics[width=0.8\linewidth, height=3cm, keepaspectratio]{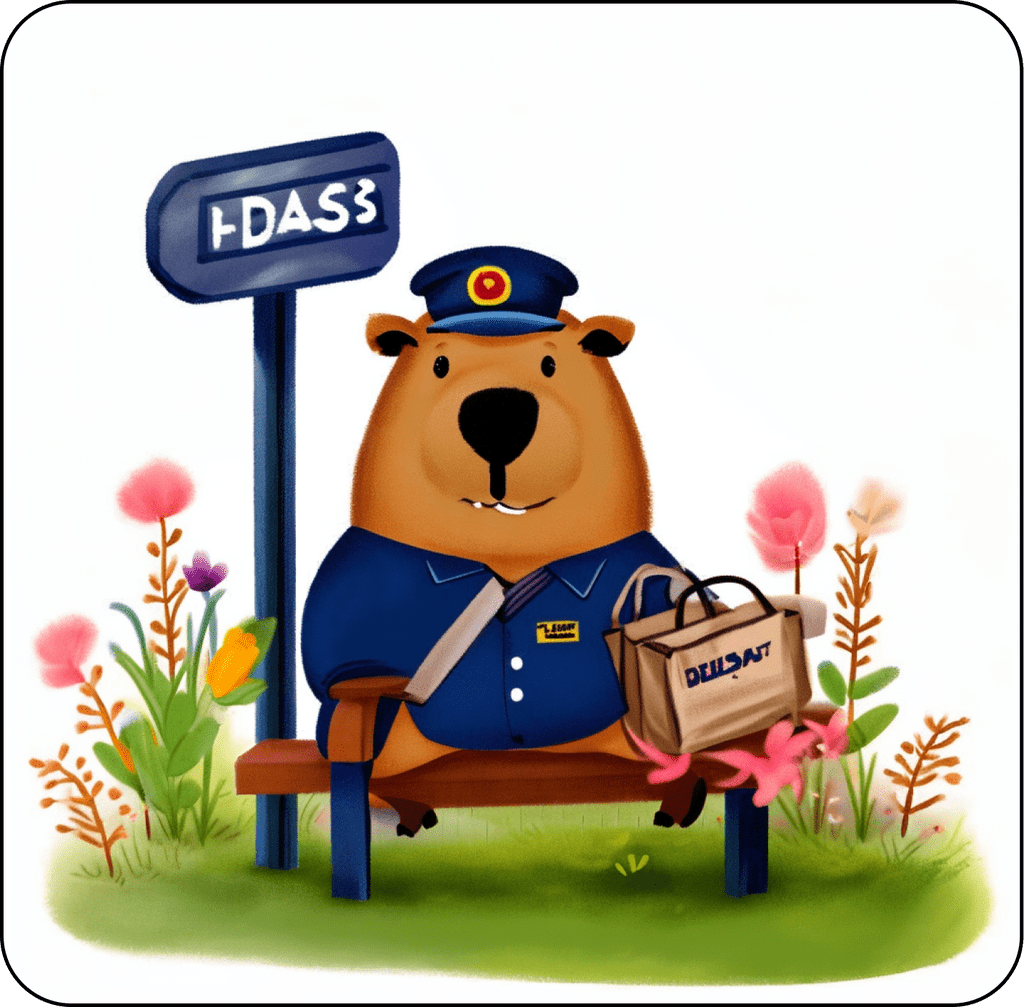} & 
A cartoon capybara wearing a blue uniform and hat and carrying a mailbag sits on a bench next to a bus stop sign that reads ``BUS". The capybara is surrounded by flowers and plants.
& 

\centering
\includegraphics[width=0.8\linewidth, height=3cm, keepaspectratio]{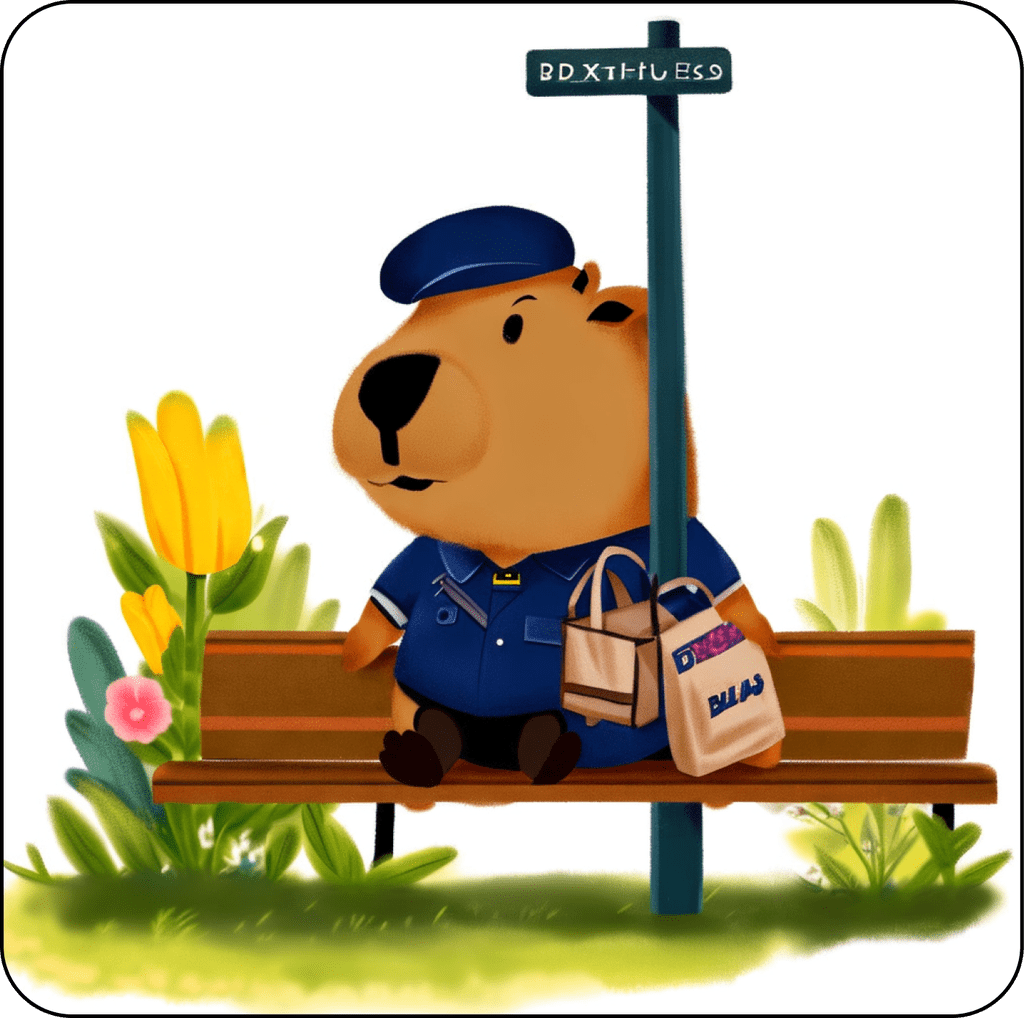} & 
A cartoon capybara wearing a blue uniform and hat and carrying a mailbag sits on a bench next to a bus stop sign that reads ``BUS". The capybara is surrounded by flowers and plants.
\\
\midrule

\centering
\includegraphics[width=0.8\linewidth, height=3cm, keepaspectratio]{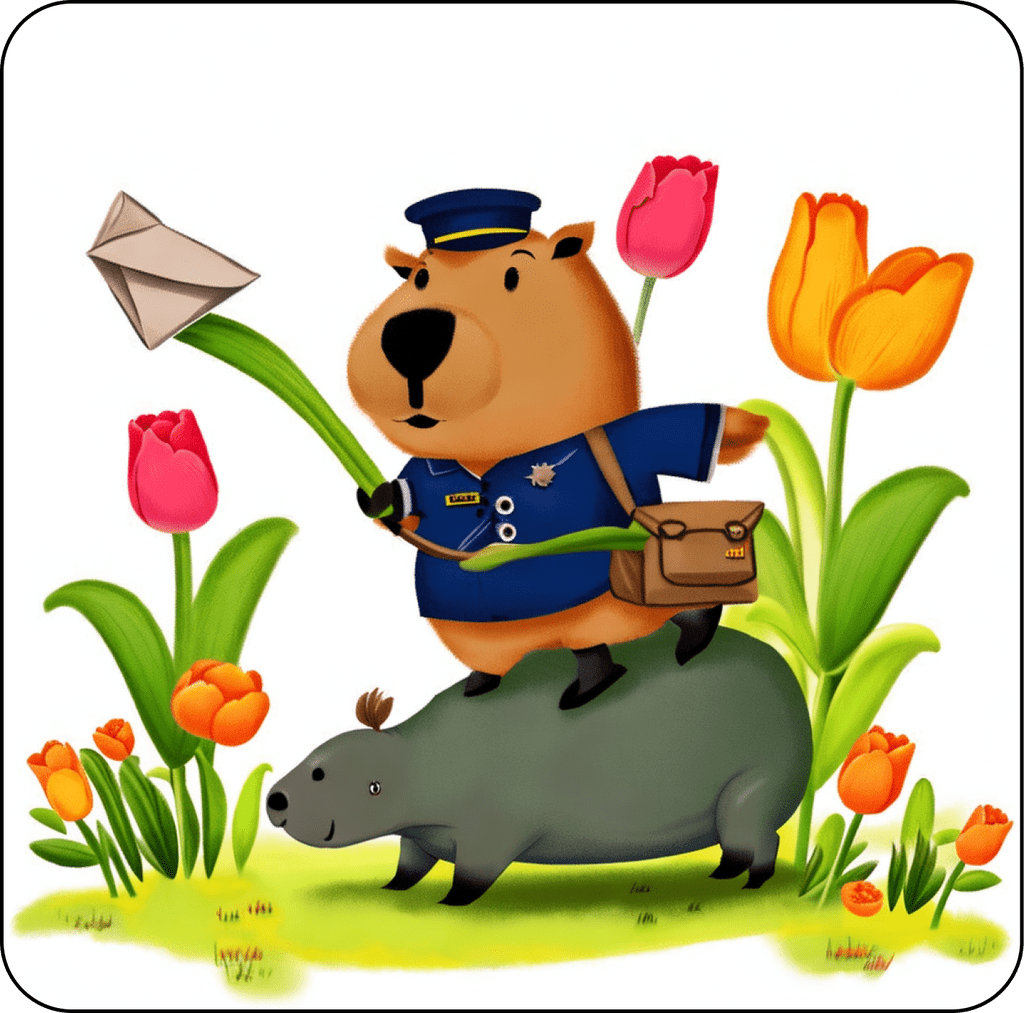} & 
A cartoon capybara wearing a blue uniform and hat and carrying a mailbag stands atop a turtle walking through a garden of tulips and other plants. The capybara has a stem of a plant in its mouth, and an envelope files behind it.  
& 

\centering
\includegraphics[width=0.8\linewidth, height=3cm, keepaspectratio]{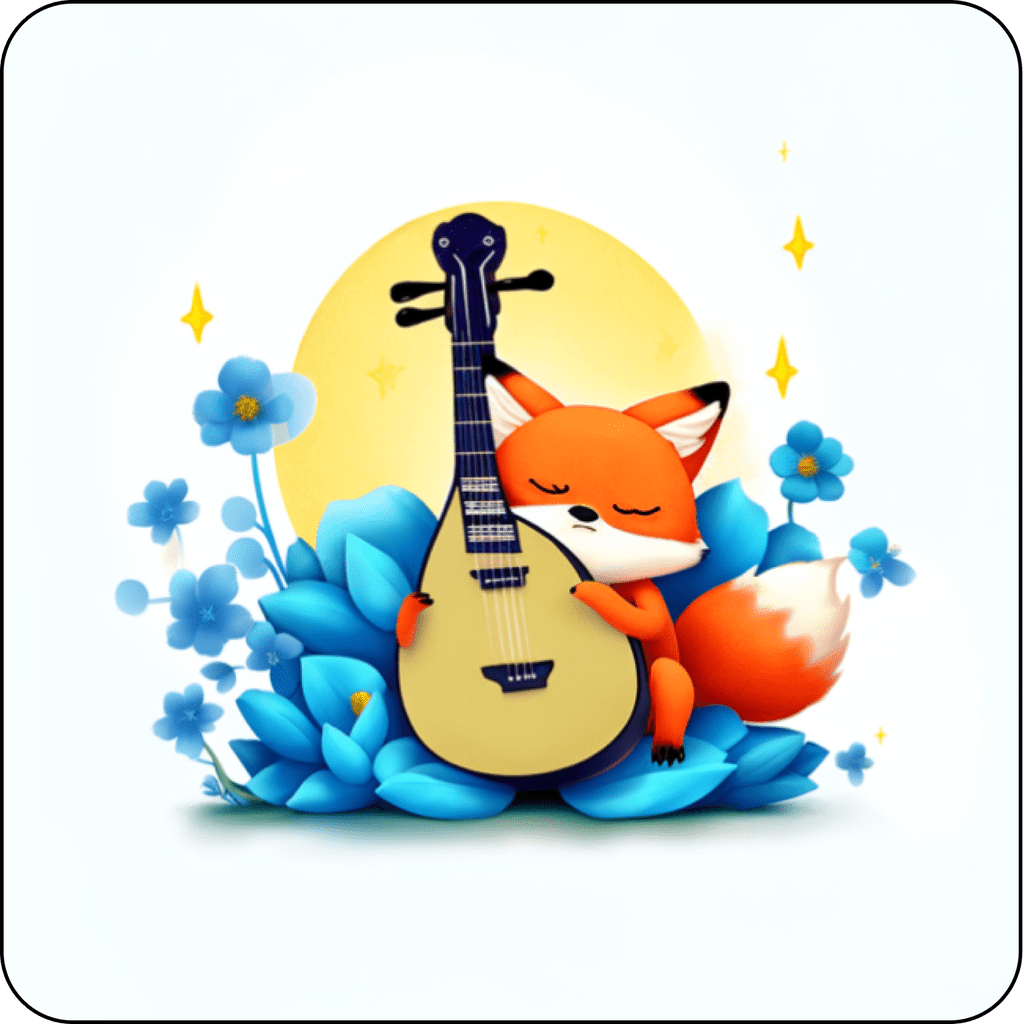} & 
An orange fox with closed eyes snuggles next to a pipa, surrounded by blooming blue flowers, with a golden full moon and twinkling stars in the background.
\\
\midrule

\centering
\includegraphics[width=0.8\linewidth, height=3cm, keepaspectratio]{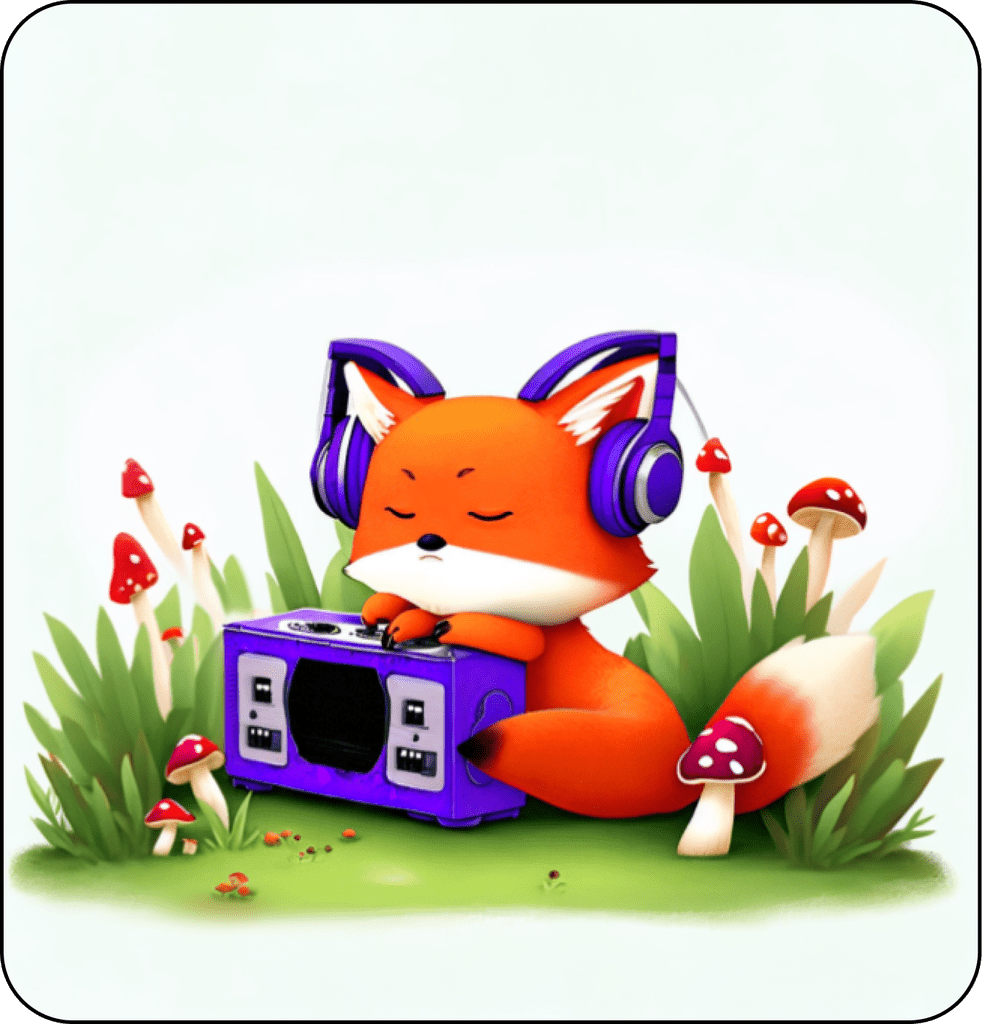} & 
An orange fox wearing headphones sits with its eyes closed on a grassy field, next to a purple vintage cassette player. Surrounding it are red mushrooms and plants bearing red fruits.
& 

\centering
\includegraphics[width=0.8\linewidth, height=3cm, keepaspectratio]{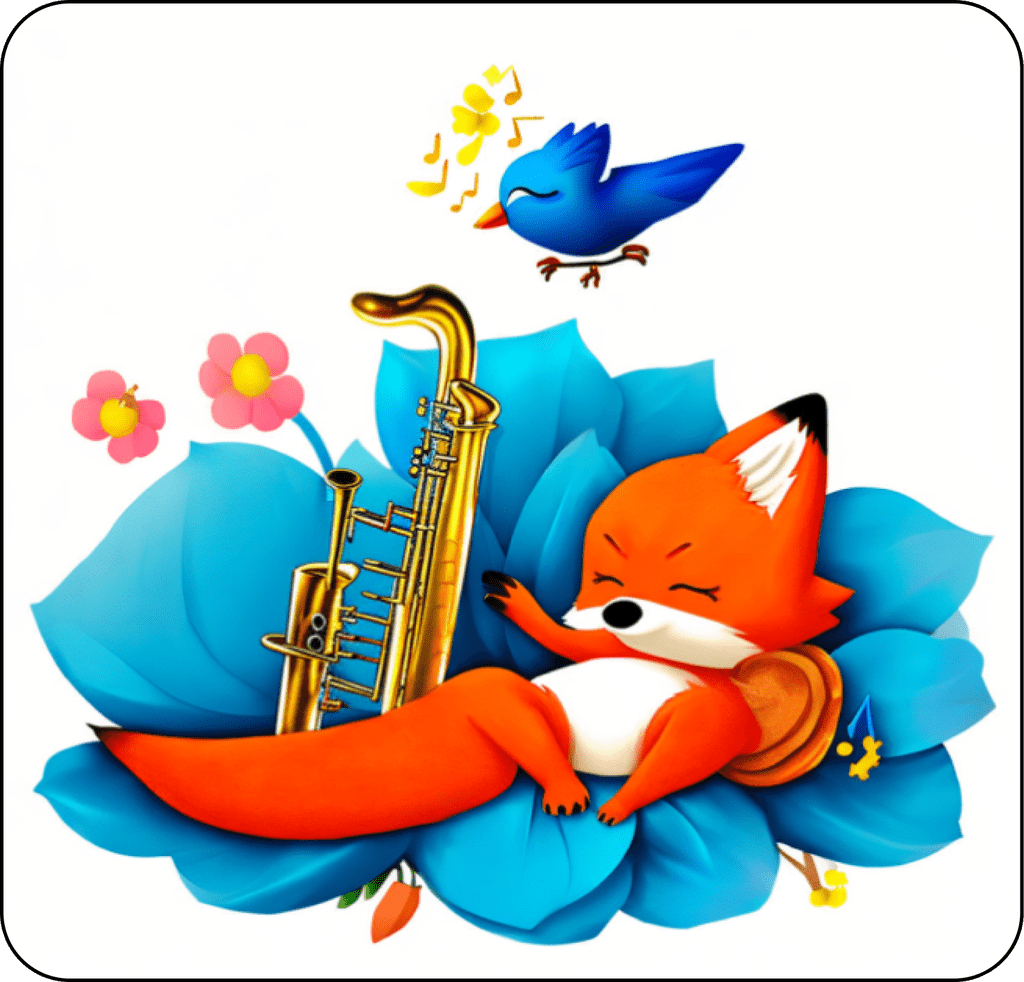} & 
An orange fox is peacefully sleeping with its eyes closed in a blue flower bed, next to a golden saxophone. A blue bird is joyfully singing on top of the saxophone.
\\
\midrule

\centering
\includegraphics[width=0.6\linewidth, height=3cm, keepaspectratio]{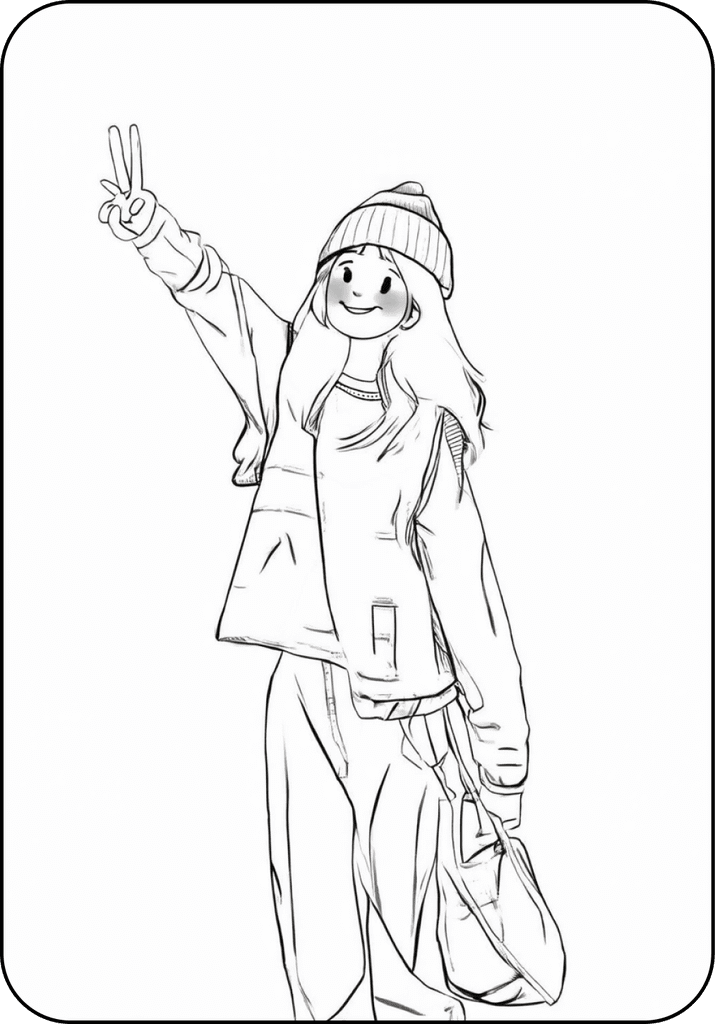} & 
A girl stands against a white background,
smiling and holding up her right hand in a peace sign. She wears a beanie hat, and her long hair falls over her shoulders. She’s dressed in a vest over a long-sleeved T-shirt, and loose-fitting pants. A bag hangs from her shoulder, and sneakers complete her casual look. 
& 

\centering
\includegraphics[width=0.6\linewidth, height=3cm, keepaspectratio]{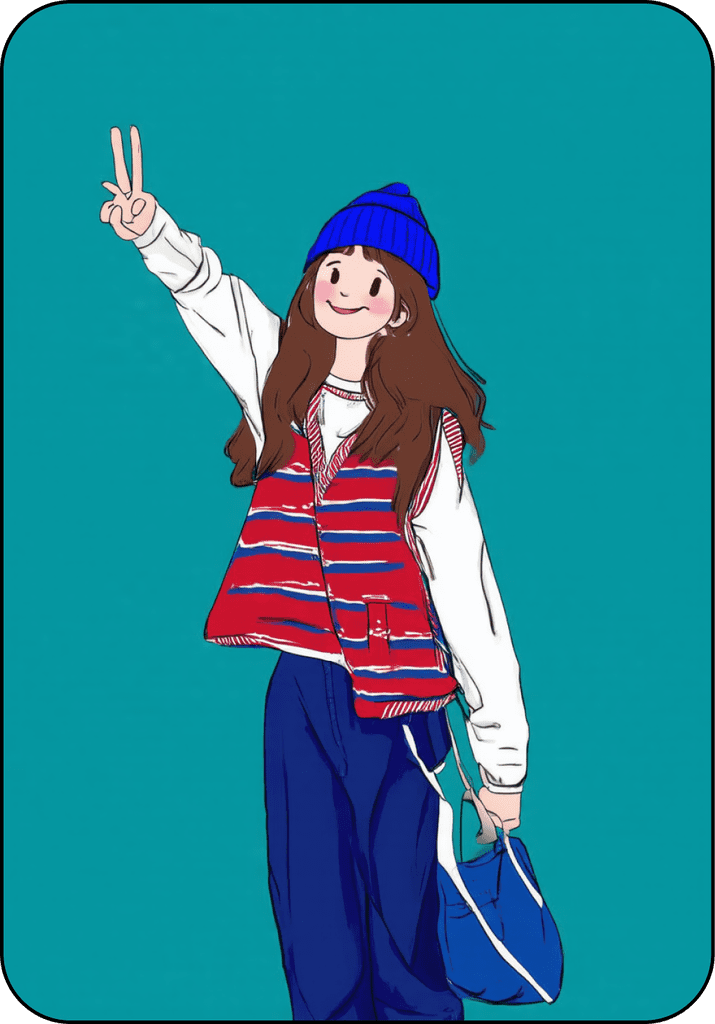} & 

A girl stands against a teal background, smiling and holding up her right hand in a peace sign. She wears a blue beanie hat, and her long brown hair falls over her shoulders. She’s dressed in a red and blue striped vest over a white long-sleeved T-shirt, and loose-fitting blue jeans. A blue bag hangs from her shoulder, and white sneakers complete her casual look. 
\\
\midrule

\centering
\includegraphics[width=0.6\linewidth, height=3cm, keepaspectratio]{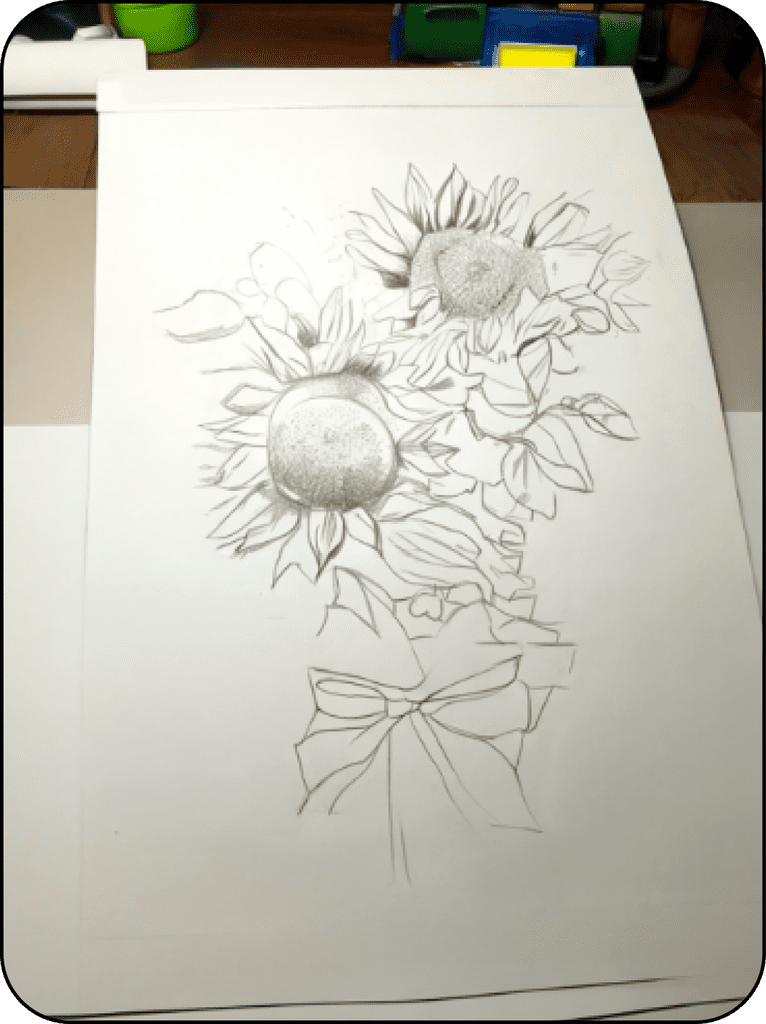} & 
On a white sheet of paper, a pencil sketch of a bouquet of flowers. The bouquet consists of two blooming sunflowers and some green leavers and small flowers wrapped in wrapping paper and tied with a bow. 
& 

\centering
\includegraphics[width=0.6\linewidth, height=3cm, keepaspectratio]{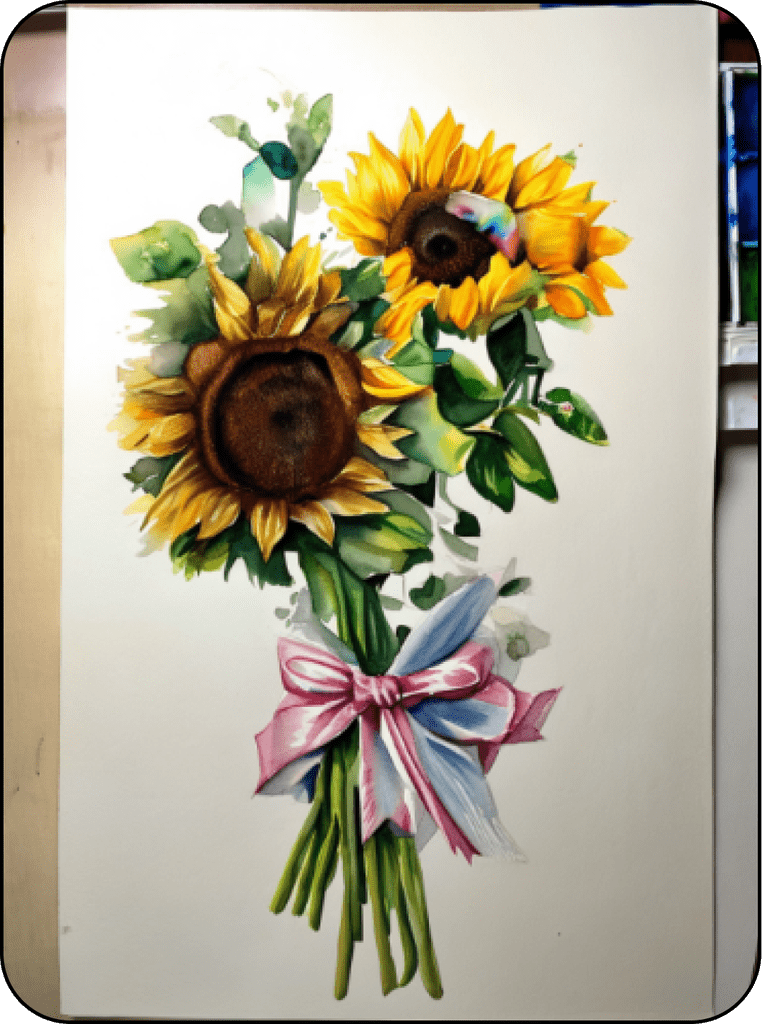} & 
On a cream-colored sheet of paper, all areas of a bouquet are painted with watercolor paints. The bouquet consists of two blooming sunflowers and small flowers, wrapped in wrapping paper and tied with a bow. 
\\
\midrule

\centering
\includegraphics[width=0.8\linewidth, height=3cm, keepaspectratio]{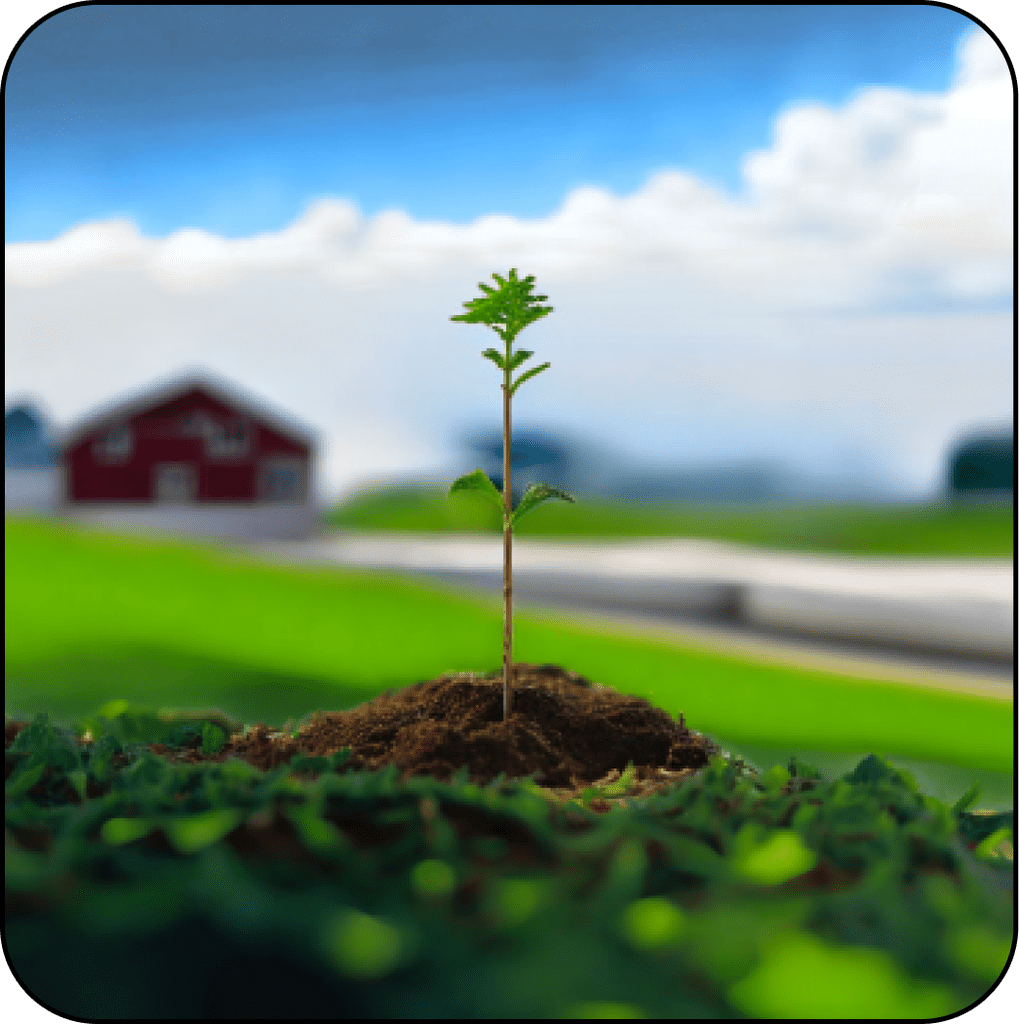} & 
On a sunny day, on a green grassland, several houses can be seen in the distance, with blue sky and white clouds dotted the sky. In the center of the screen, a small sapling has just been planted in the soil. It looks very fragile, but full of vitality.
& 

\centering
\includegraphics[width=0.8\linewidth, height=3cm, keepaspectratio]{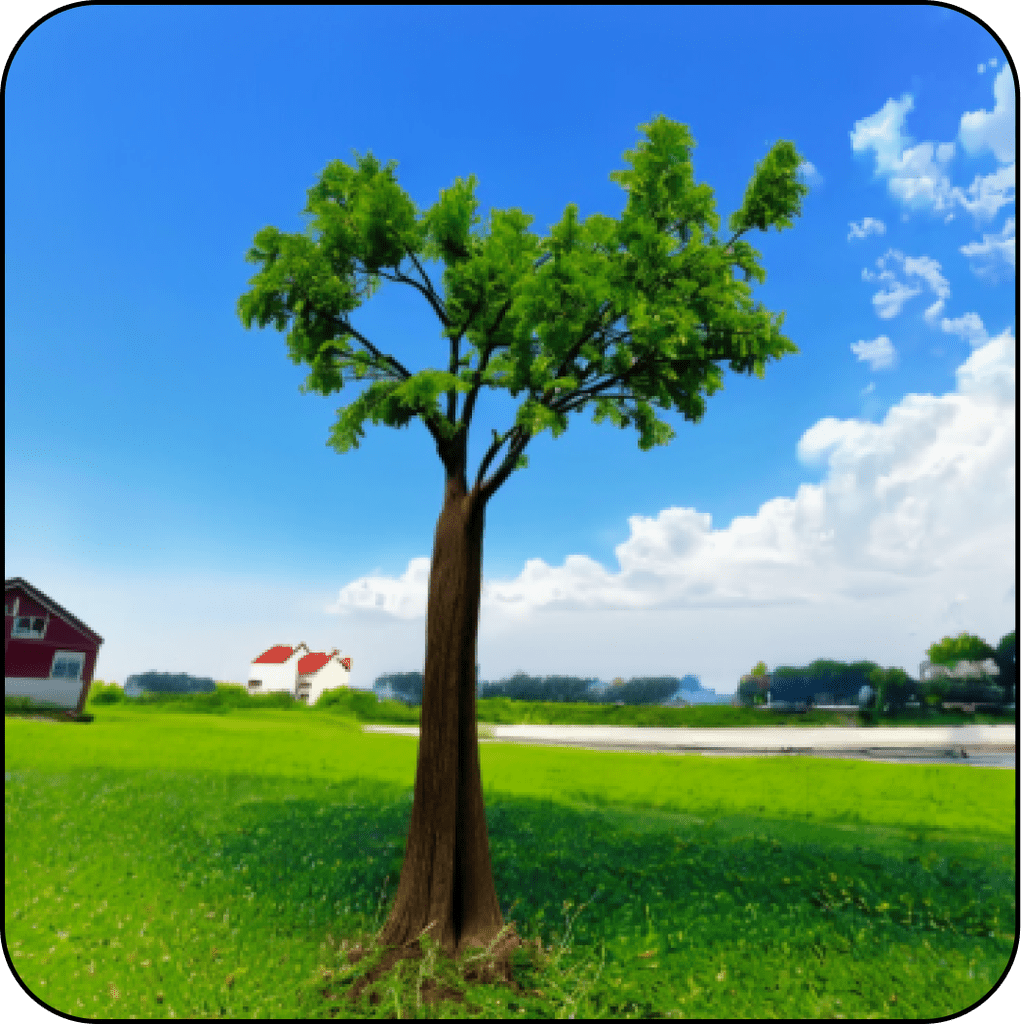} & 

On a sunny day, on a green grassland, several houses can be seen in the distance, with blue sky and white clouds dotted the sky. In the center of the screen, the once weak sapling has grown much taller, about 1 meter high, and its branches and leaves are even more lush.
\\
\midrule

\centering
\includegraphics[width=0.8\linewidth, height=3cm, keepaspectratio]{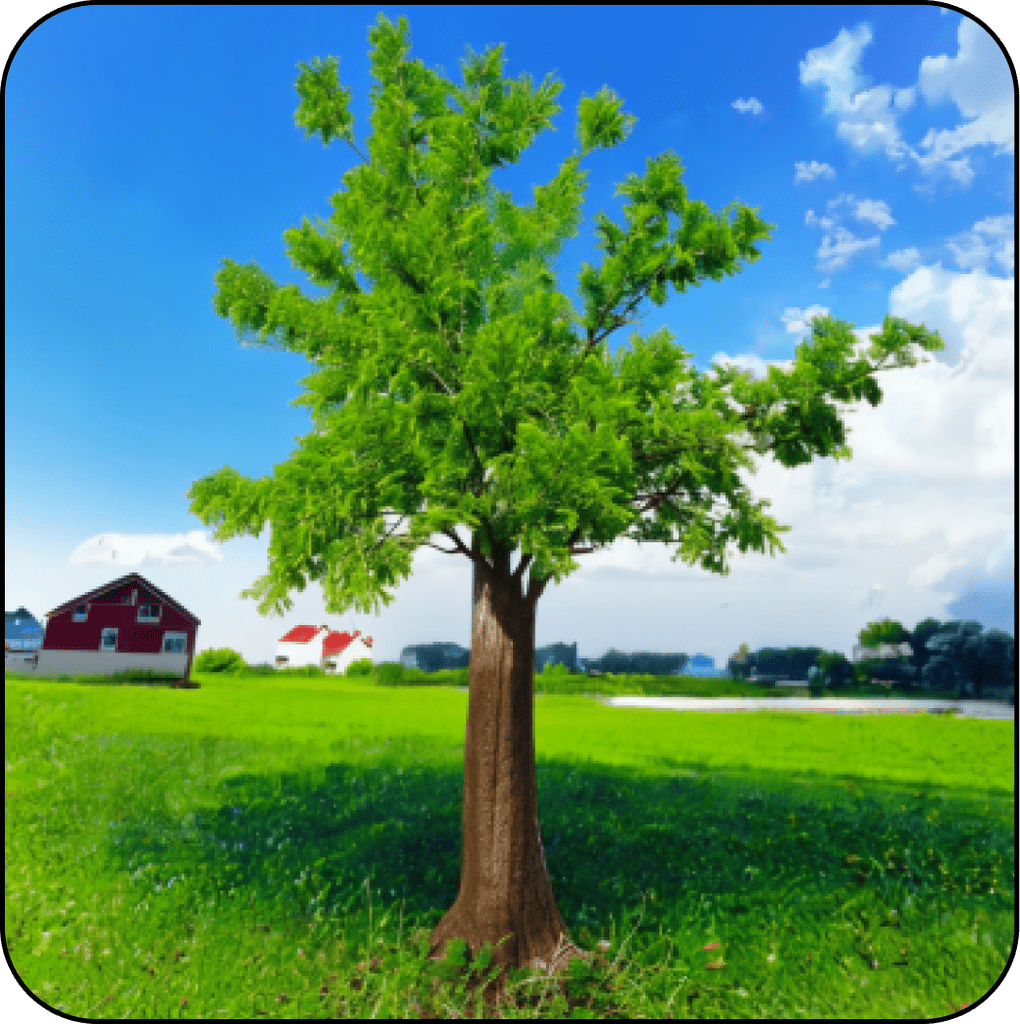} & 
On a sunny day, on a green grassland, several houses can be seen in the distance, with blue sky and white clouds dotted the sky. In the center of the screen, the tree has grown to 2 meters high, with lush branches and leaves, full of vitality.
& 

\centering
\includegraphics[width=0.8\linewidth, height=3cm, keepaspectratio]{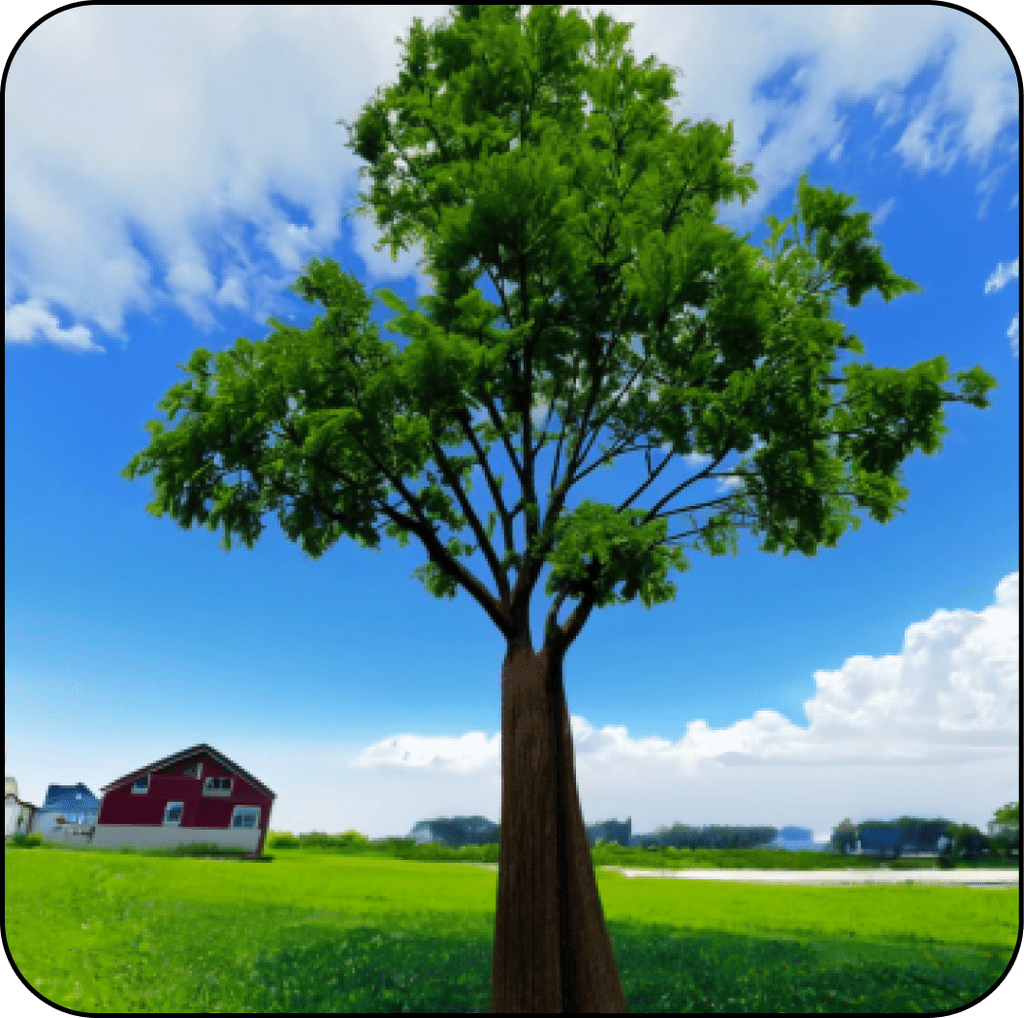} & 

On a sunny day, on a green grassland, several houses can be seen in the distance, with blue sky and white clouds dotted the sky. In the center of the screen, the tree has grown into a 3-meter-tall tall tree, with lush branches and leaves, covering the sky and sun. Standing west, it exudes strong vitality.
\\
\midrule

\centering
\includegraphics[width=0.6\linewidth, height=3cm, keepaspectratio]{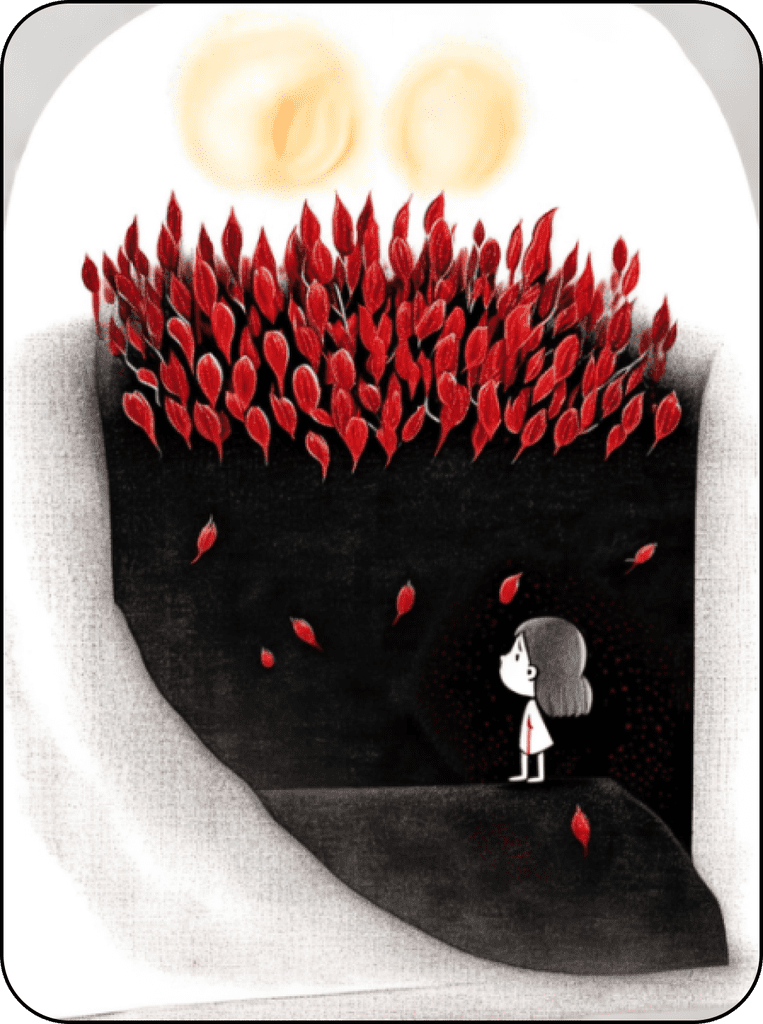} & 
A little girl standing on a cliff edge, looking up at a cluster of red plants with a glowing sphere in the center. Below the cliff is a black forest, with a white background speckled with red plants.
& 

\centering
\includegraphics[width=0.6\linewidth, height=3cm, keepaspectratio]{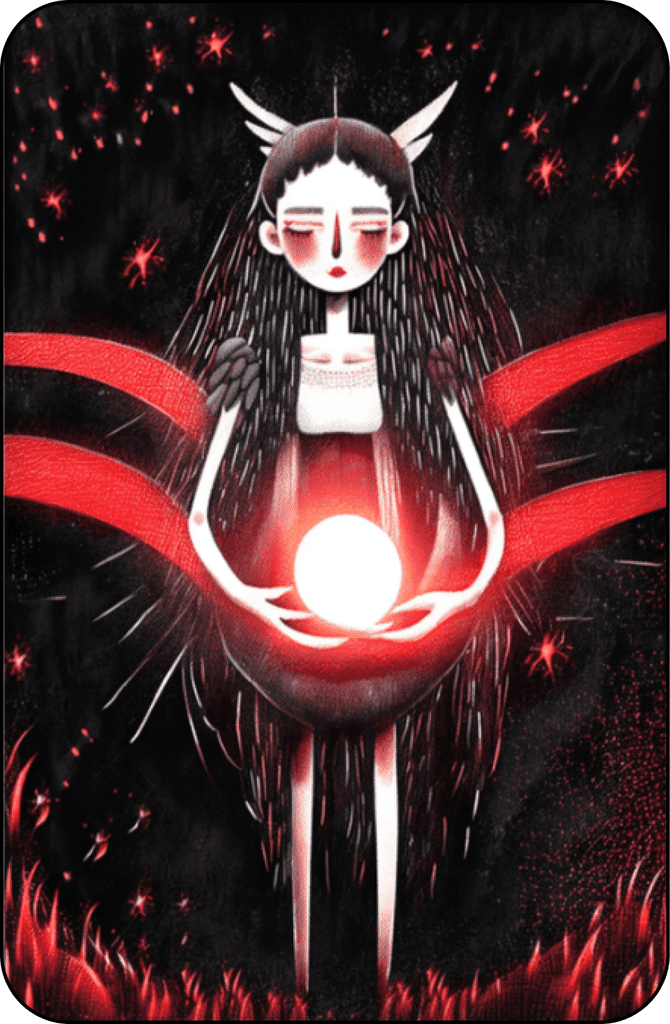} & 

A winged girl holding a glowing sphere, surrounded by swirling red lines and stars. Her background is black, dotted with red plants. 
\\
\midrule

\centering
\includegraphics[width=0.6\linewidth, height=3cm, keepaspectratio]{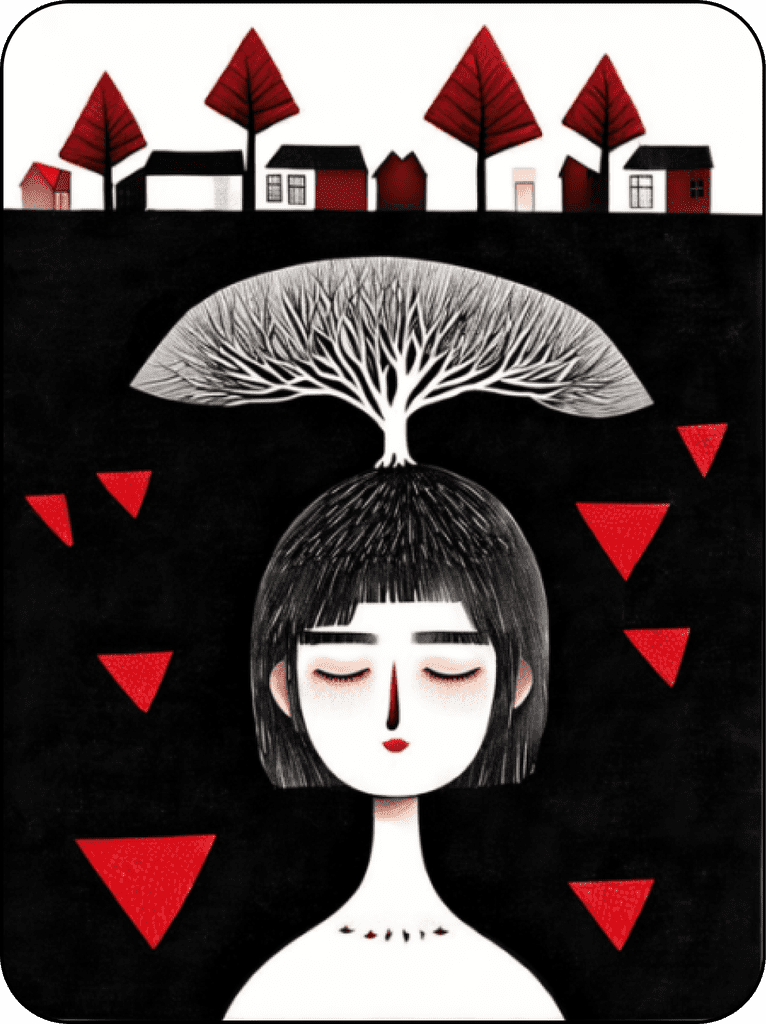} & 
A girl with her eyes closed, a bare tree on her head. There are small houses around the tree, set against a black background adorned with red leaves and triangles.  
& 

\centering
\includegraphics[width=0.6\linewidth, height=3cm, keepaspectratio]{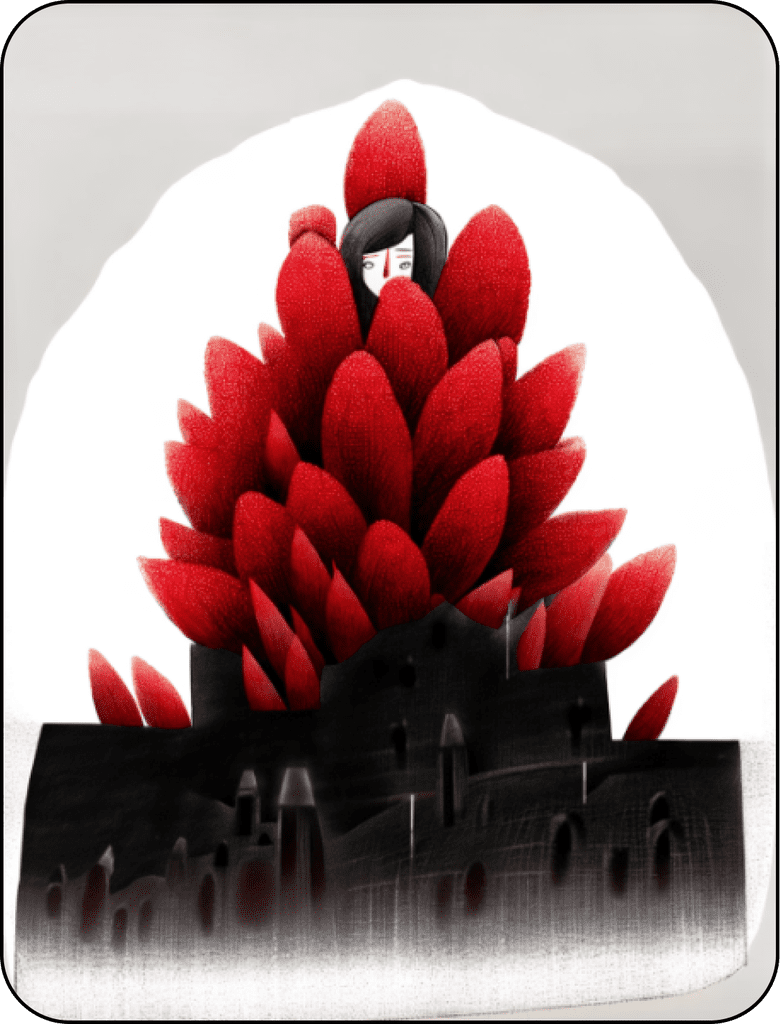} & 

A winged girl, mouth open, looking up at a cluster of red plants with a glowing sphere in the center. Beneath her are small houses and black plants. 
\\
\midrule

\centering
\includegraphics[width=\linewidth, height=3cm, keepaspectratio]{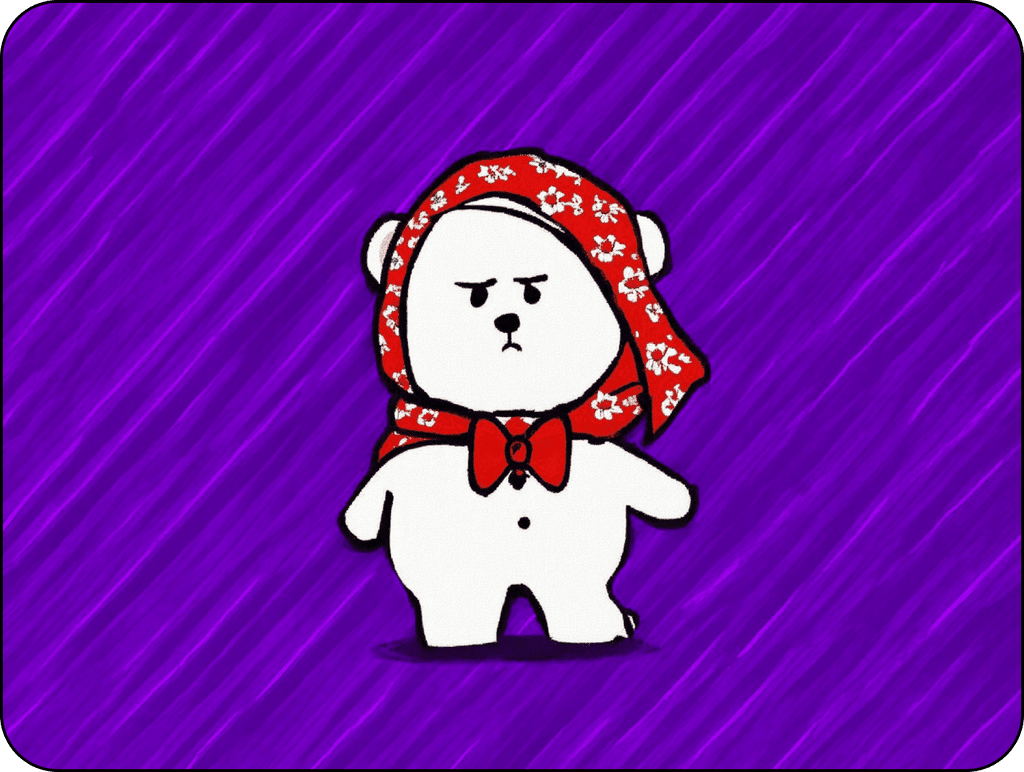} & 
A cartoon polar bear wearing a red floral headscarf and a red bow tie, with a serious expression, stands in the center of a purple striped background. 
& 

\centering
\includegraphics[width=\linewidth, height=3cm, keepaspectratio]{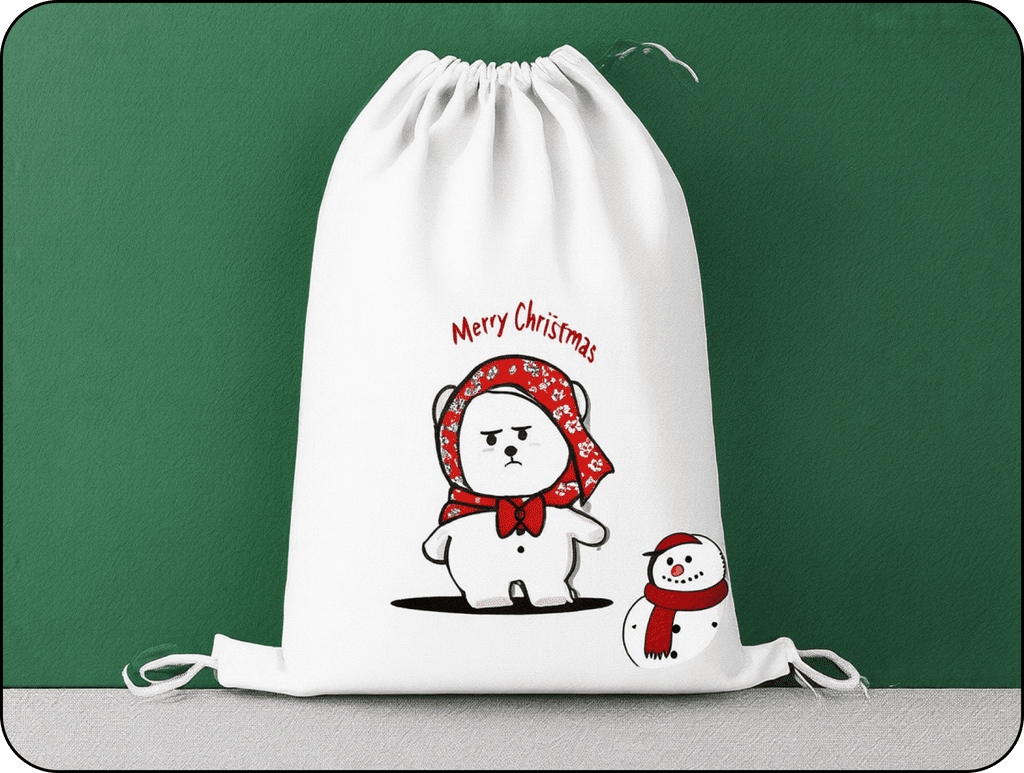} & 

A white drawstring backpack with a cartoon polar bear wearing a red floral headscarf and a snowman with a red scarf and a carrot nose printed on it, with ``Merry Christmas" written above. 
\\
\midrule

\centering
\includegraphics[width=0.8\linewidth, height=3cm, keepaspectratio]{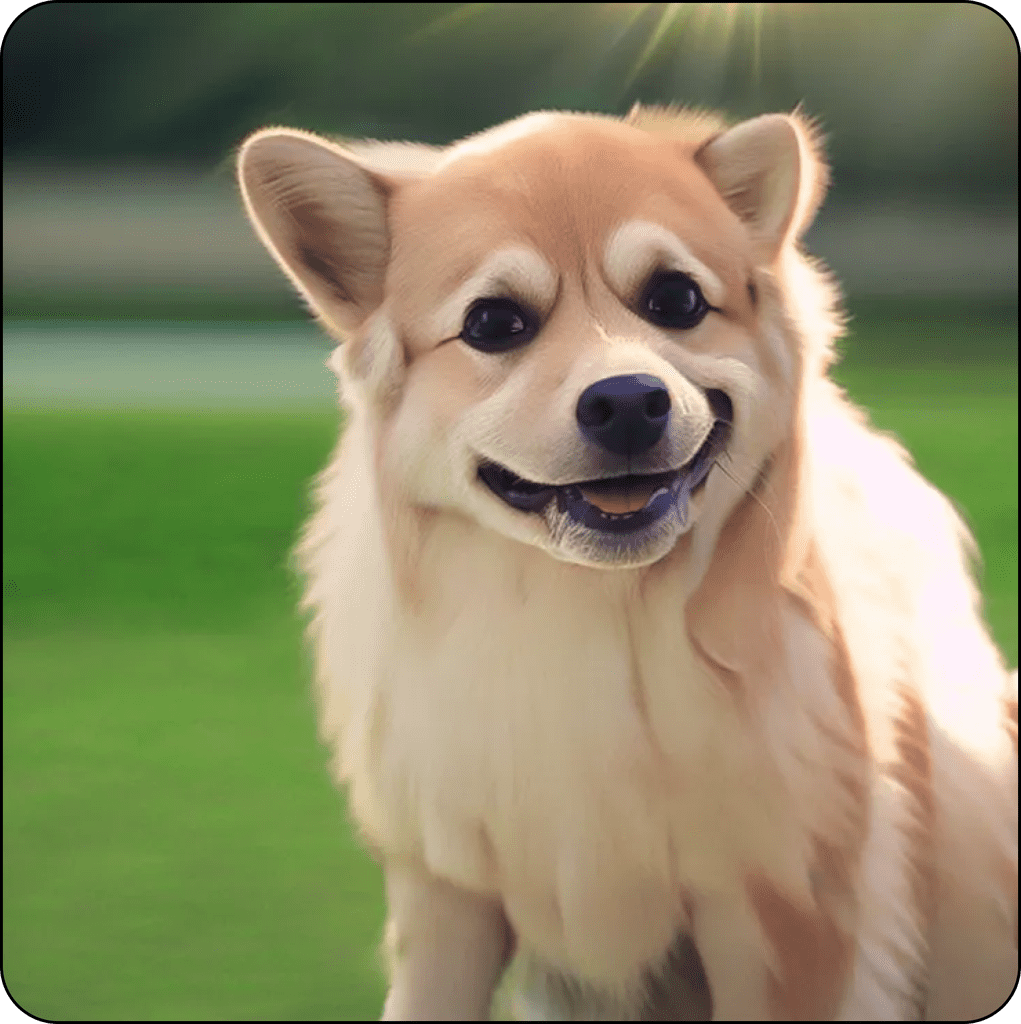} & 
The picture shows a cute little dog with fluffy fur, round eyes, and a lively nose, smiling kindly at the camera. The puppy's fur color is light brown, and its body posture appears very relaxed. Its limbs naturally hang on both sides of its body, and its tail gently sways, as if welcoming its owner's arrival. The background of the puppy is a green lawn, with sunlight shining through the clouds, illuminating the entire scene.
& 

\centering
\includegraphics[width=0.8\linewidth, height=3cm, keepaspectratio]{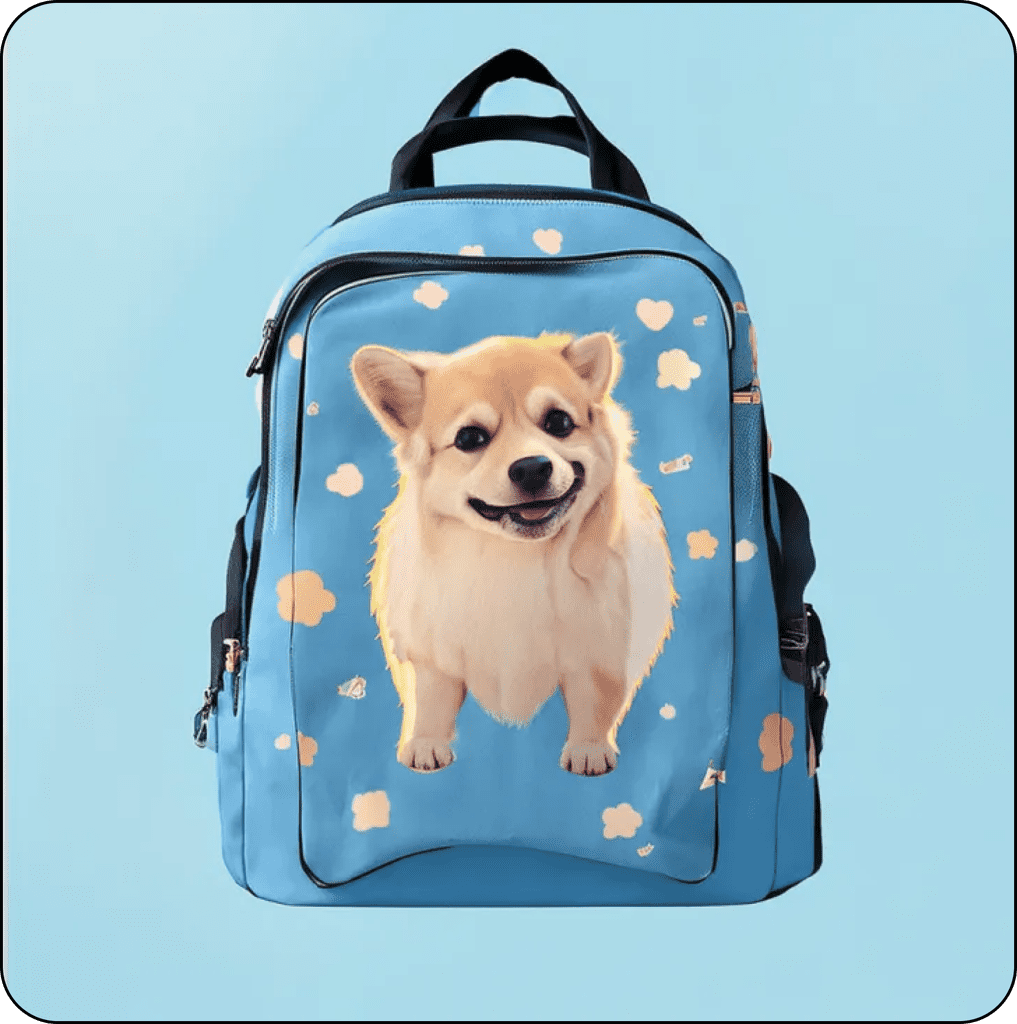} & 

The picture shows a cute cartoon style backpack with a blue themed color and some cute patterns printed on it. On the front of the backpack, there is a photo of a small dog with fluffy fur, round eyes, and a lively nose. It is facing the camera with a friendly smile. The photo of the puppy is printed in the center of the backpack, with a moderate size that matches the overall style of the backpack. The background of the backpack is a bright yellow color, creating a lively atmosphere.
\\
\midrule

\centering
\includegraphics[width=0.6\linewidth, height=3cm, keepaspectratio]{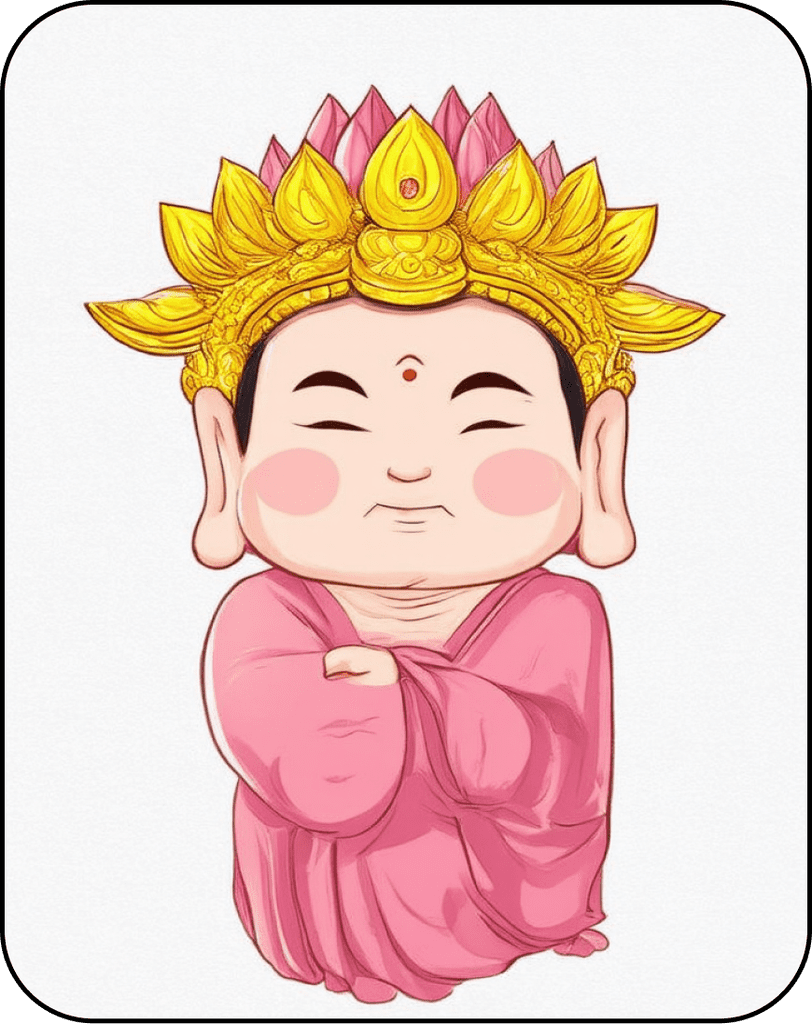} & 
A carton character, resembling Buddha, wearing a golden lotus headdress and pink monk robes. His hands are hidden in his sleeves. He has a slight smile and closed eyes, showing contentment. 
& 

\centering
\includegraphics[width=0.6\linewidth, height=3cm, keepaspectratio]{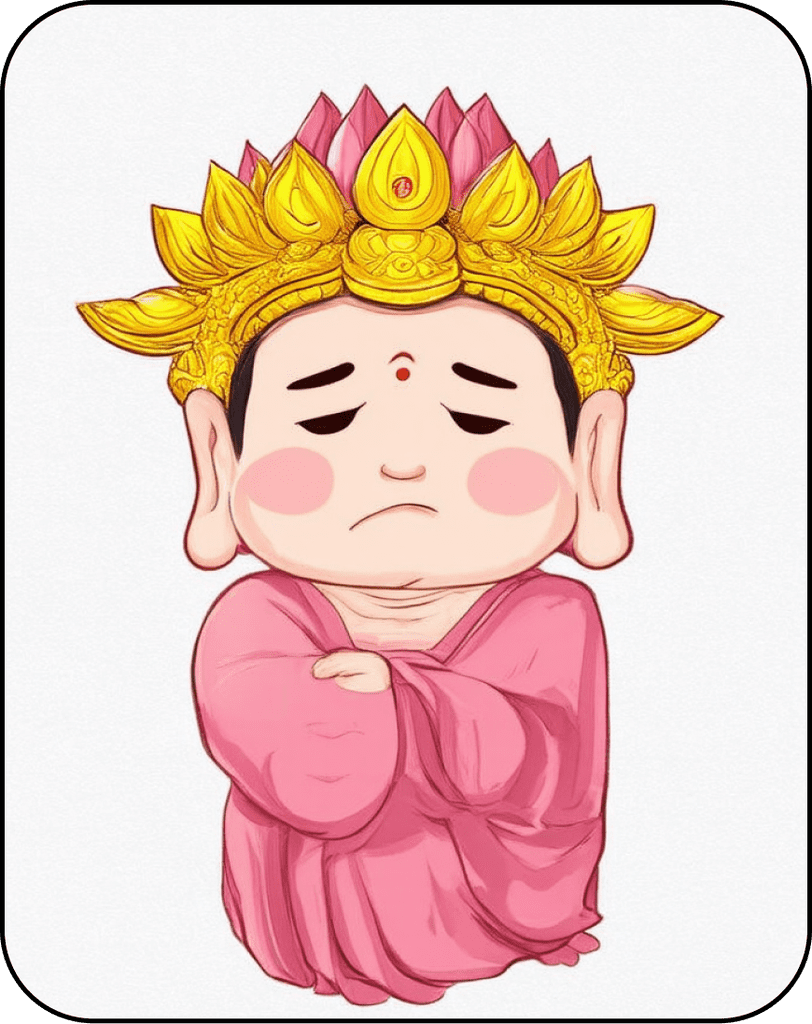} & 

A carton character, resembling Buddha, wearing a golden lotus headdress and pink monk robes. His hands are hidden in his sleeves. He has a furrowed brow and downturned mouth, showing sadness.  
\\
\midrule

\centering
\includegraphics[width=0.6\linewidth, height=3cm, keepaspectratio]{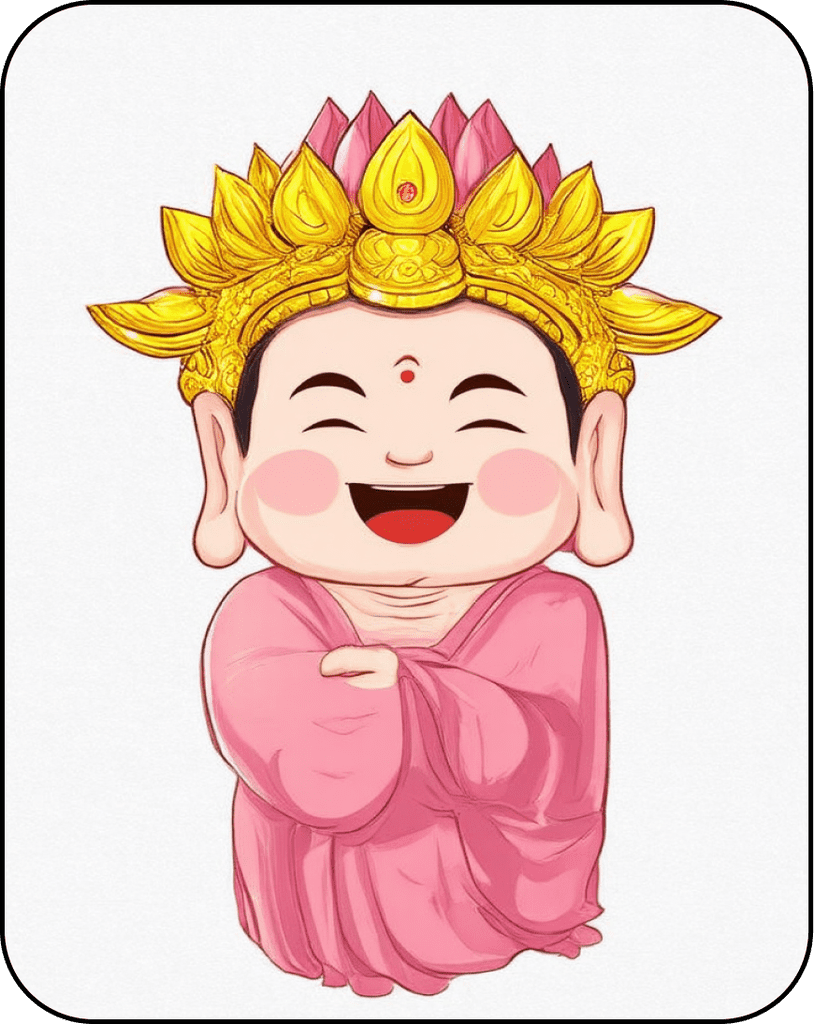} & 
A carton character, resembling Buddha, wearing a golden lotus headdress and pink monk robes. His hands are hidden in his sleeves. He has a big smile and closed eyes, showing joy. 
& 

\centering
\includegraphics[width=0.6\linewidth, height=3cm, keepaspectratio]{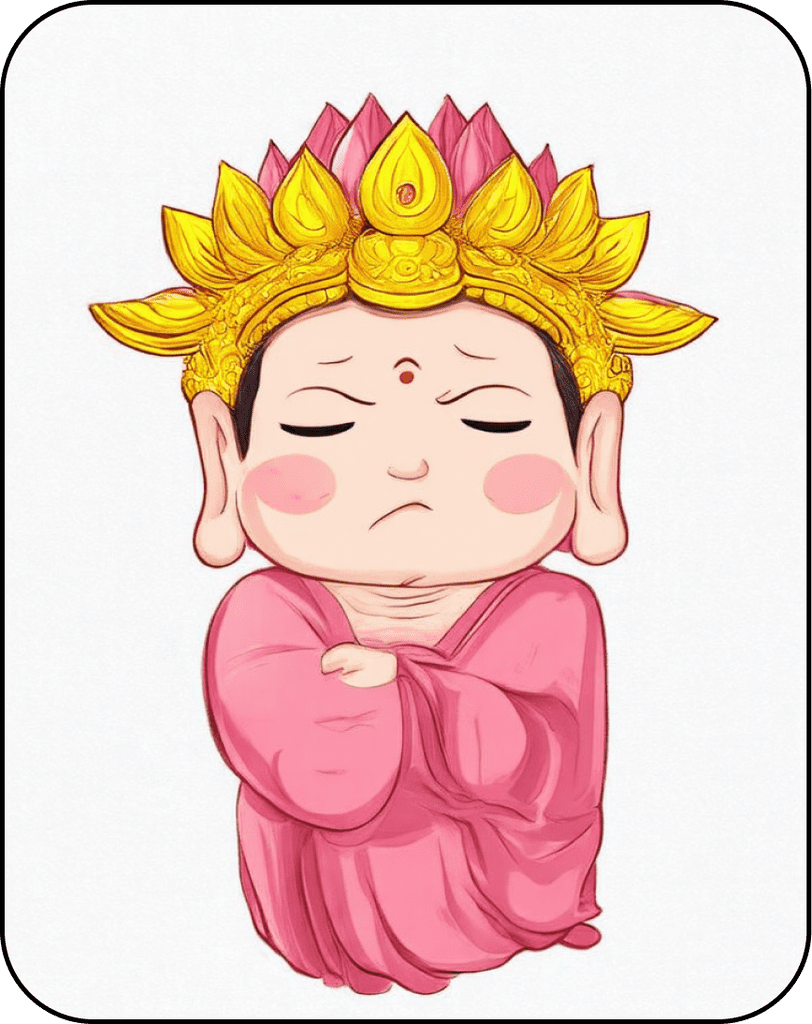} & 

A carton character, resembling Buddha, wearing a golden lotus headdress and pink monk robes. His hands are hidden in his sleeves. His eyes are closed and his eyebrows are slightly furrowed, showing exhaustion.
\\
\midrule

\centering
\includegraphics[width=0.6\linewidth, height=3cm, keepaspectratio]{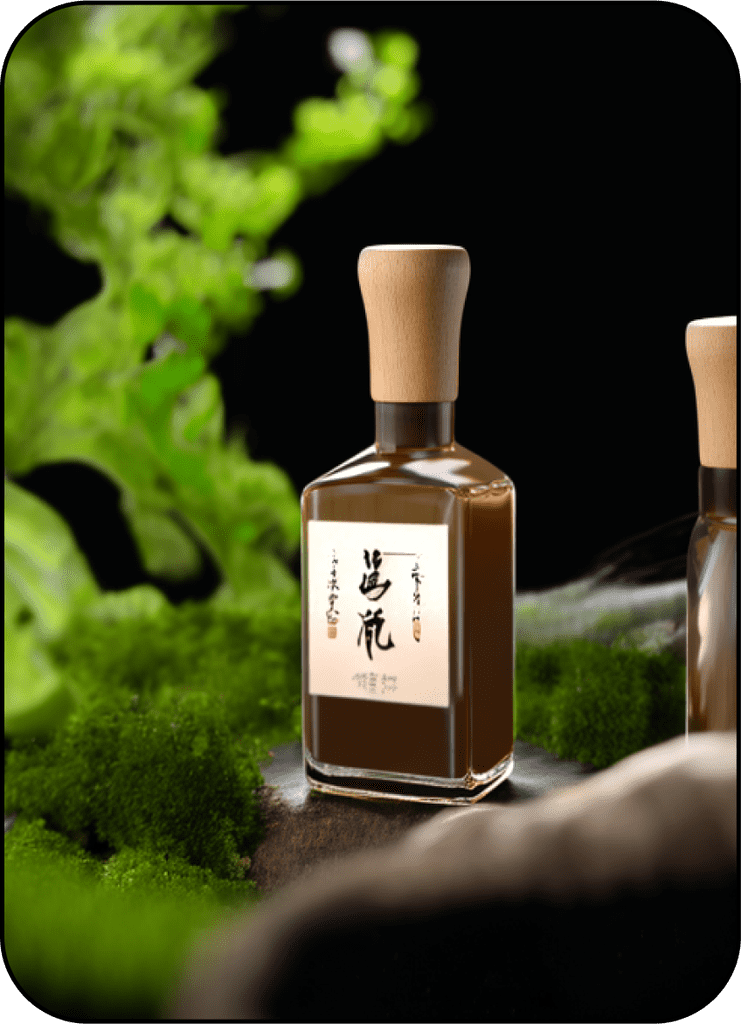} & 
A photo of a square glass bottle containing a light brown liquid. The bottle is sealed with a wooden stopper and has a label with Chinese characters and design elements. The bottle is surrounded by green plants, creating a natural feel.
& 

\centering
\includegraphics[width=0.6\linewidth, height=3cm, keepaspectratio]{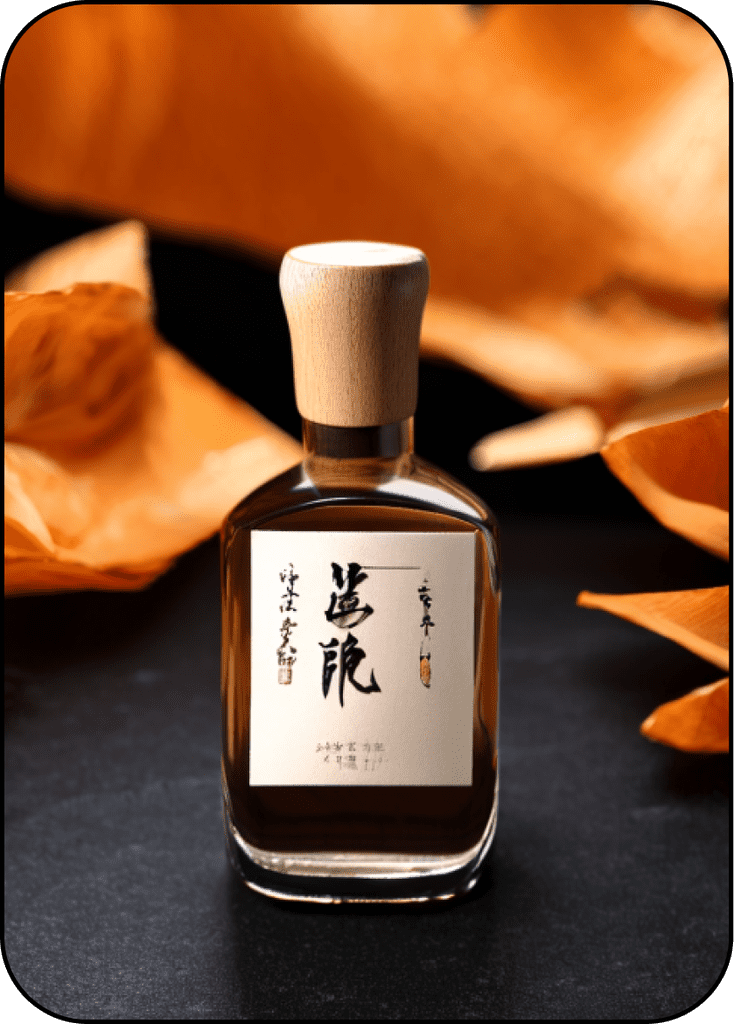} & 

A photo of a square glass bottle containing a light brown liquid. The bottle is sealed with a wooden stopper and has a label with Chinese characters and design elements.  It’s set against a dark background and surrounded by scattered orange petals, creating a professional studio ambiance.
\\
\midrule

\centering
\includegraphics[width=\linewidth, height=3cm, keepaspectratio]{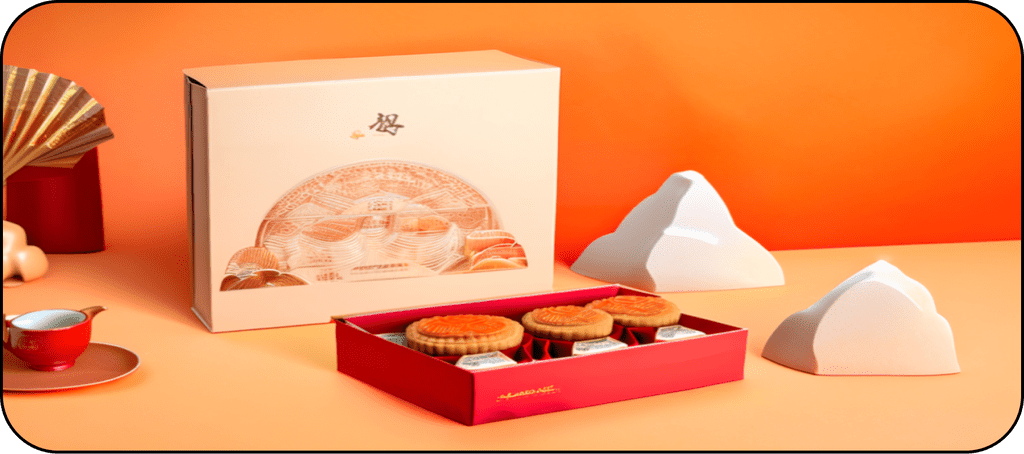} & 
A product photography image with a warm orange background. The main focus is a golden gift box with a fan-shaped cutout pattern and landscape painting decorations; three golden mooncakes are visible inside. Next to the gift box is a red tray, also containing three mooncakes. Abstract white mountain-shaped decorations are on the right, and a traditional Chinese red fan and a teacup with Chinese-style patterns are on the left. The overall tone is warm and festive, highlighting the product's high-end feel and traditional cultural elements.
& 

\centering
\includegraphics[width=\linewidth, height=3cm, keepaspectratio]{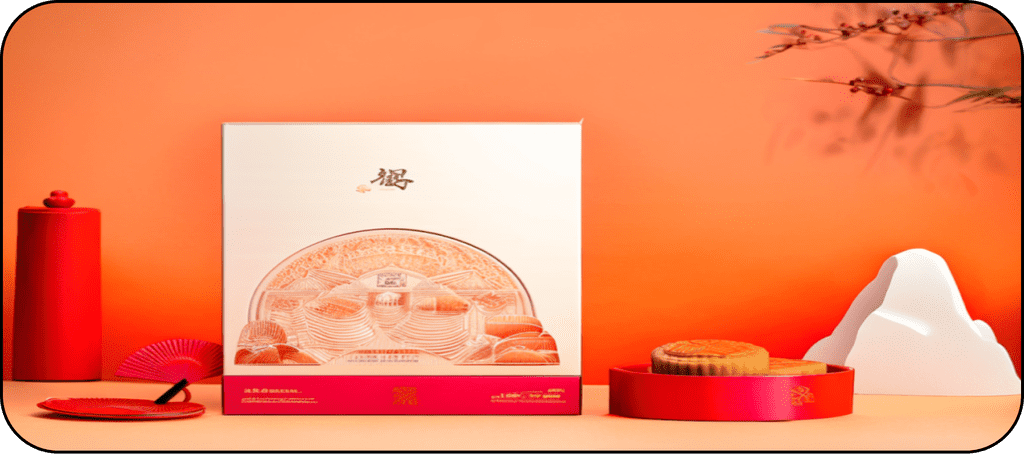} & 

A product photography image with a warm orange background. The center of the image features a white gift box with a fan-shaped cutout pattern and landscape painting decorations; three golden mooncakes are visible inside. Abstract white mountain-shaped decorations are on the right. On the left is a red rectangular object and a red round tray with a mooncake on it. The overall style of the image is simple and clear, highlighting the elegance and sophistication of the gift box. The color scheme is harmonious, creating a comfortable visual experience.
\\

\end{longtable}
}

\clearpage
\newpage

{\tiny
\begin{longtable}{@{}m{0.15\textwidth} m{0.3\textwidth} | m{0.15\textwidth} m{0.3\textwidth}@{}} 
\caption{Textual descriptions of the images in Figure 1.2.}\label{appendix_tab2}\\
\toprule
\textbf{Image} & \textbf{Description} & \textbf{Image} & \textbf{Description} \\ 
\midrule
\endhead

\multicolumn{4}{r}{\emph{Continued on next page}} \\ 
\endfoot

\bottomrule
\endlastfoot

\centering
\includegraphics[width=0.8\linewidth, height=3cm, keepaspectratio]{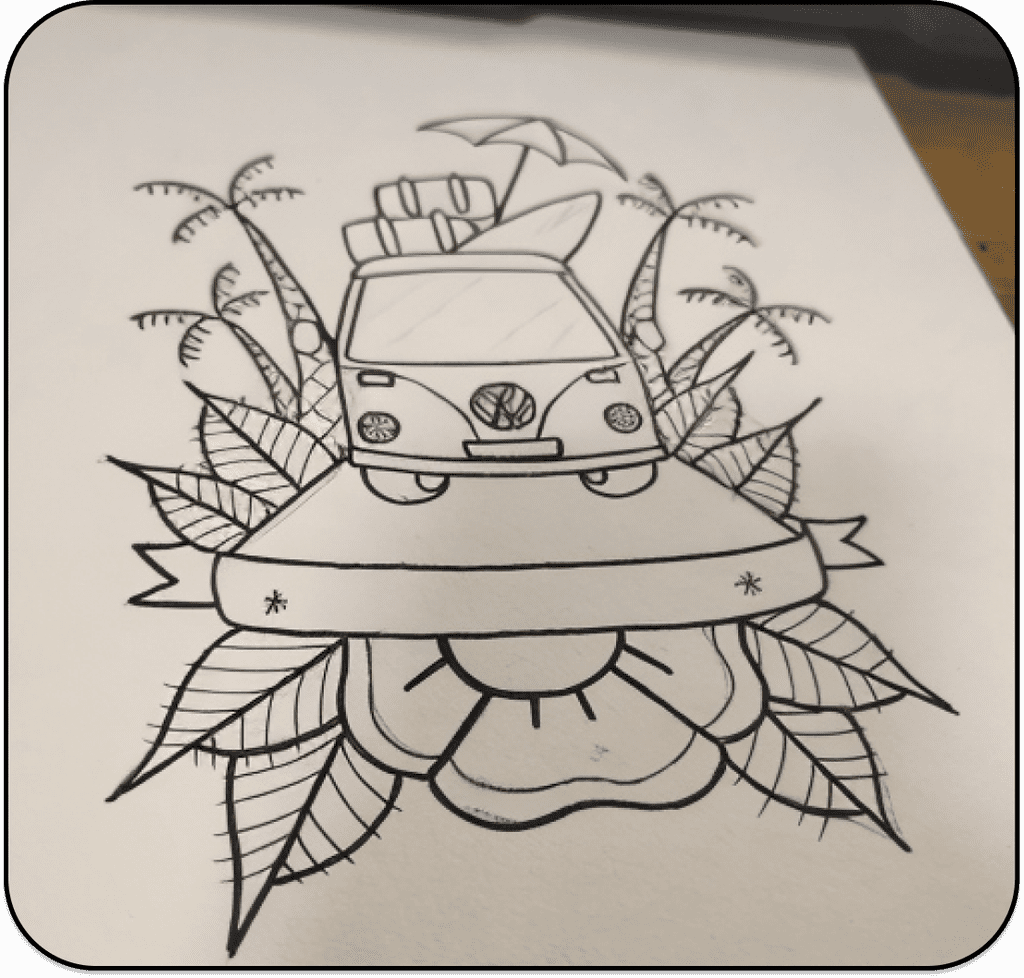} & 

This painting depicts a retro Volkswagen van with luggage and surfboards loaded on the roof. In front of the van are two lush palm trees, creating a strong summer beach vacation atmosphere. Below the van, there is a blank banner with a budding rose flower below, surrounded by branches and leaves. The entire painting is outlined with clear lines, contrasting black and white, and has a minimalist style. &
\centering 
\includegraphics[width=0.8\linewidth, height=3cm, keepaspectratio]{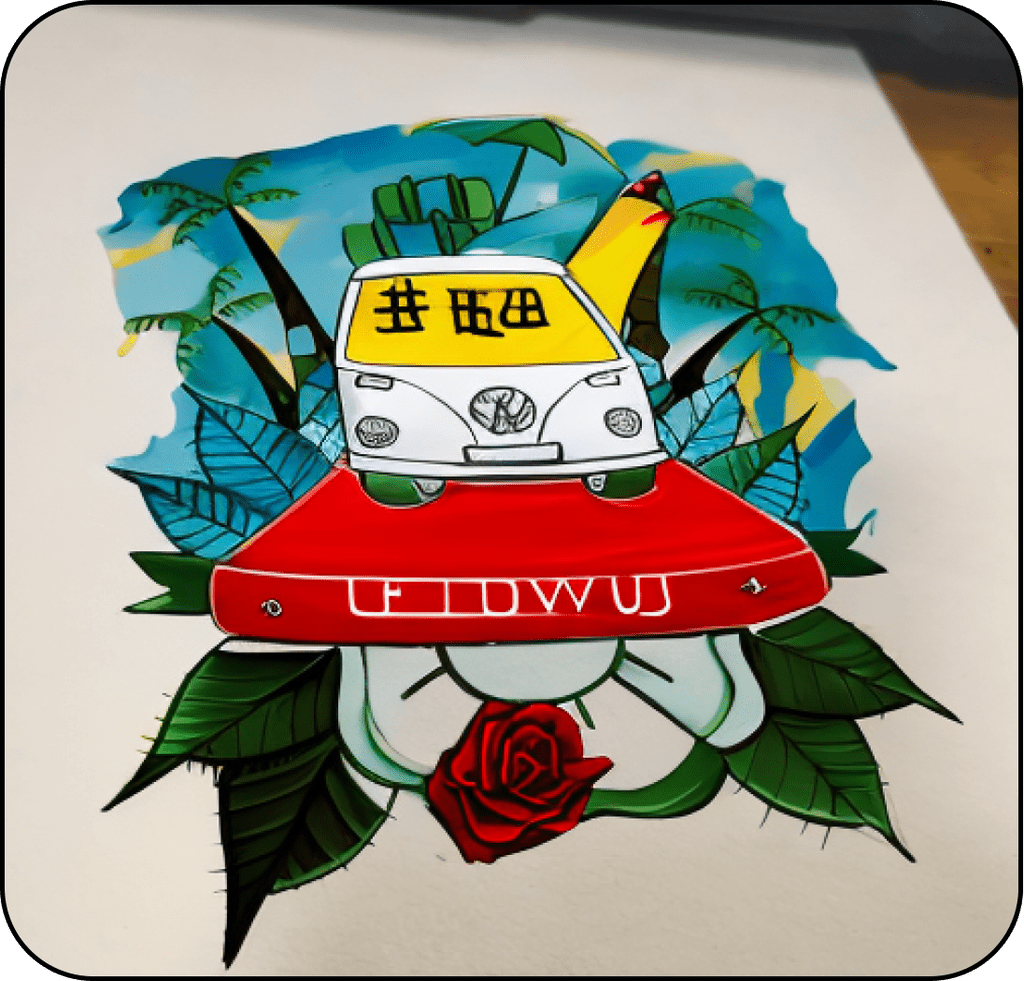} & 
This painting depicts a red and white vintage Volkswagen van with luggage and surfboards loaded on its roof. In front of the van are two lush palm trees, creating a strong summer beach vacation atmosphere. Below the van is a yellow banner that reads ``TRAVEL WITH YOU". Below the banner is a blooming red rose surrounded by branches and leaves. The painting is vibrant and full of energy. 
\\

\midrule
\centering
\includegraphics[width=\linewidth, height=3cm, keepaspectratio]{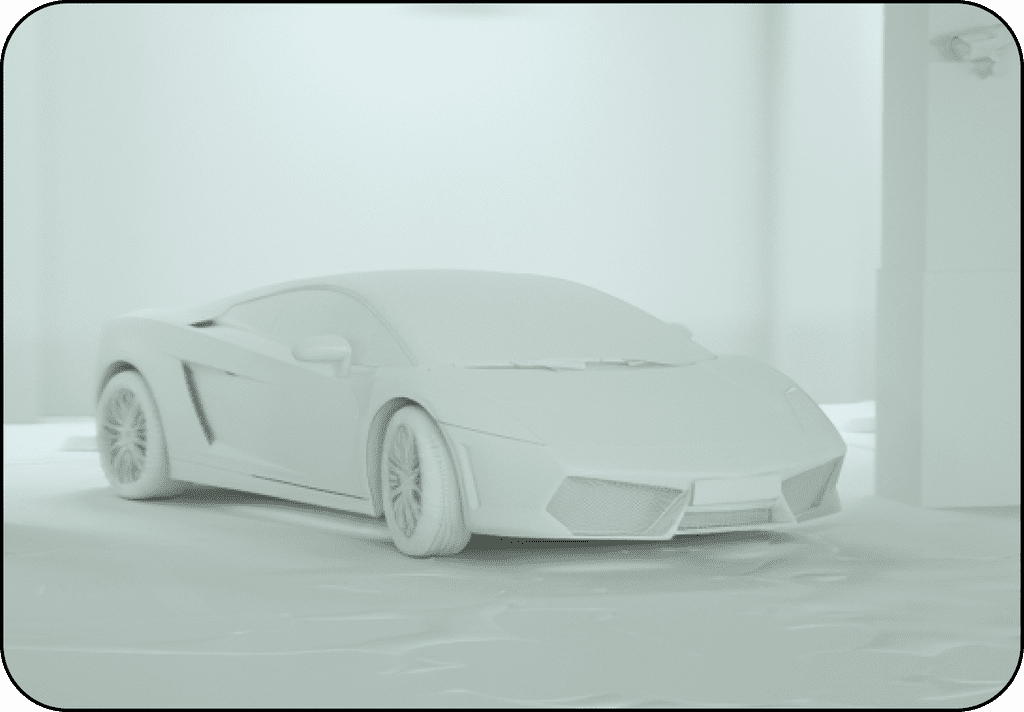} & 

This is a photo of an underground parking garage with a white Lamborghini sports car parked. The sports car has smooth lines and a dynamic shape, and its body is pure white, which is particularly striking under the light. The floor of the parking lot is smooth and reflects the light.
& 

\centering
\includegraphics[width=\linewidth, height=3cm, keepaspectratio]{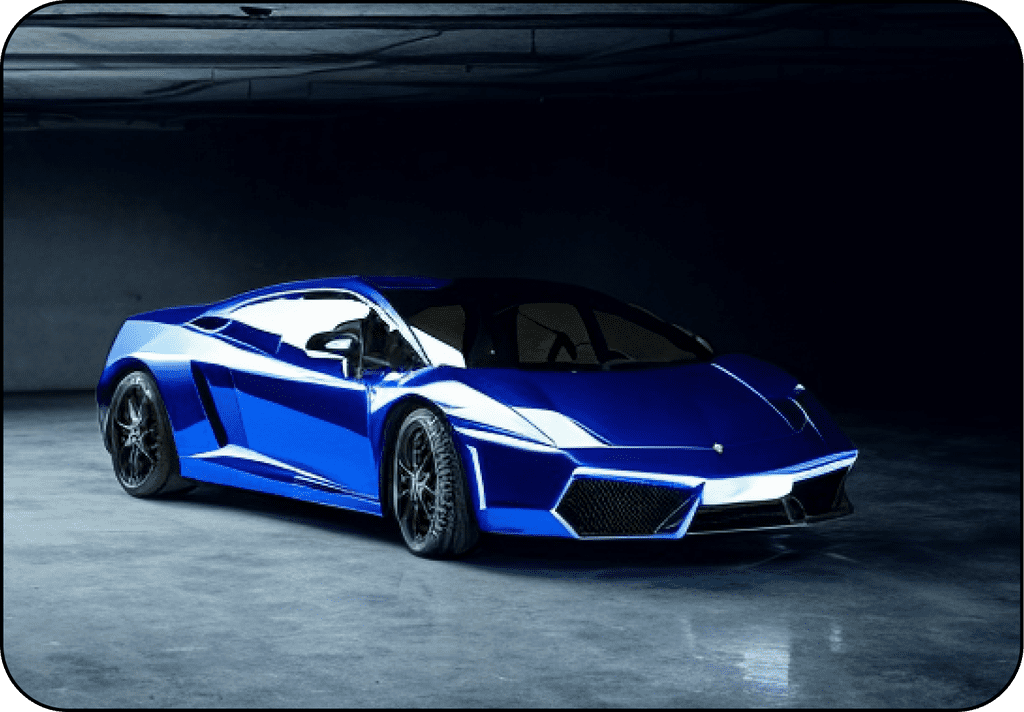} & 

This is a photo of an underground parking garage with a blue Lamborghini sports car parked. The sports car has smooth lines and a dynamic shape, and its body is dark blue, shining with a metallic luster under the light. The floor of the parking lot is smooth and reflects the light.
\\

\midrule
\centering
\includegraphics[width=0.6\linewidth, height=3cm, keepaspectratio]{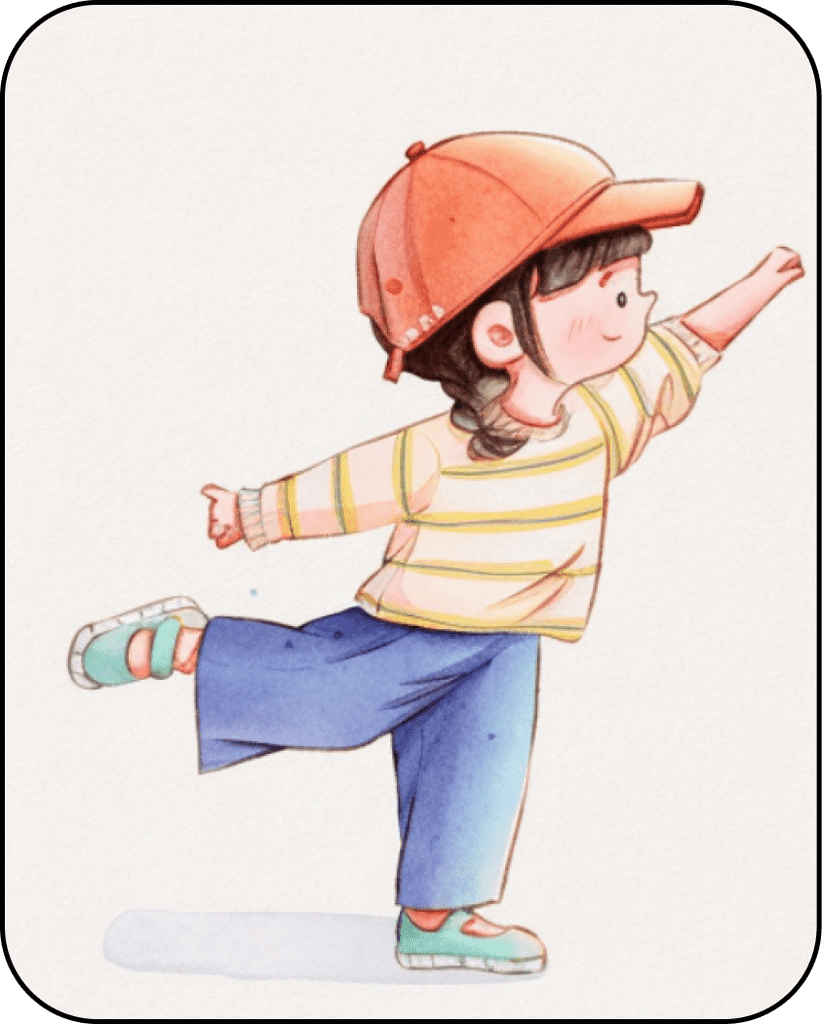} & 

In the picture, a little girl wearing a yellow and white striped shirt and blue pants is dancing happily. She wears an orange baseball cap, with her hair tied in a cute side ponytail. She stands on her toes with her left foot, her right leg extended backward, her left arm naturally bent, and her right arm raised high as if trying to touch the sky, a joyful smile on her face.
& 

\centering
\includegraphics[width=0.6\linewidth, height=3cm, keepaspectratio]{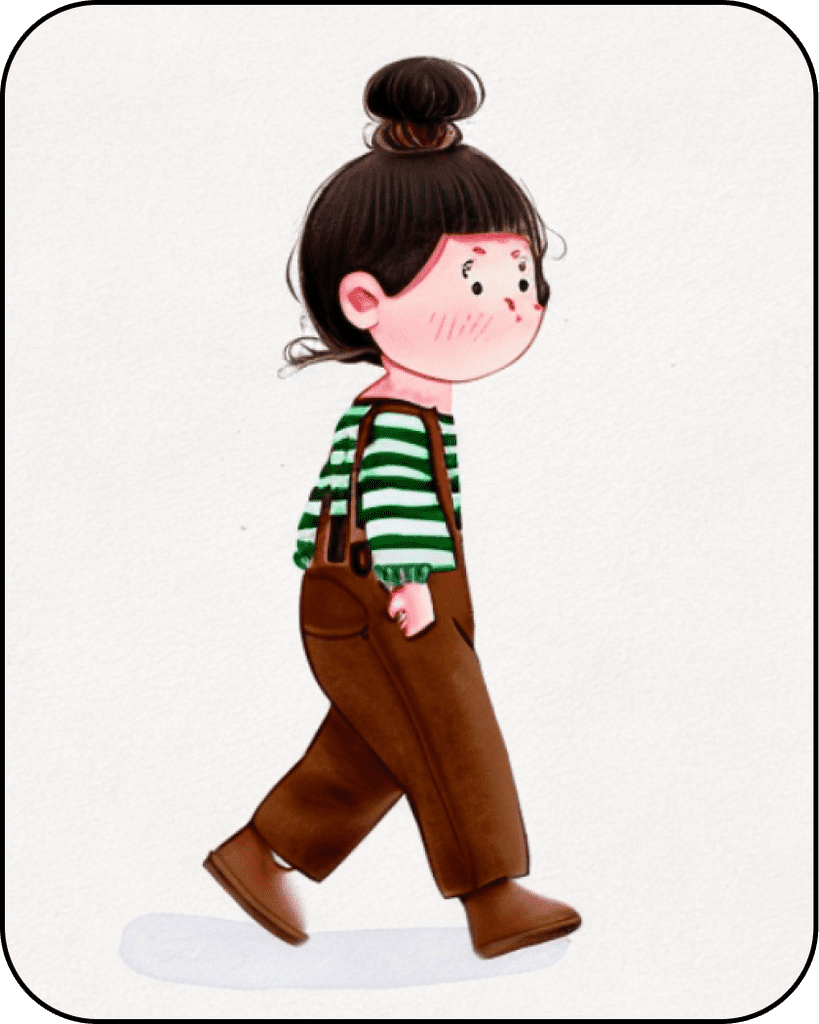} & 

In the picture, a little girl wearing a green and black striped shirt and brown overalls is taking a leisurely stroll. Her long black hair is tied up in a high bun, with a few strands casually falling down. She wears a pair of brown ankle boots, giving her a stylish and playful look. She walks with a brisk pace, slightly turning her body to the side, her eyes curiously gazing into the distance, as if filled with interest in everything around her.
\\

\midrule
\centering
\includegraphics[width=0.8\linewidth, height=3cm, keepaspectratio]{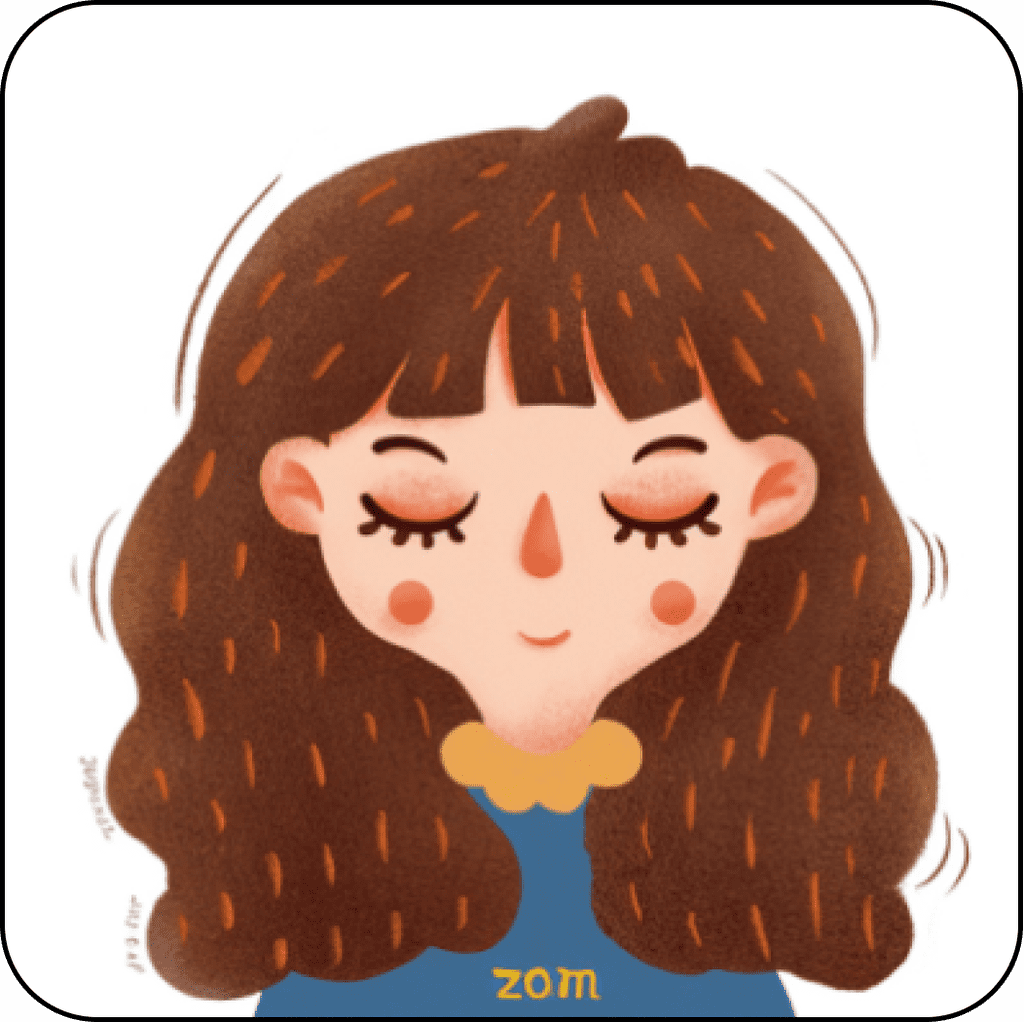} & 
The central focus is a close-up of a girl's head and shoulders. She has shoulder-length, slightly wavy brown hair with bangs covering her eyebrows, her eyes are closed, and her expression is calm and serene, with rosy cheeks.  She wears a dark blue top with light yellow accents at the collar. The overall style is soft, with warm colors, smooth lines, and visible brushstrokes. The background is pure white, highlighting the main subject. A simple ``zom" is printed at the bottom.
& 

\centering
\includegraphics[width=0.8\linewidth, height=3cm, keepaspectratio]{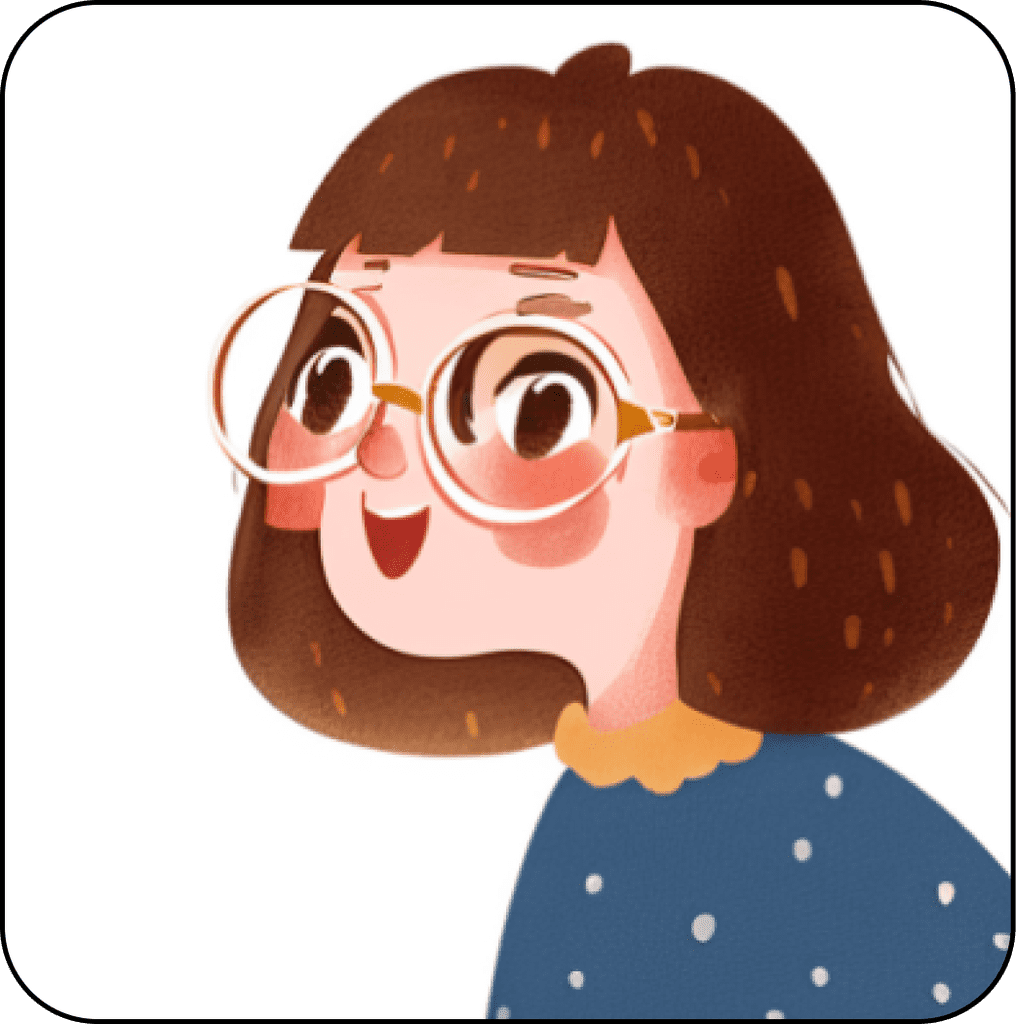} & 

The central focus is a close-up of a girl's head and shoulders. She has shoulder-length, slightly wavy brown hair with bangs covering her eyebrows and is wearing round gold-rimmed glasses. Her eyes are wide open, she is smiling, and her cheeks are rosy. She wears a dark blue top with white polka dots and light yellow accents at the collar. The overall style is soft, with warm colors, smooth lines, and visible brushstrokes. The background is pure white, highlighting the main subject.
\\

\midrule
\centering
\includegraphics[width=0.6\linewidth, height=3cm, keepaspectratio]{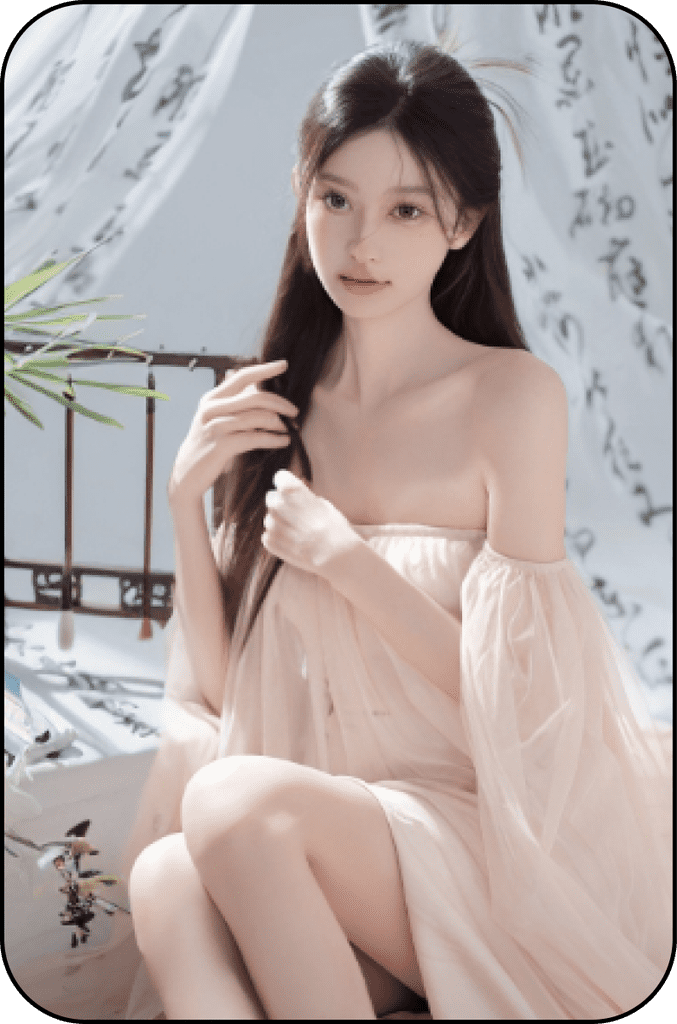} & 
A young woman with long, dark hair cascading over her shoulders wears a light-colored, off-the-shoulder dress, her delicate features clearly visible. She sits on a bamboo mat, both hands gently touching her hair, her gaze soft as she looks directly at the camera. 
& 

\centering
\includegraphics[width=0.6\linewidth, height=3cm, keepaspectratio]{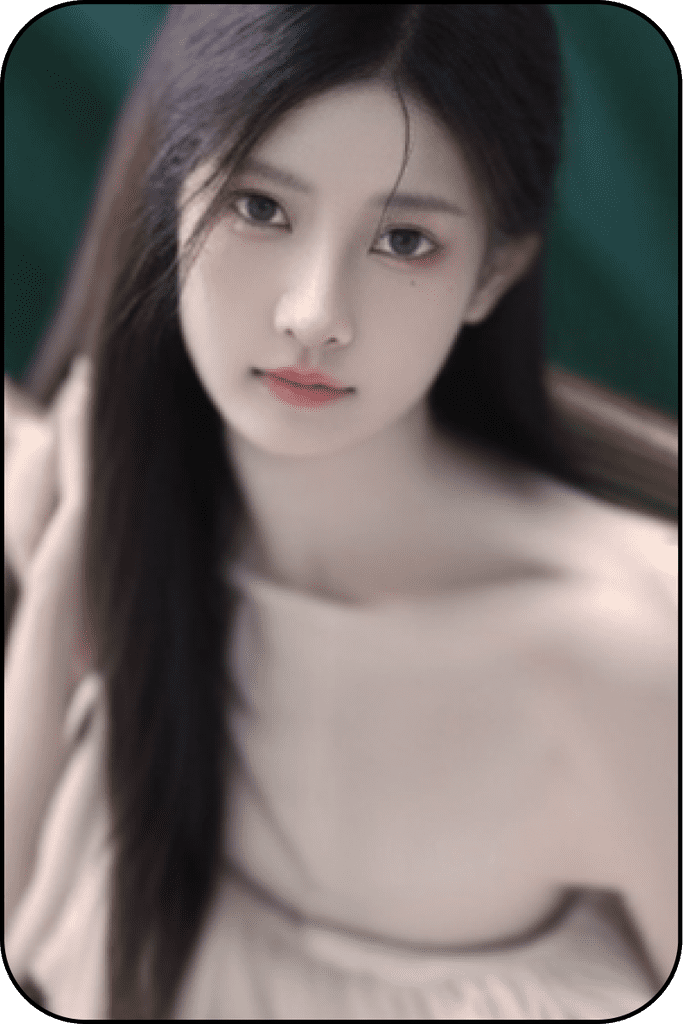} & 

A young woman with long, dark hair cascading over hers shoulders wears a light-colored, off-the-shoulder dress, her delicate features clearly visible. She sits on a bamboo mat, her right hand gently touching her hair, her gaze soft as she looks directly at the camera. 
\\

\midrule
\centering
\includegraphics[width=0.8\linewidth, height=3cm, keepaspectratio]{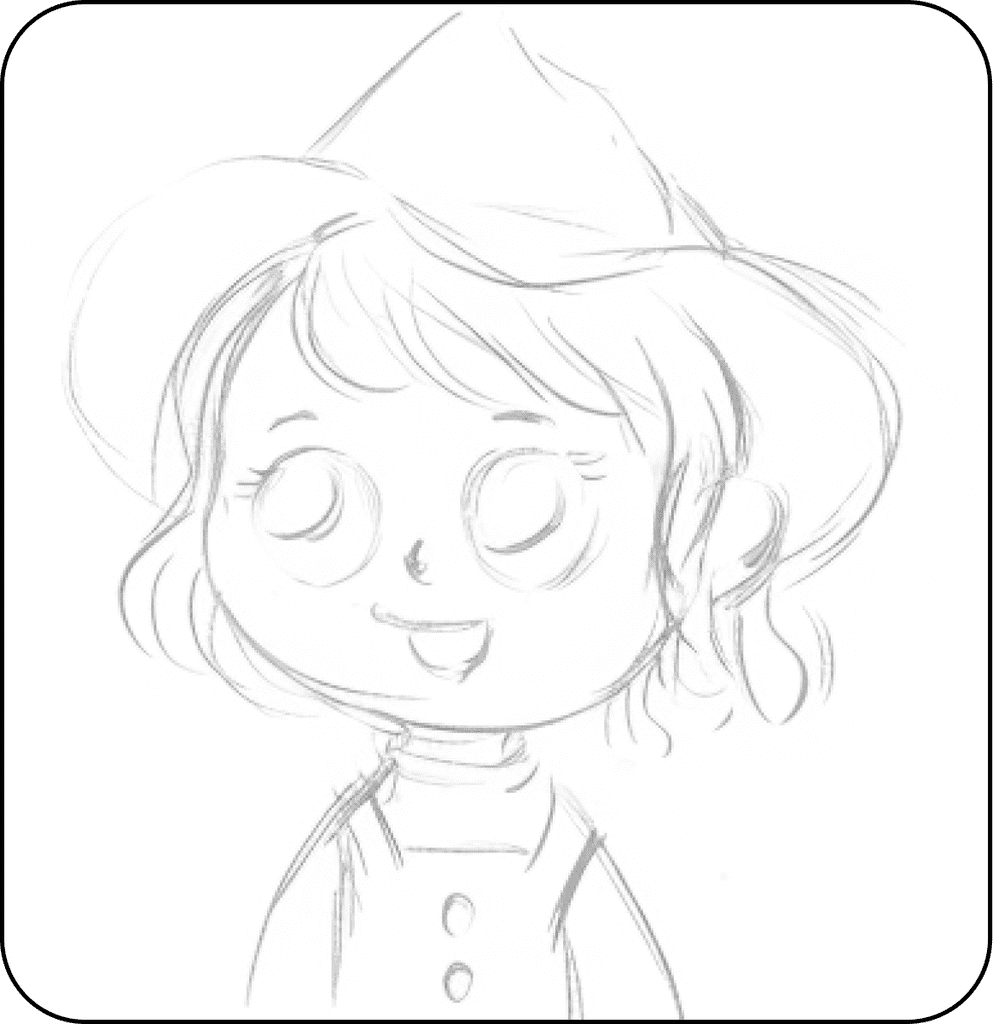} & 
A cartoon-style portrait of a little girl, drawn in pencil sketch style. She is wearing a pointed hat and a pair of bib pants with two buttons. The little girl has short hair to her shoulders, her eyes closed in two curved arcs, her mouth slightly open with a happy smile.
& 

\centering
\includegraphics[width=0.8\linewidth, height=3cm, keepaspectratio]{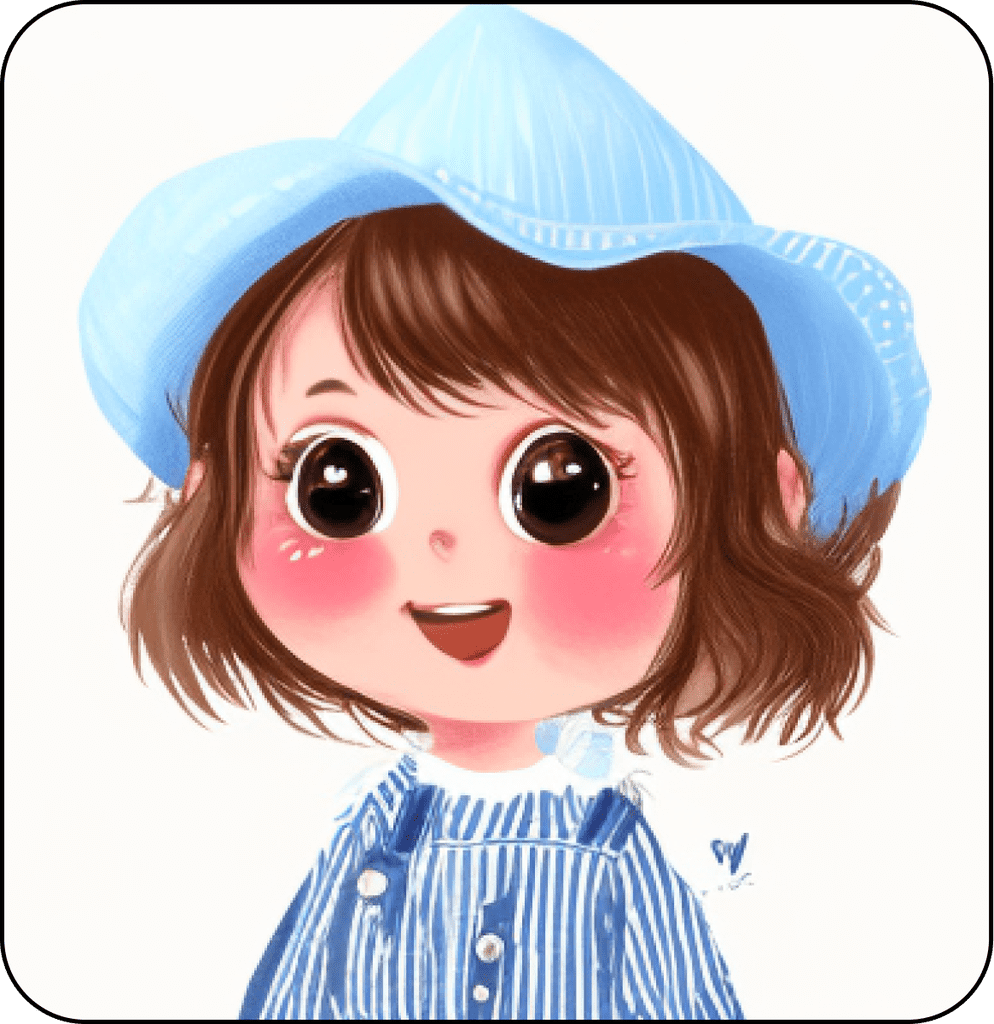} & 

A cartoon-style portrait of a little girl, colorfully drawn.  She is wearing a sky blue pointed hat with a circle of white stripes. The little girl is wearing a black and white stripped top with sky blue overalls decorated with two buttons. The little girl has short hair to her shoulders, thick brown hair naturally hanging down. The little girl has two big eyes, her mouth slightly open with a happy smile. 
\\

\midrule
\centering
\includegraphics[width=0.8\linewidth, height=3cm, keepaspectratio]{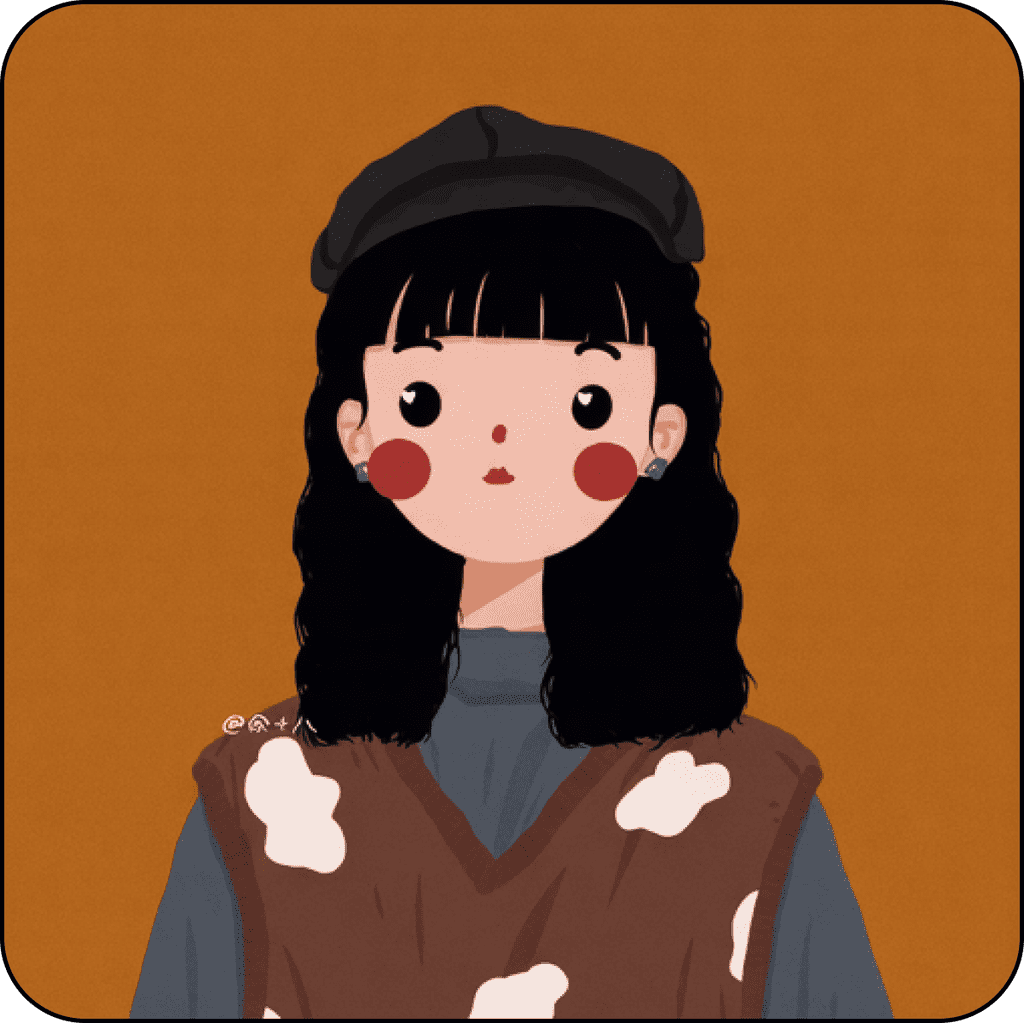} & 
A girl with black bob hair and bangs is wearing a black beret. She has large eyes, a round face, rosy cheeks, and is wearing silver earrings. She is wearing a brown vest with white cloud patterns over a blue turtleneck. 
& 

\centering
\includegraphics[width=0.8\linewidth, height=3cm, keepaspectratio]{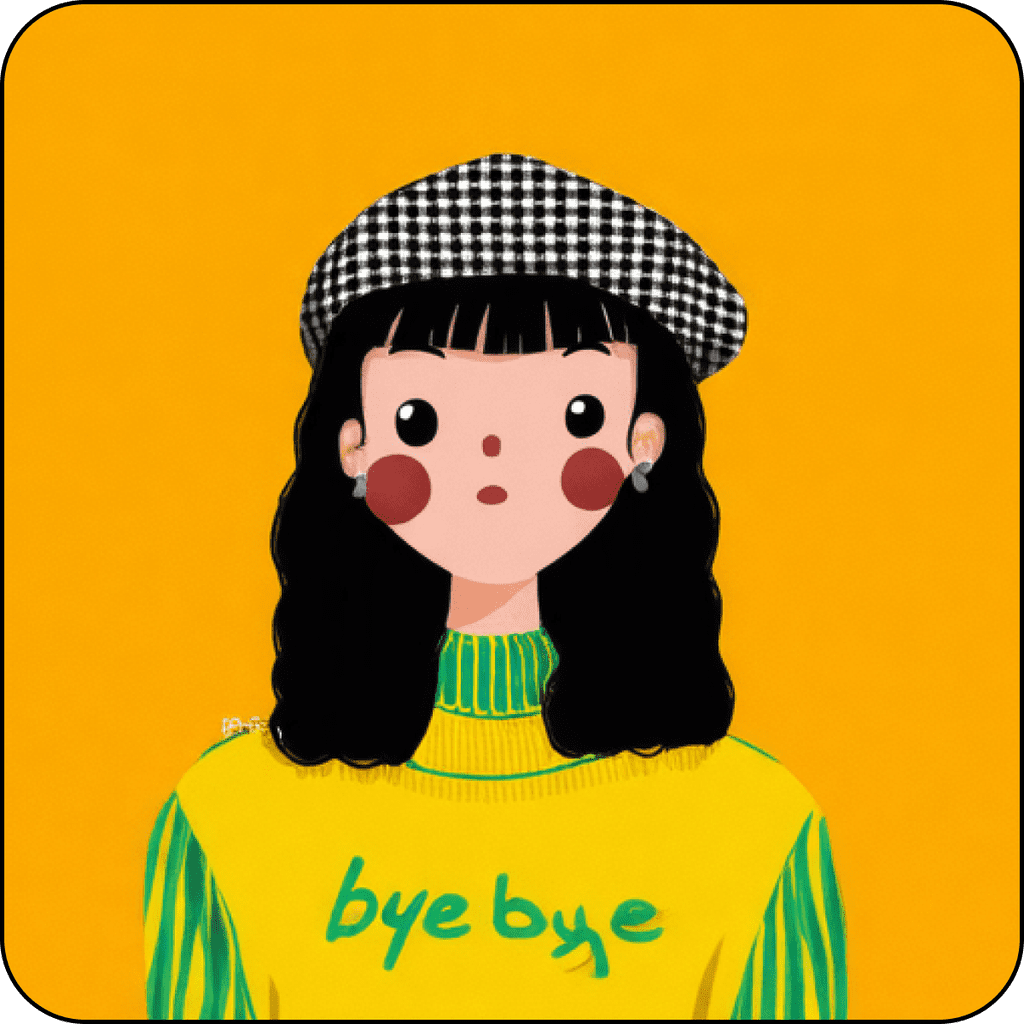} & 

A girl with black bob hair and bangs is wearing a black checkered beret. She has large eyes, a round face, rosy cheeks, and is wearing gold earrings. She is wearing a yellow sweater with the words "byebye" printed on it and green striped collar and cuffs. 
\\

\midrule
\centering
\includegraphics[width=\linewidth, height=3cm, keepaspectratio]{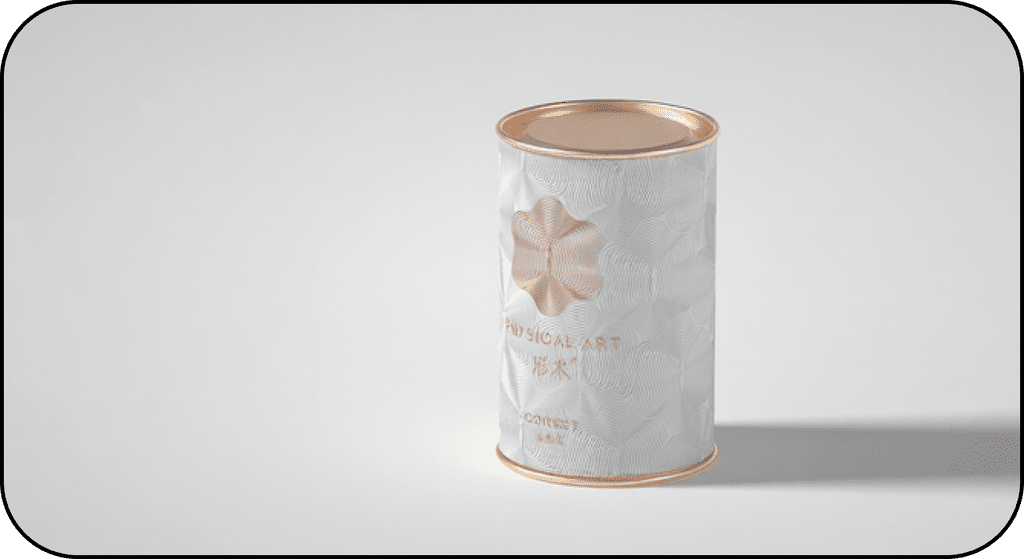} & \begin{CJK*}{UTF8}{gbsn}A product display image with a pure white background. A cylindrical metal can is placed in the center. The can is white with a delicate textured pattern on its surface, resembling clouds or petals, with a golden metallic sheen. The lid is made of rose gold metal. The center of the can features a golden brand logo, including the brand name ``PHYSICAL ART" and the Chinese brand name ``悠然". The bottom of the can is marked with ``CONCERT 45ML". The overall style is simple, elegant, and refined.
\end{CJK*}
& 

\centering
\includegraphics[width=\linewidth, height=3cm, keepaspectratio]{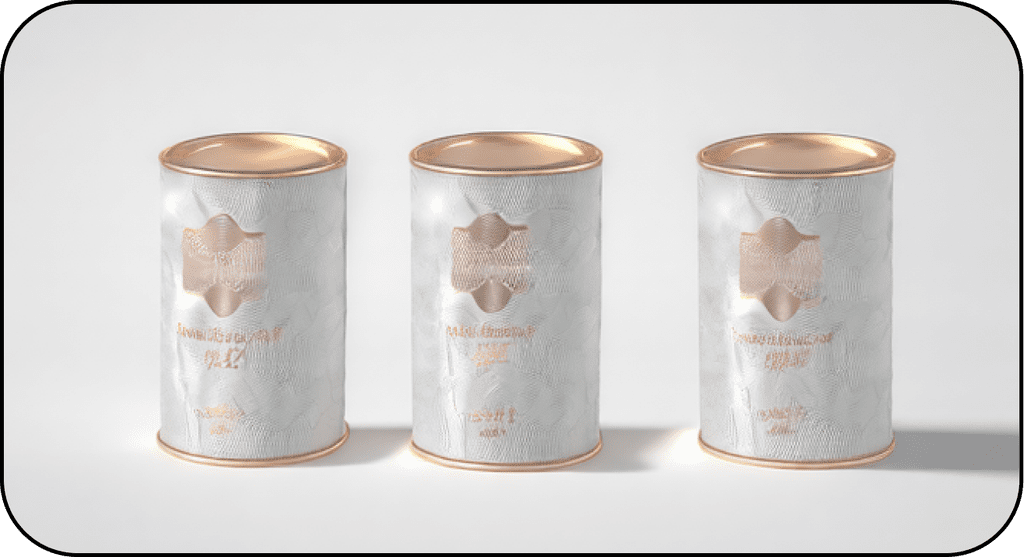} & 
\begin{CJK*}{UTF8}{gbsn}A product display image with a pure white background. Three identical cylindrical metal cans are arranged side by side. The cans are white with a delicate textured pattern on their surface, resembling clouds or petals, with a golden metallic sheen. The lids are made of rose gold metal. The center of each can features a golden brand logo, including the brand name ``PHYSICAL ART" and the Chinese brand name ``悠然". The bottom of each can is marked with ``CONCERT 45ML". The overall style is simple, elegant, and refined.
\end{CJK*}
\\

\midrule
\centering
\includegraphics[width=0.6\linewidth, height=3cm, keepaspectratio]{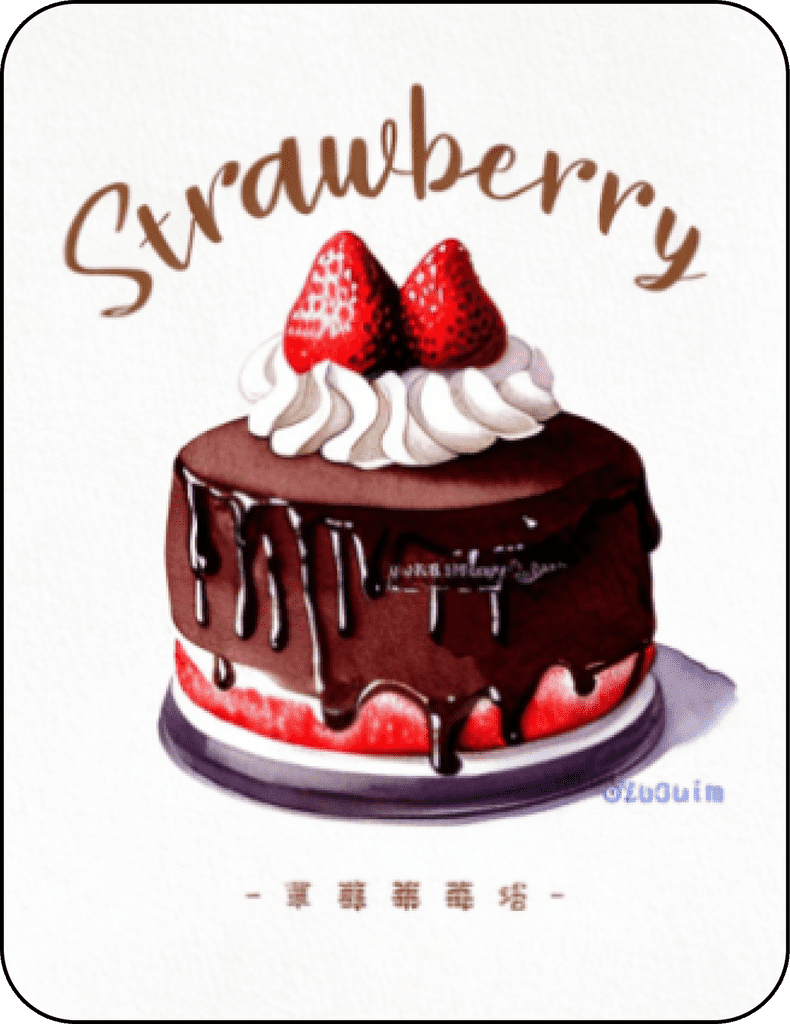} & 
The center of the image is a chocolate cake. The bottom of the cake is red, the top is covered with thick chocolate sauce, and it's decorated with fluffy cream and two fresh strawberries. The cake rests on a dark base. Above the image is the title ``Strawberry", with an elegant and flowing font. The background is simple, the overall style is fresh and sweet, creating a comfortable visual experience. The watercolor painting technique gives the image a soft color transition and a light texture.
& 

\centering
\includegraphics[width=0.6\linewidth, height=3cm, keepaspectratio]{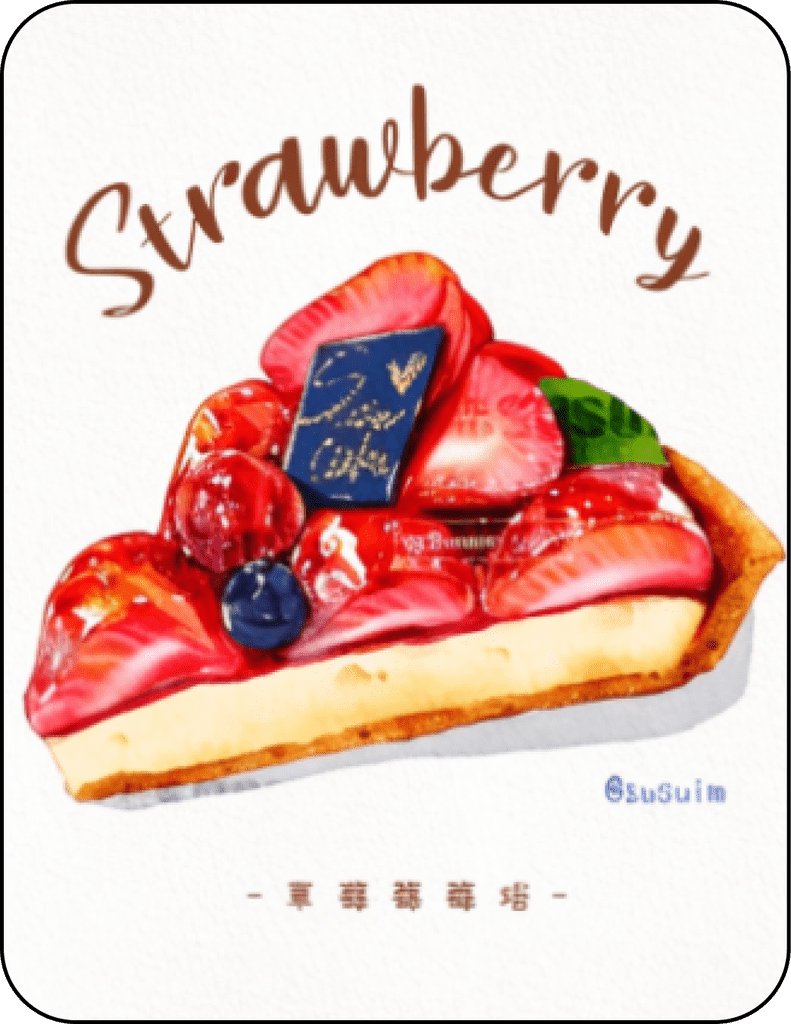} & 
The main subject of the image is a slice of strawberry tart. The tart crust is golden yellow, and it's topped with bright red strawberries, decorated with a few blueberries and cherries. There's a golden chocolate decoration on the strawberries. The cross-section of the tart shows a rich layering and the texture of the filling. Above the image is the title \"Strawberry\", consistent with the first image. The background is equally simple, and the watercolor painting technique creates a light and dreamy atmosphere.
\\
\midrule

\centering
\includegraphics[width=0.6\linewidth, height=3cm, keepaspectratio]{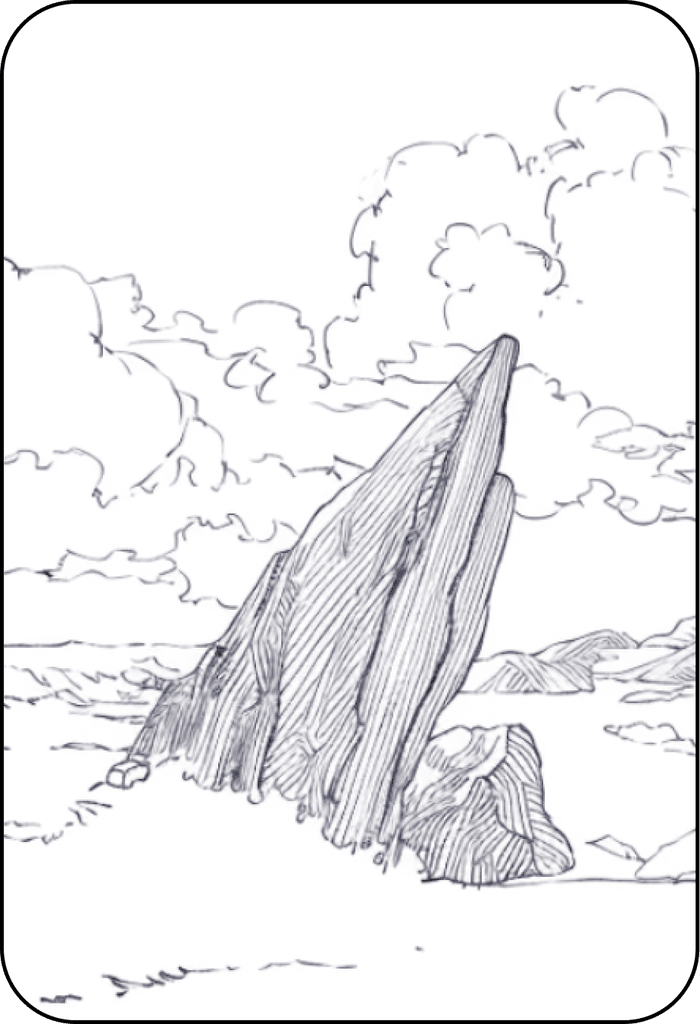} & 
This illustration depicts a huge rock standing on the coast in a pen-and-ink sketch style. The surface of the rock is rough, with rich texture details and smooth and natural lines. Behind the rock is the rough sea, and the rolling mountains can be vaguely seen in the distance. The sky is covered with clouds of various shapes.
& 

\centering
\includegraphics[width=0.6\linewidth, height=3cm, keepaspectratio]{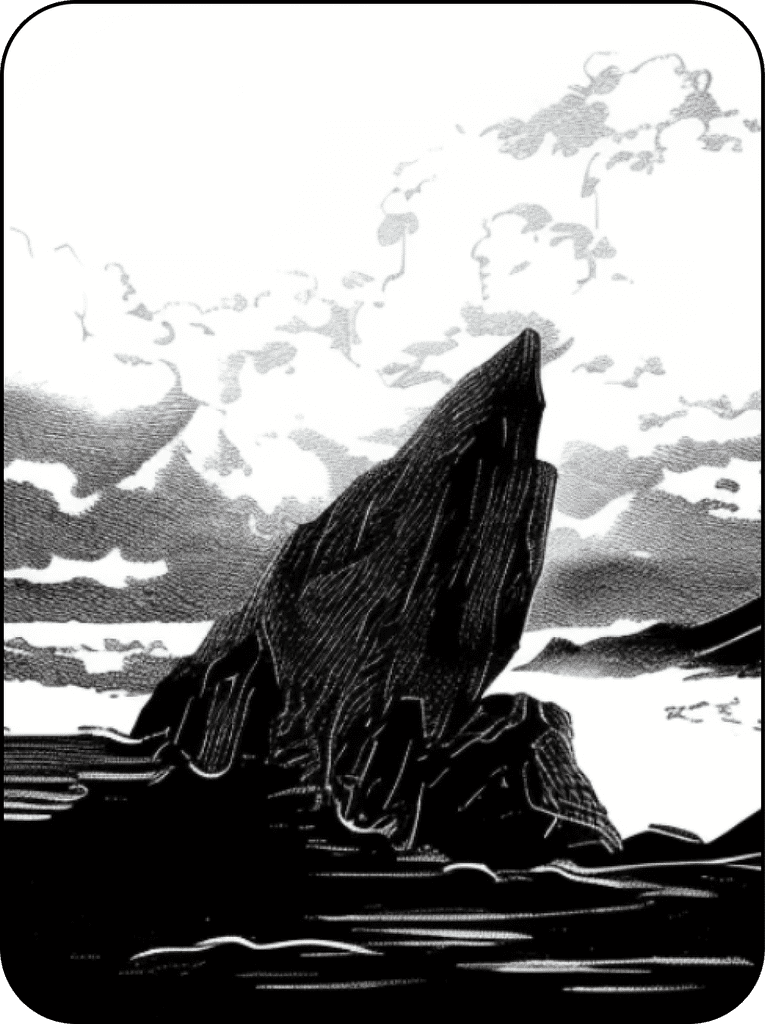} & 
This illustration depicts a huge rock standing on the coast in the style of an engraving. The surface of the rock is rough, with rich texture details, and the use of dense lines to depict the effect of light and shadow. The bottom of the rock is beaten by the surging waves, and the the spray is splashing. In the distance are rolling mountains, and the sky is filled with clouds of various shapes, with sunlight shining through the clouds.
\\
\midrule

\centering
\includegraphics[width=0.6\linewidth, height=3cm, keepaspectratio]{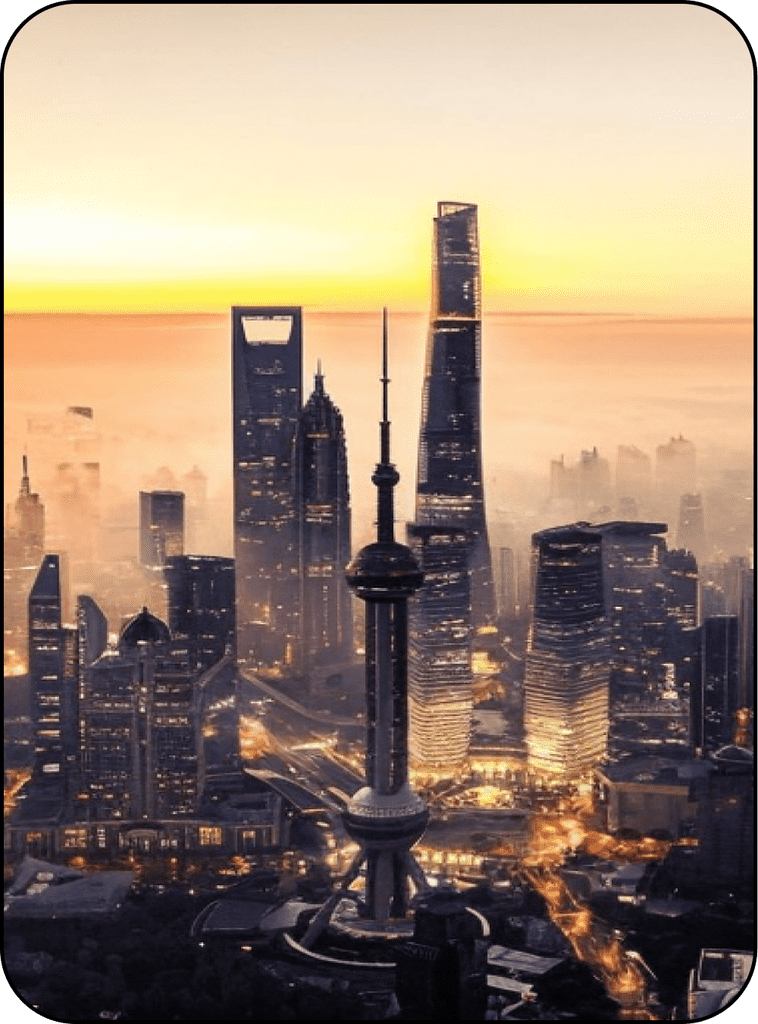} & 
This is a bustling modern city, with high-rise buildings standing tall and arranged in a staggered manner. The silhouette of the distant high-rise buildings is clearly visible under the afterglow of the sunset. The glass of some buildings reflects golden sunlight, and the streets are bustling with traffic. The sky presents a gradient from orange yellow to pink, the clouds are dyed with brilliant colors, and the city is bathed in a warm sunset.
& 

\centering
\includegraphics[width=0.6\linewidth, height=3cm, keepaspectratio]{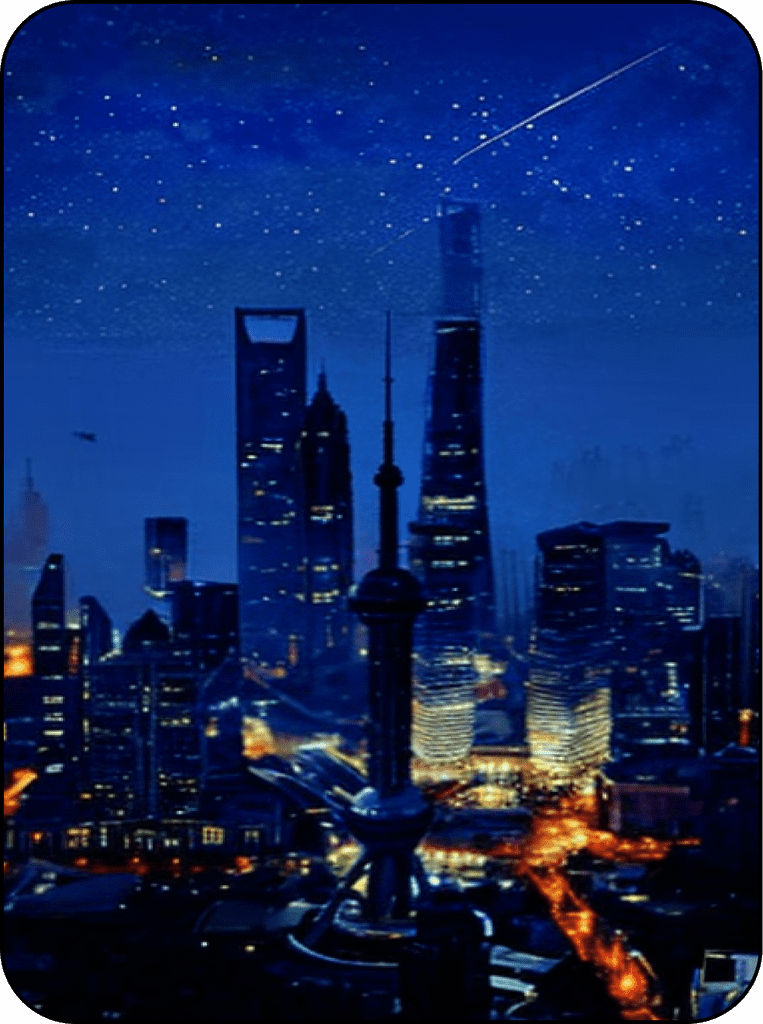} & 
This is a bustling modern city, with high-rise buildings standing tall and arranged in a staggered manner. The silhouette of distant high-rise buildings is clearly visible in the night sky. Most buildings were lit up with scattered lights, and the streets were bustling with traffic, with light trajectories crisscrossing. The sky is dotted with several bright stars, and the night sky is deep blue. The city is immersed in a peaceful and tranquil atmosphere.
\\
\midrule

\centering
\includegraphics[width=0.6\linewidth, height=3cm, keepaspectratio]{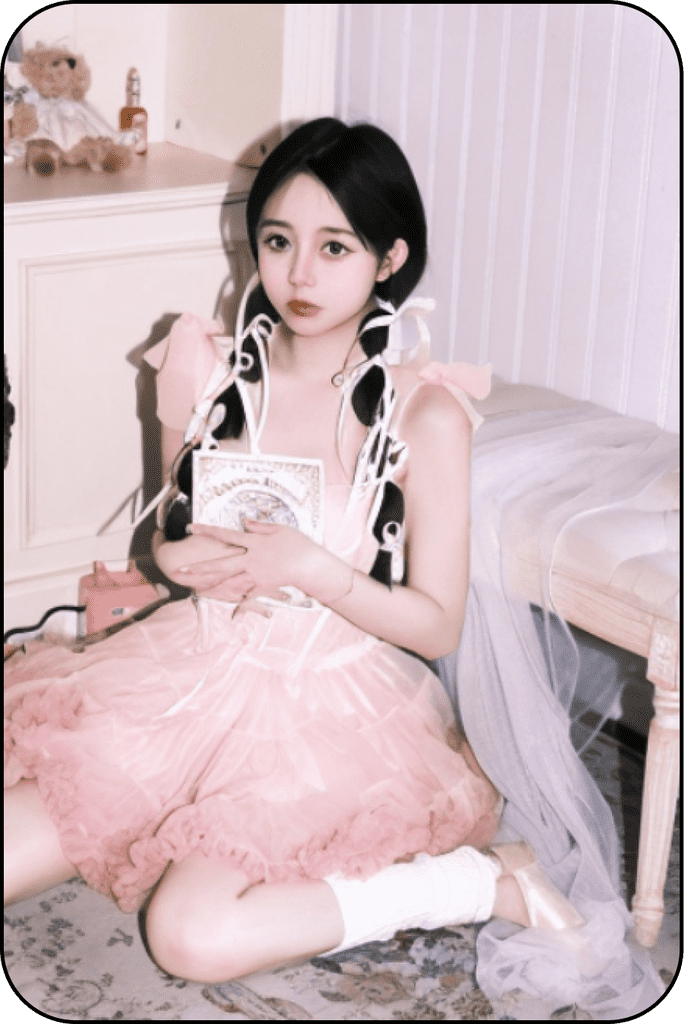} & 
A young Asian woman with long, black hair sits on the floor wearing a pink lace dress with white ribbon decorations tied in bows on her head. She holds a storybook and looks at the viewer with clear eyes. The room features white wood paneling and a white cabinet. 
& 

\centering
\includegraphics[width=0.6\linewidth, height=3cm, keepaspectratio]{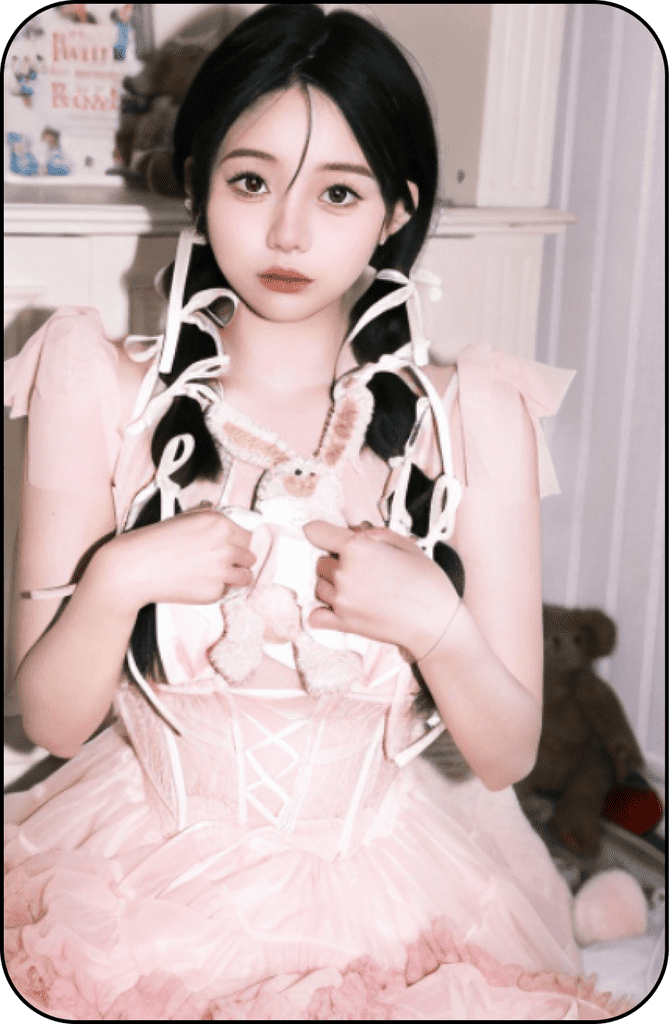} & 

A young Asian woman with long, black hair sits on the floor wearing a pink lace dress with white ribbon decorations tied in bows on her head. She embraces a plush toy and looks gently ahead. The room features white wood paneling and a white cabinet. 
\\
\midrule

\centering
\includegraphics[width=0.6\linewidth, height=3cm, keepaspectratio]{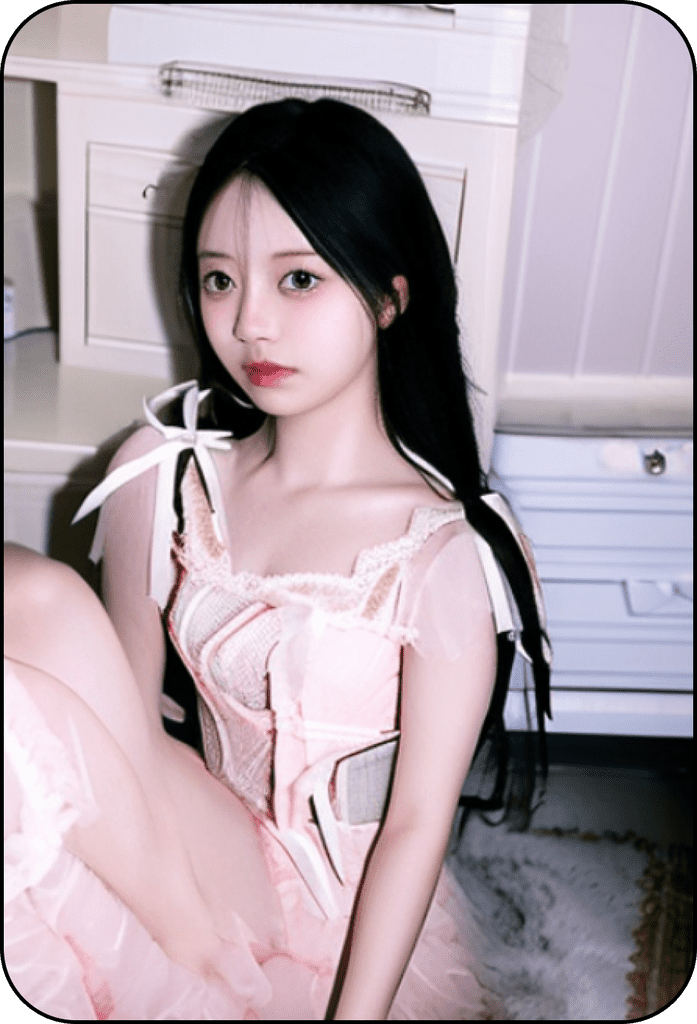} & 
A young Asian woman with long, black hair sits on the floor wearing a pink lace dress with white ribbon decorations tied in bows on her head. She rests one hand on her leg and looks directly at the viewer with clear eyes. The room features white wood paneling and a white cabinet. 
& 

\centering
\includegraphics[width=0.6\linewidth, height=3cm, keepaspectratio]{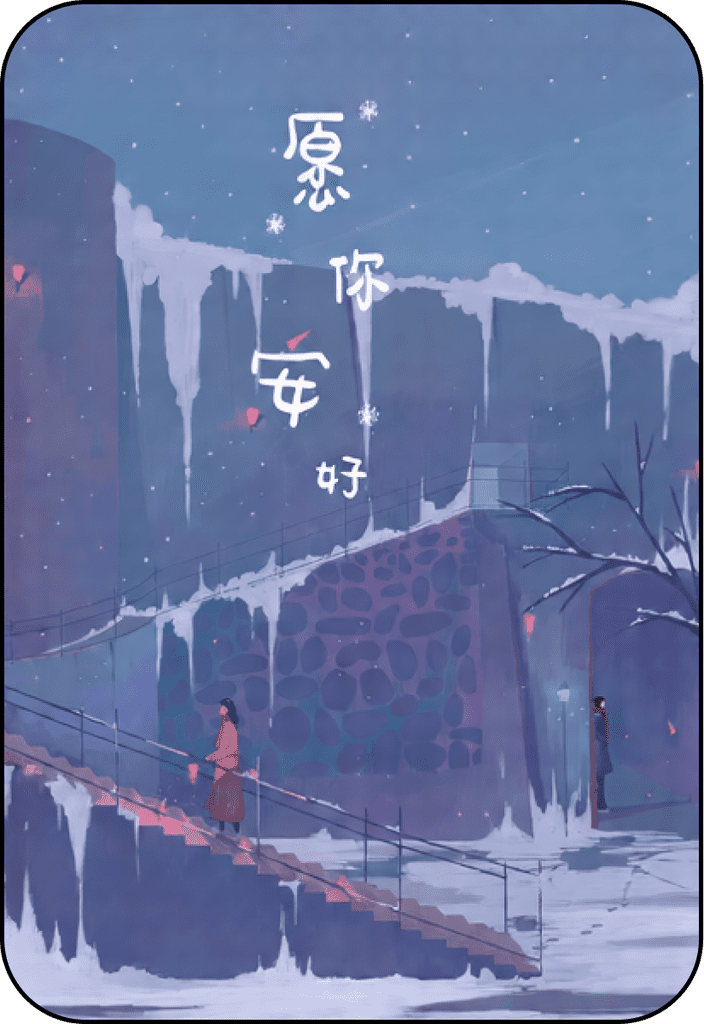} & \begin{CJK*}{UTF8}{gbsn}The scene presents a tranquil winter nightscape. Snow-laden stone buildings stand tall on either side, with dense icicles hanging from their roofs. A woman in a reddish-brown coat ascends a stone staircase, carrying a warm lantern.  Dim yellowish lights emanate from the buildings, contrasting with the twinkling stars in the sky. Fine snowflakes fall from the sky, and a thin layer of snow covers the ground. The overall color palette is cool, creating a serene and peaceful atmosphere. In the distance, another figure can be vaguely seen standing at a building entrance. A lone, bare tree is visible in the lower right corner. Vertical Chinese characters reading ``愿你安好" are displayed at the top.
\end{CJK*}
\\
\midrule

\centering
\includegraphics[width=0.6\linewidth, height=3cm, keepaspectratio]{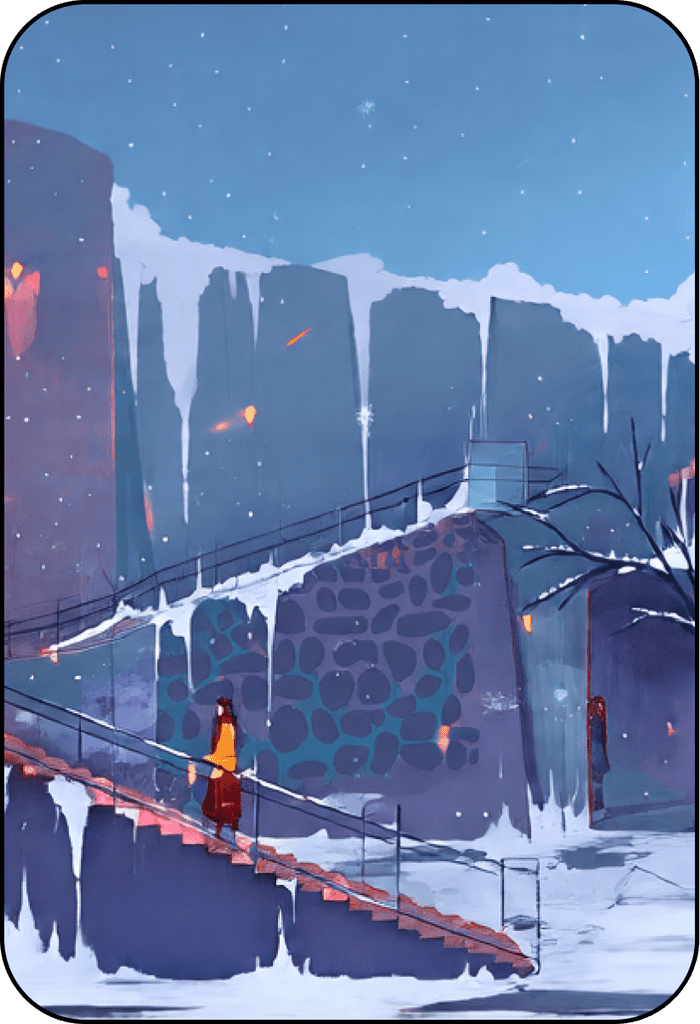} & 
The image depicts a cold winter night scene. Tall buildings on both sides are covered with thick snow and long icicles. A stone staircase winds upwards, with a figure in an orange-yellow coat walking along it. The buildings' lights are warm and yellowish, contrasting with the sparse stars in the night sky. Tiny snowflakes fall from the sky, and the ground is covered in snow. The color palette is cool, but the lights provide a touch of warmth, creating a quiet and slightly mysterious atmosphere.  A bare tree and a partial view of the building interiors are visible in the background. The overall style is dreamy and slightly impressionistic.
& 

\centering
\includegraphics[width=0.8\linewidth, height=3cm, keepaspectratio]{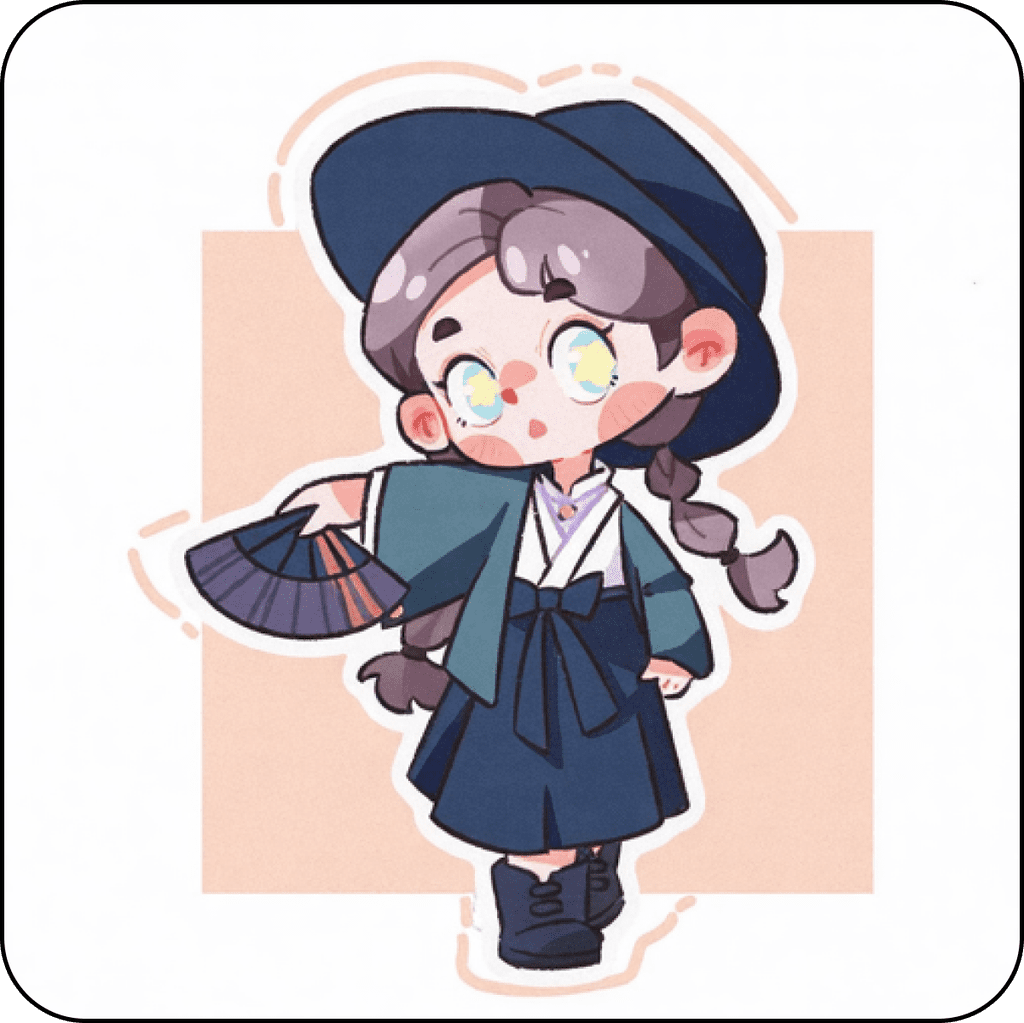} & 

The image presents an adorable chibi girl dressed in dark blue clothing. She wears a dark blue hat, a dark blue jacket, a white top underneath, a dark blue skirt, and black ankle boots. She has two braided pigtails and holds a dark purple folding fan. Her eyes are clear and bright, and her overall style is fashionable with traditional elements. The background is light orange with small dots around the edges, creating a fresh and cute style.
\\
\midrule

\centering
\includegraphics[width=0.8\linewidth, height=3cm, keepaspectratio]{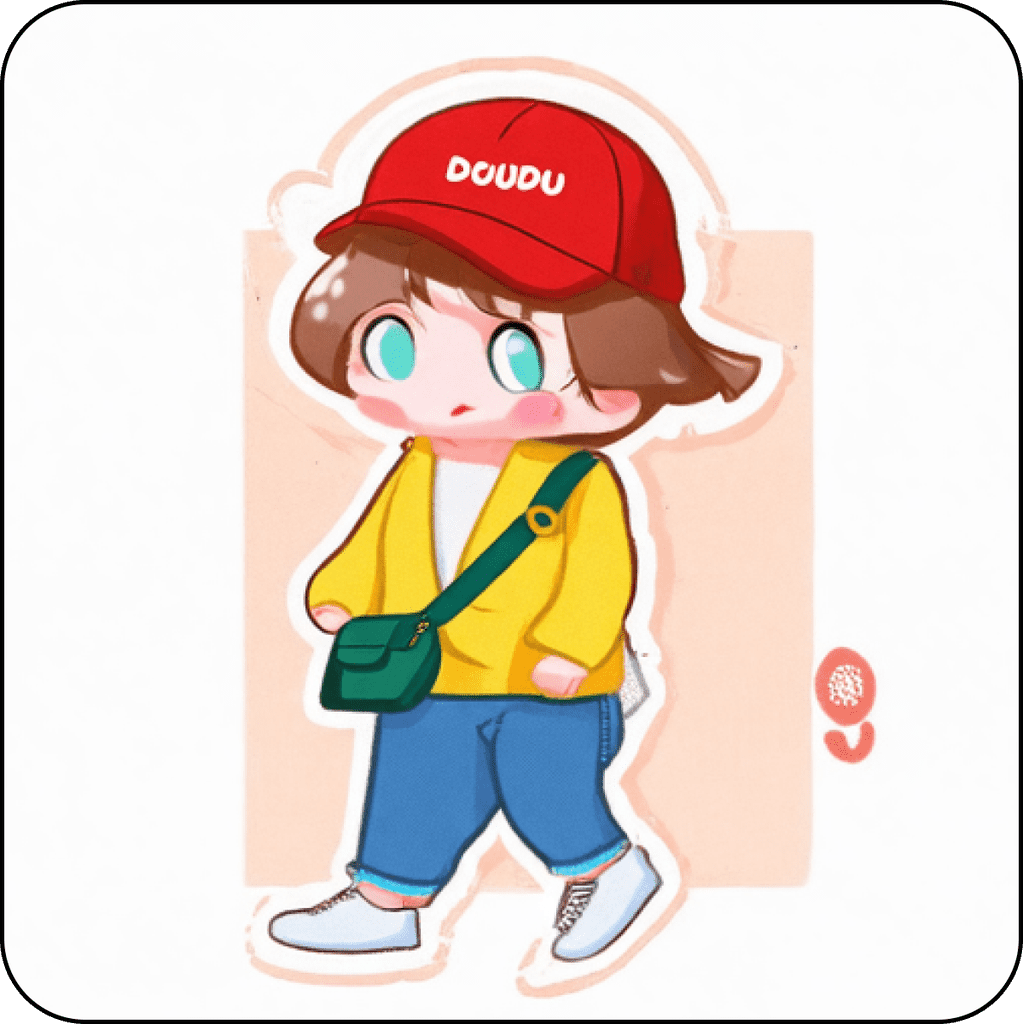} & 
The image shows an adorable chibi girl in casual attire. She sports a red baseball cap with ``DOUDU" printed on it, a yellow jacket over a white top, blue jeans, and white sneakers, along with a dark green crossbody bag. She has shoulder-length brown hair, slightly rosy cheeks, and a somewhat shy expression, creating a youthful and lively overall style. The background is light orange with small dots around the edges, maintaining a fresh and cute style.
& 

\centering
\includegraphics[width=1.0\linewidth, height=3cm, keepaspectratio]{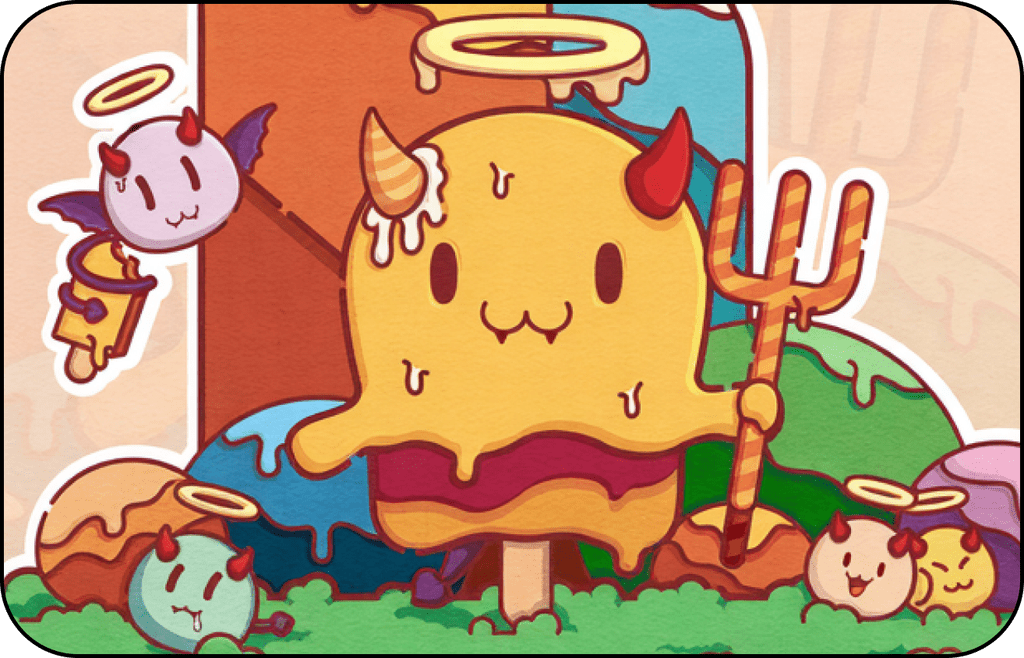} & 

The center of the image is a yellow, melting-icecream-like character with devil horns, holding a trident. It's surrounded by various colored ice creams and many small creatures wearing tiny halos or having little devil wings. The background is a soft orange and pink, creating an overall cute and playful style. The ice creams are richly colored, detailed, and present a sweet and dreamy atmosphere.
\\
\midrule

\centering
\includegraphics[width=0.6\linewidth, height=3cm, keepaspectratio]{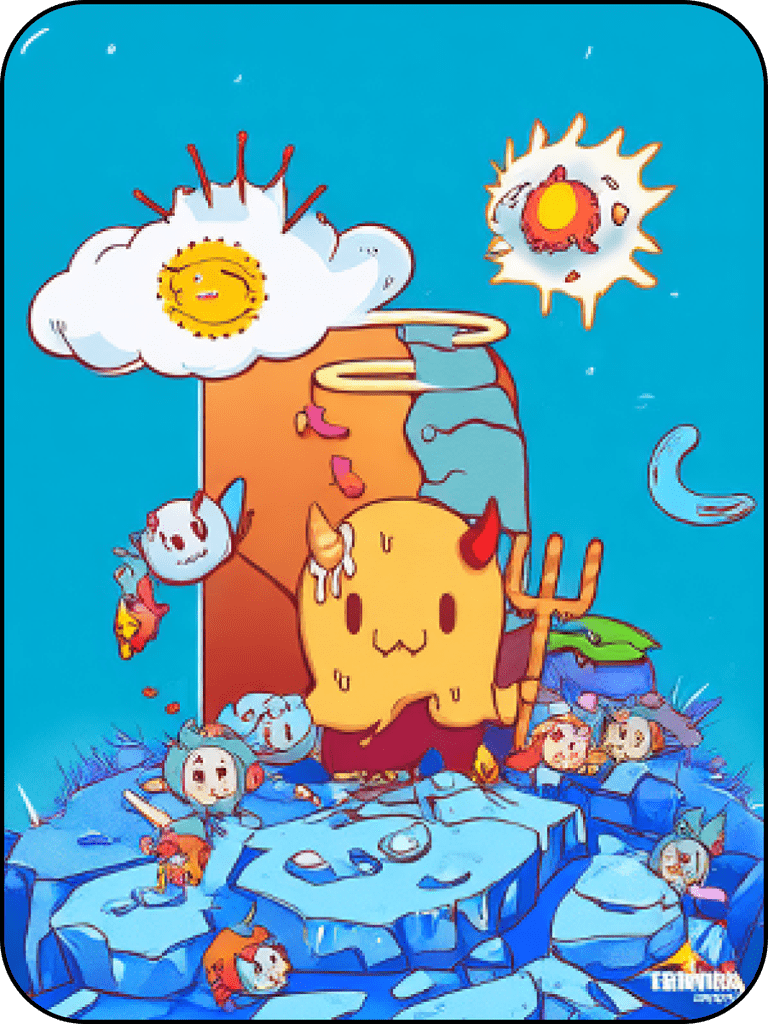} & 
The center of the image is a yellow, melting-ice cream-like character with devil horns, holding a trident. It stands on a rock, surrounded by many small creatures wearing tiny halos or having little devil wings.  The background is a refreshing blue sky with a sun and moon and some clouds.  The overall style is cute and fantastical, with bright colors and lively designs, creating a magical atmosphere.
& 

\centering
\includegraphics[width=0.8\linewidth, height=3cm, keepaspectratio]{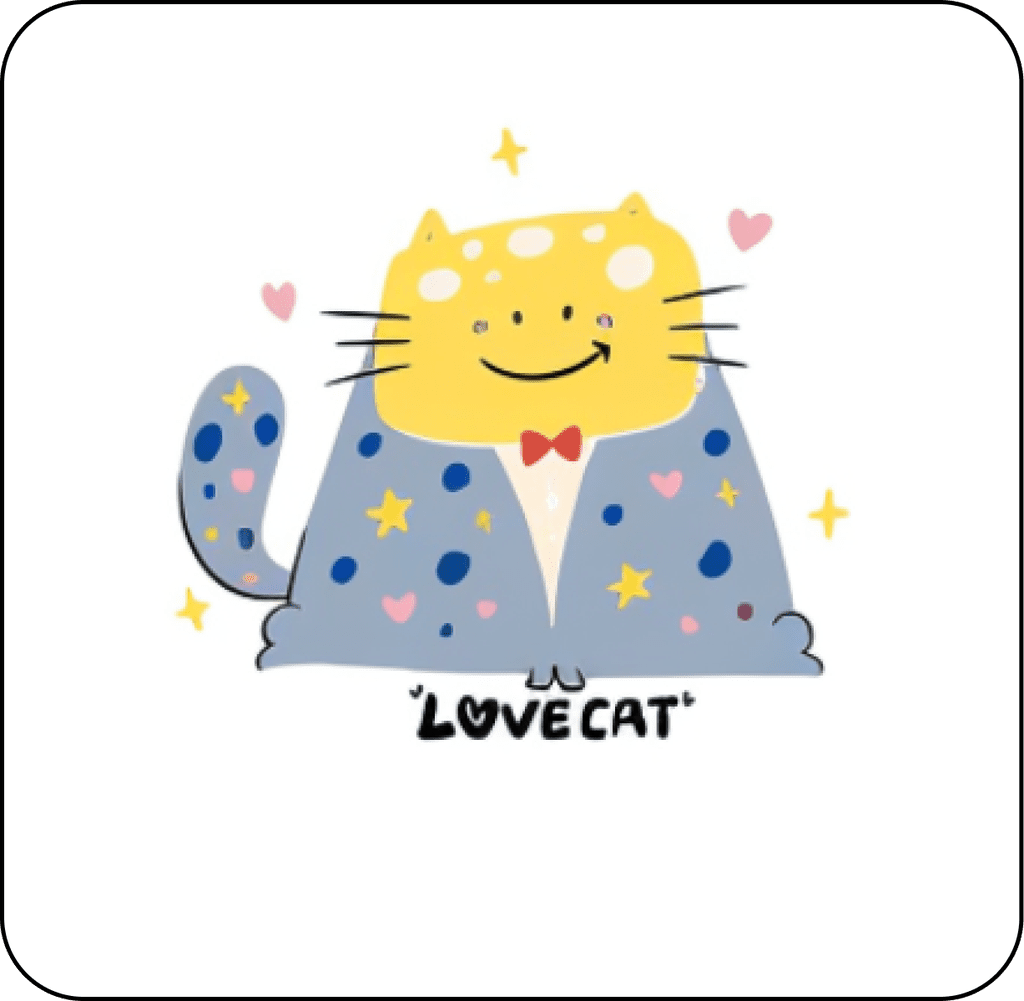} & 

A yellow cartoon cat with black eyes, whiskers, and a red mouth. It wears a red bow tie and a blue coat decorated with patterns like stars, dots, and hearts. At the bottom of the image is the word ``LOVE CAT". 
\\
\midrule

\centering
\includegraphics[width=0.8\linewidth, height=3cm, keepaspectratio]{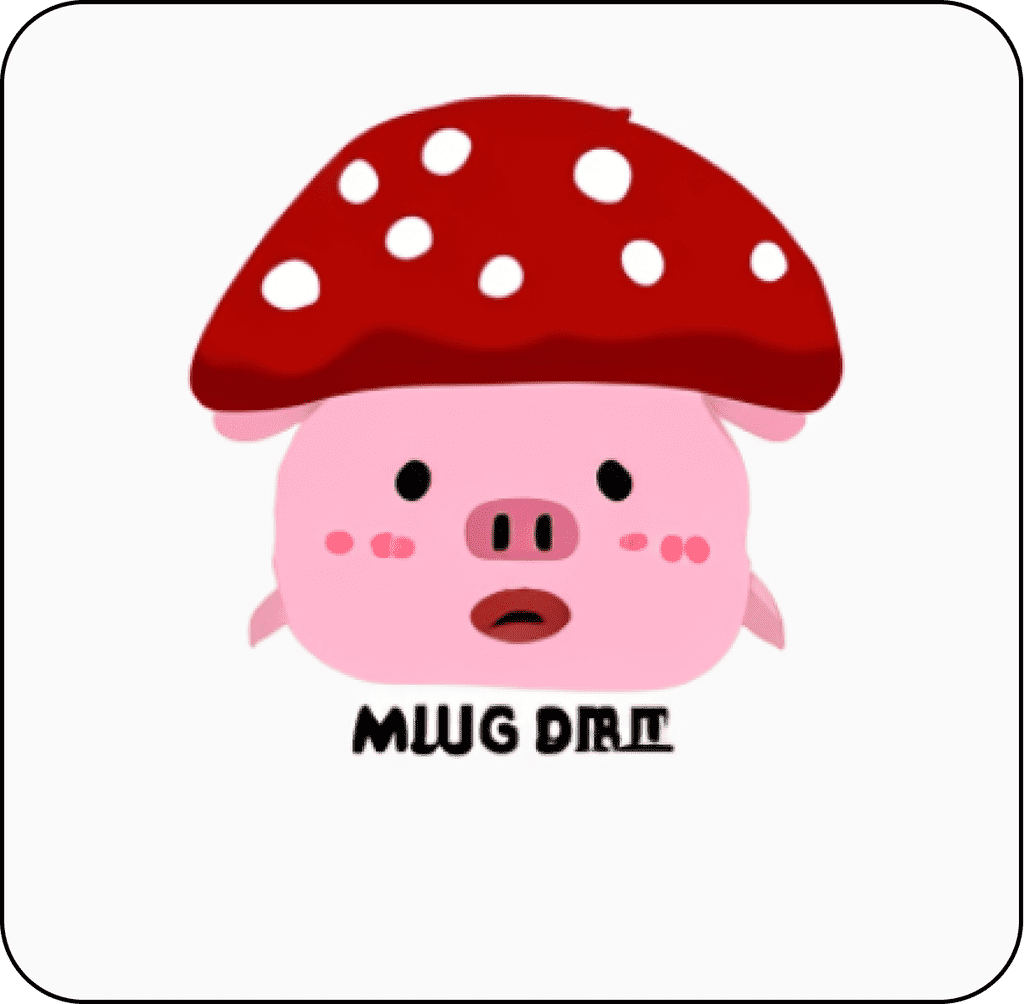} & 
A pink cartoon pig with black eyes, a red nose, and a red mouth. It wears a red mushroom cap decorated with white dots. At the bottom of the image is the word ``MUSH ROOM". 
& 

\centering
\includegraphics[width=0.8\linewidth, height=3cm, keepaspectratio]{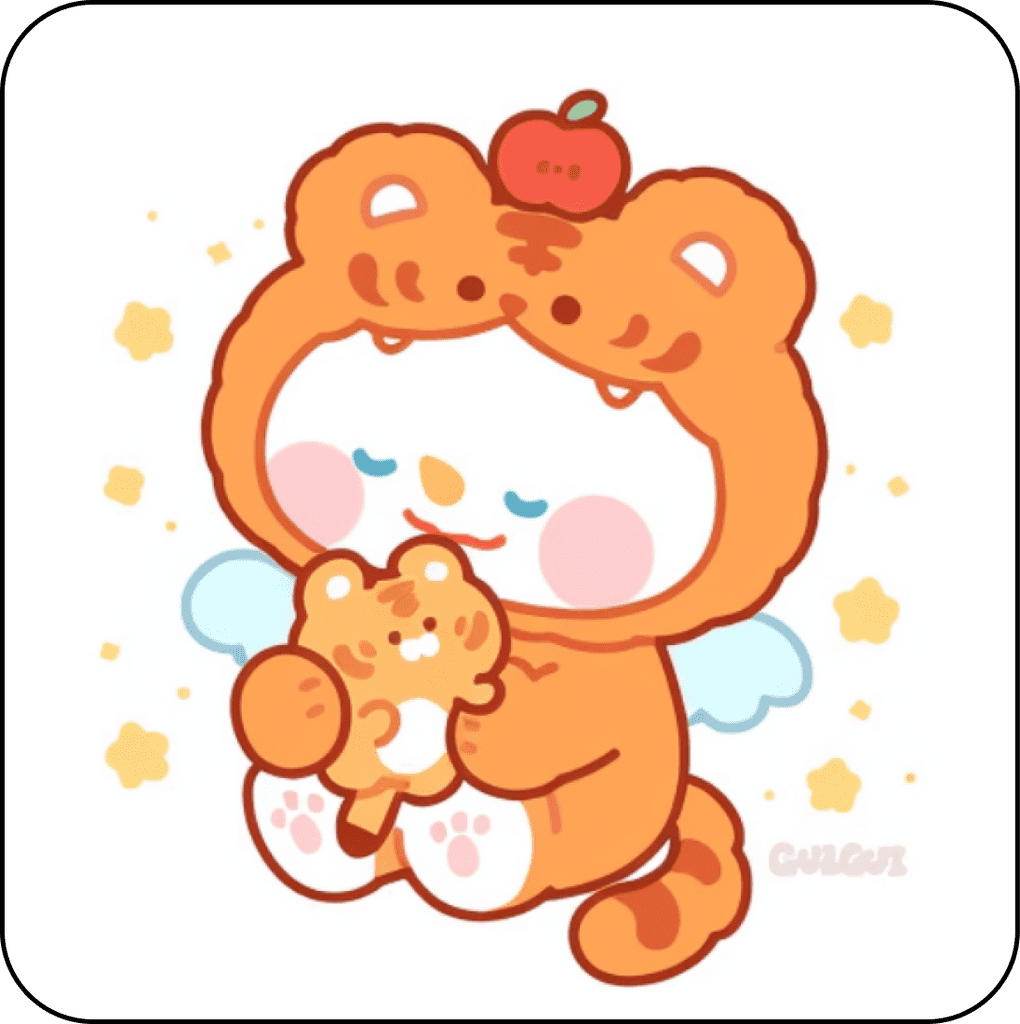} & 

A cartoon character is sitting on the ground with its eyes closed and a smile on its face. It is wearing an orange tiger jumpsuit with a red apple on the hood. It has a pair of white wings and is holding a small orange tiger plush toy. The background is white with yellow stars.
\\
\midrule

\centering
\includegraphics[width=0.8\linewidth, height=3cm, keepaspectratio]{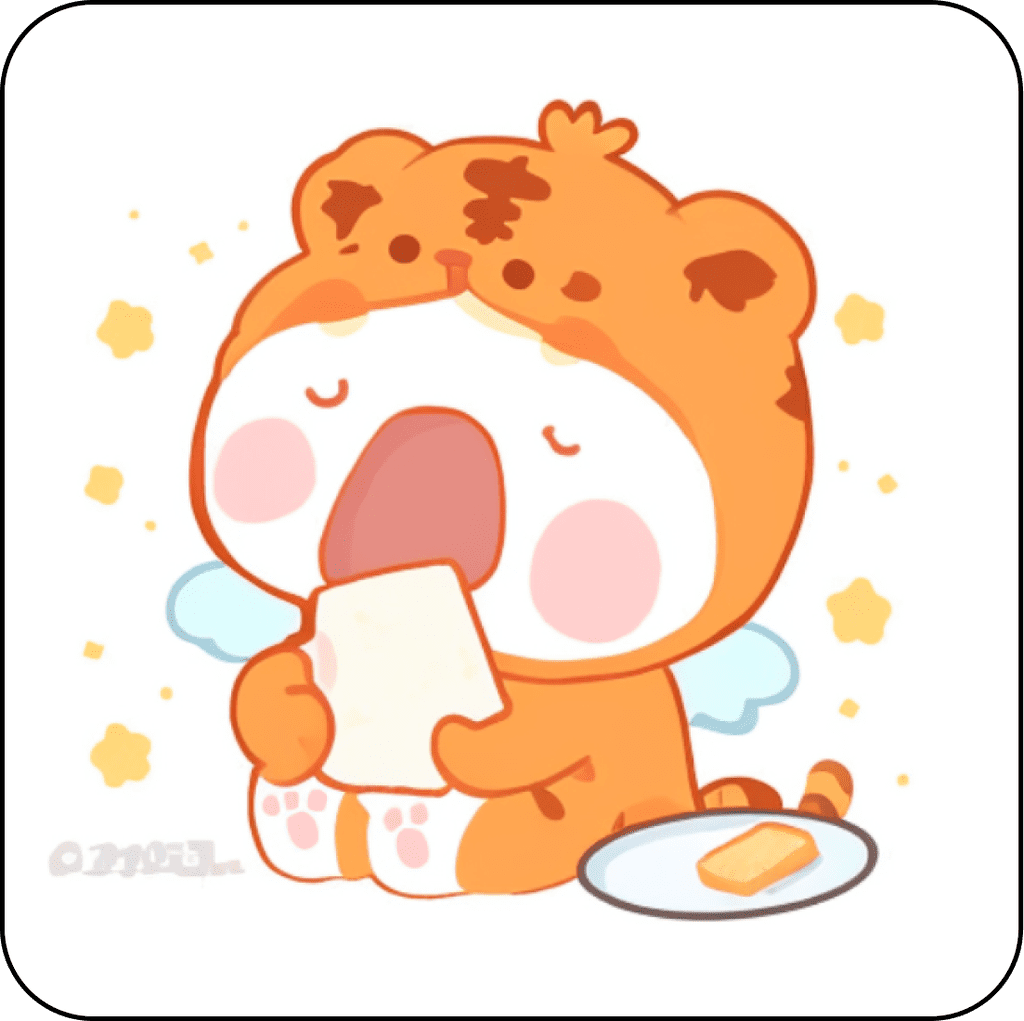} & 
A cartoon character is sitting on the ground with its mouth open, seemingly eating a piece of white rice cake. Its eyes are narrowed, and it has a smile on its face. It is wearing an orange tiger jumpsuit. The background is white with yellow stars and pink flowers. To the left of the cartoon character, there is a plate with a piece of white rice cake on it.
& 

\centering
\includegraphics[width=1.0\linewidth, height=3cm, keepaspectratio]{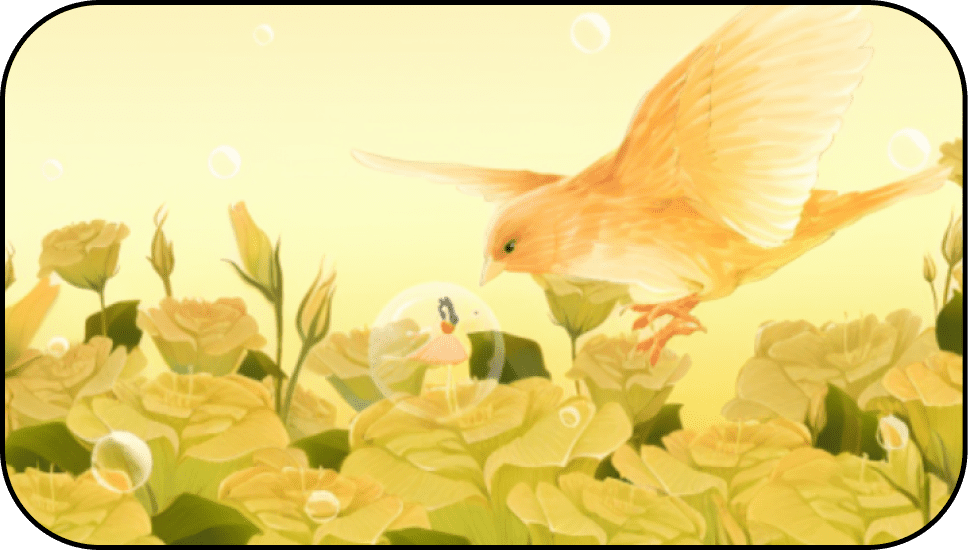} & 

The image shows a small orange-yellow bird with its wings spread, looking down at a little girl in a ballet dress enclosed in a transparent bubble.  Surrounding them are blooming pale yellow roses, the background is warm and soft, dotted with white bubbles of varying sizes, creating a peaceful and serene atmosphere. The style is fresh, elegant, with soft colors and smooth lines.
\\
\midrule

\centering
\includegraphics[width=1.0\linewidth, height=3cm, keepaspectratio]{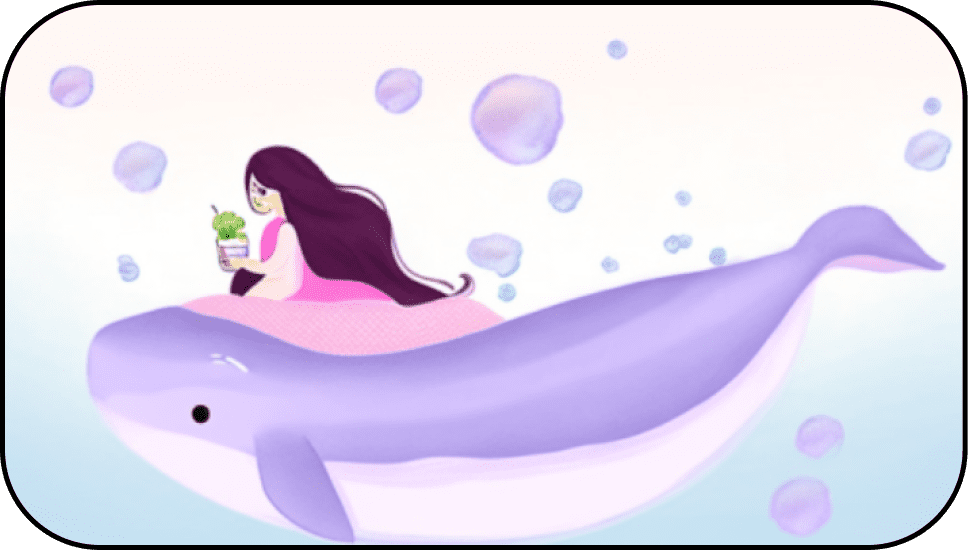} & 
The image depicts a girl with long, flowing dark hair wearing a pink dress, sitting on the back of a pale purple whale, holding a cup of green drink. The whale swims in light blue water, surrounded by colorful bubbles of different sizes. The background colors gradually change from light blue to light purple. The overall atmosphere is dreamy and romantic, with fresh and soft colors. The style is fresh, elegant, with smooth lines.
& 

\centering
\includegraphics[width=1.0\linewidth, height=3cm, keepaspectratio]{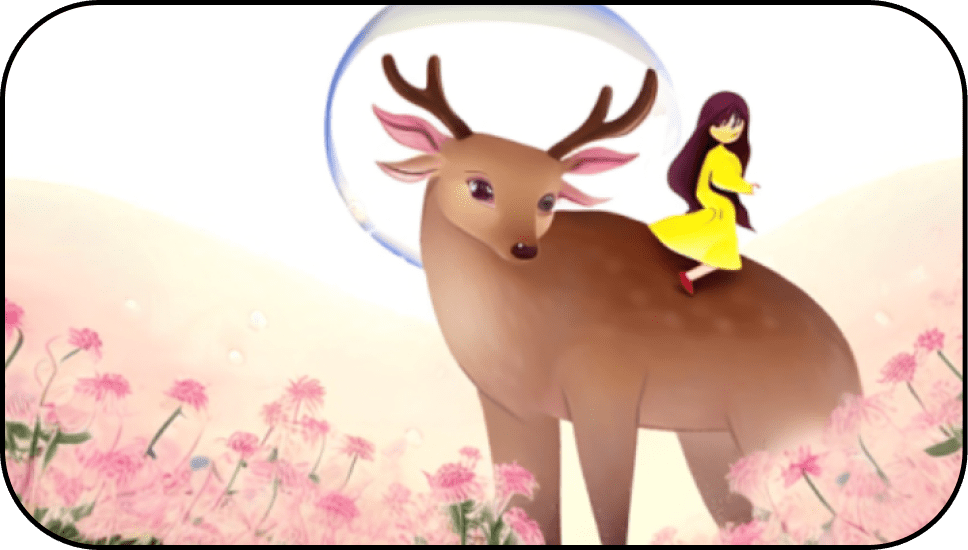} & 

The image portrays a brown deer with a girl in a yellow dress riding on its back. They stand in a field of pink flowers, against a soft beige background. A large transparent bubble is featured in the center above them, with sunlight streaming through, creating a warm and dreamy atmosphere. The style is fresh, elegant, with soft colors and smooth lines.
\\
\midrule

\centering
\includegraphics[width=0.8\linewidth, height=3cm, keepaspectratio]{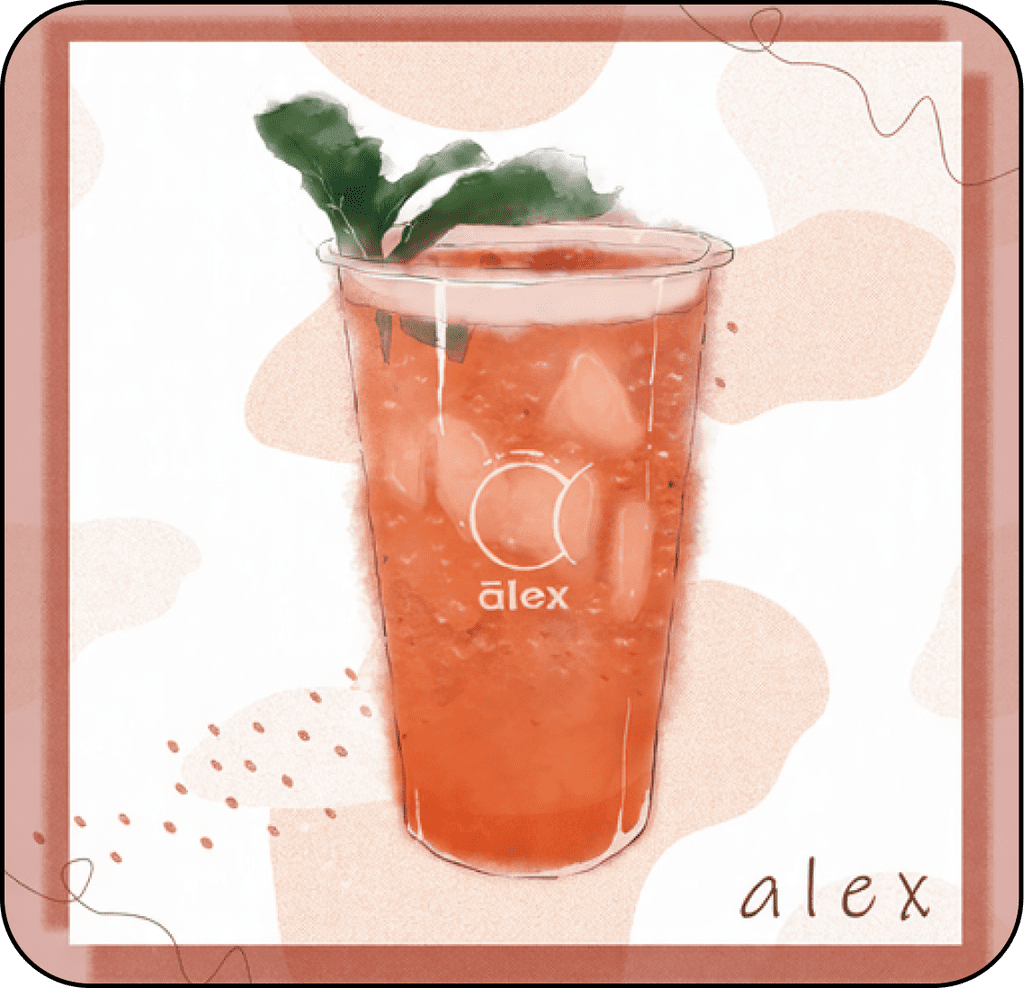} & 
A clear plastic cup with the ``alex" logo contains a pink drink. A green leaf and ice cubes are visible within the beverage. The background is white with abstract red shapes and lines. 
& 

\centering
\includegraphics[width=0.8\linewidth, height=3cm, keepaspectratio]{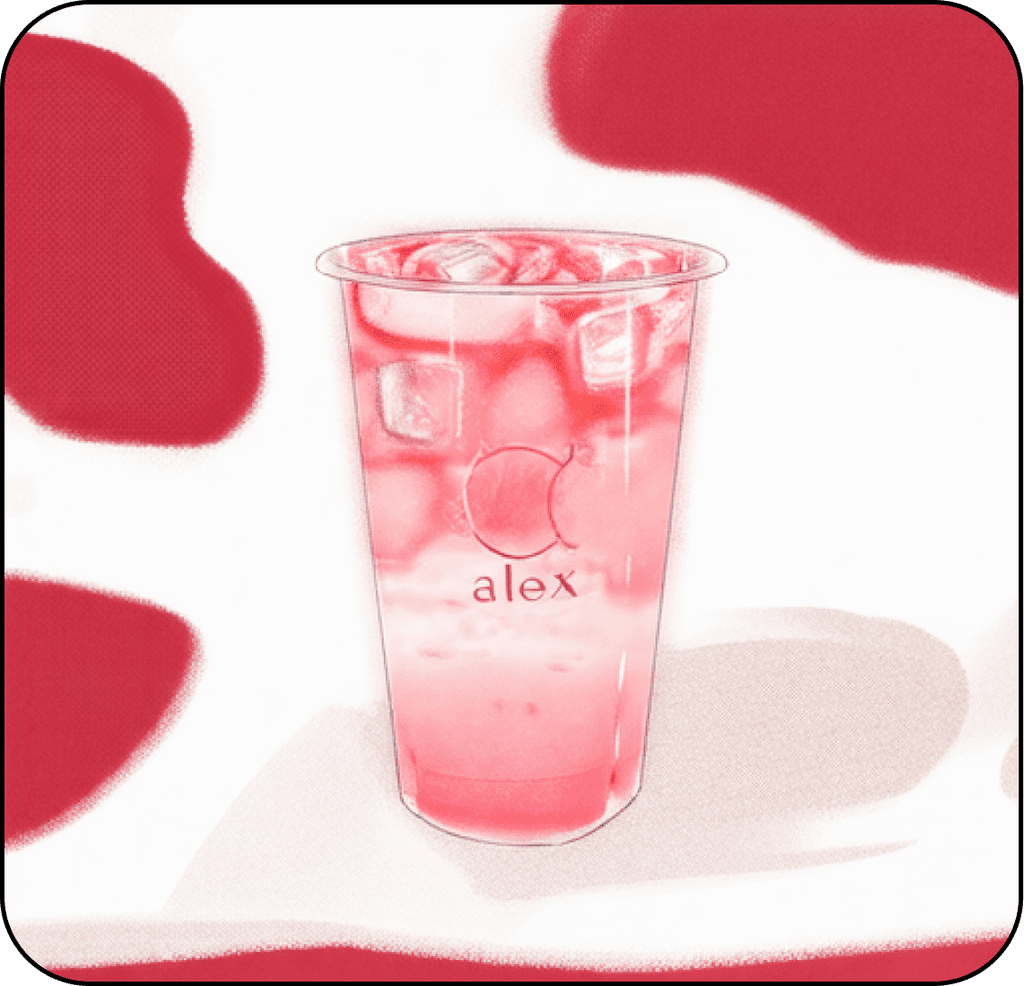} & 

A clear plastic cup with the ``alex" logo contains a red drink. The background is white with abstract red shapes and lines. 
\\
\midrule

\centering
\includegraphics[width=0.6\linewidth, height=3cm, keepaspectratio]{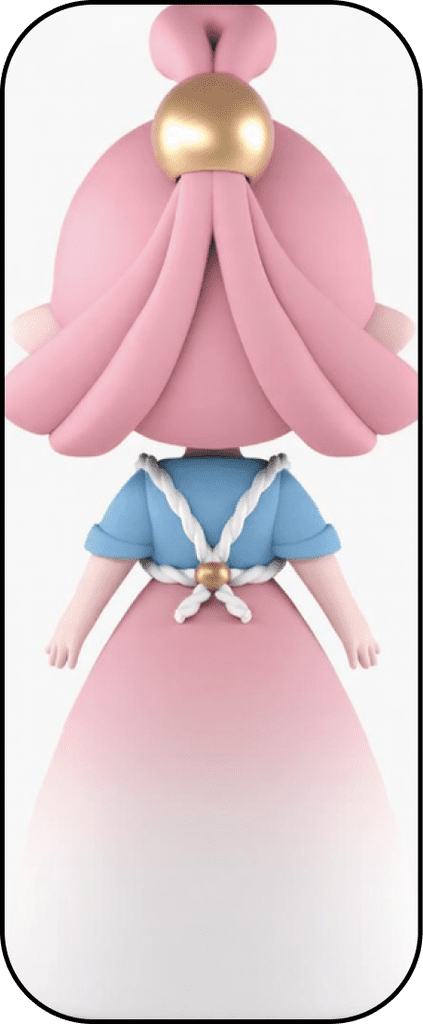} & 
This is a 3D rendered image showcasing the back view of a cartoon girl. She has pink hair tied up in a high bun, adorned with a golden spherical ornament. She is wearing a blue top with a white apron tied around her waist, featuring rope-like detailing. Her dress is pink, gradually fading to white at the bottom. She has pointed ears, exhibiting an overall cute and sweet style.
& 

\centering
\includegraphics[width=0.6\linewidth, height=3cm, keepaspectratio]{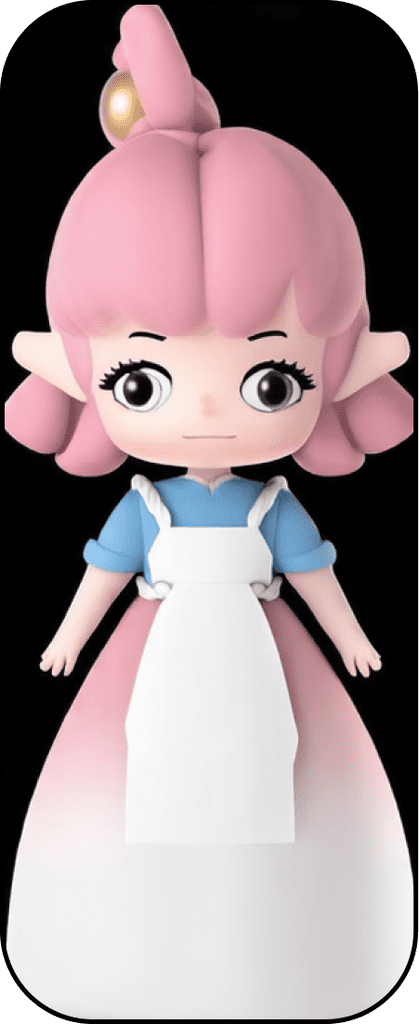} & 

This is a 3D rendered image showcasing the front view of a cartoon girl. She has pink hair tied up in a high bun, adorned with a golden spherical ornament. She has large eyes and cute pointed ears. She is wearing a blue short-sleeved top with a white apron. She is wearing a pink dress, gradually fading to white at the bottom. The overall style of the girl is cute and sweet.
\\

\end{longtable}
}

\end{document}